\documentclass{article}
\usepackage{graphicx} 
\usepackage{lineno,hyperref}
\usepackage{amsmath}
\usepackage{amsfonts}
\usepackage{xcolor}
\usepackage{geometry}
\usepackage{textcomp}
\usepackage{comment}
\usepackage{float}

\usepackage[table]{xcolor}

\usepackage{capt-of}
\usepackage{booktabs}
\usepackage{varwidth}
\usepackage{amssymb}

\usepackage[ruled,vlined]{algorithm2e}

\usepackage[british]{babel}
\usepackage{hhline}
\usepackage{multirow}
\usepackage{authblk}
\usepackage{soulpos}

\usepackage{soul}
\definecolor{OliveGreen}{rgb}{0,0.6,0}

\usepackage{algorithmic}
\usepackage{array}

\newtheorem{theorem}{Theorem}[section]

\usepackage{geometry}

\title{\textbf{Anant-Net: Breaking the Curse of Dimensionality with Scalable and Interpretable Neural Surrogate for High-Dimensional PDEs}}
\author[]{Sidharth S. Menon}
\author[]{Ameya D. Jagtap\thanks{Corresponding author: Ameya D. Jagtap (ajagtap@wpi.edu, ameyadjagtap@gmail.com) \\  \\ Accepted in \textit{Computer Methods in Applied Mechanics and Engineering, Elsevier}.}}

\affil[]{\textit{\small{Aerospace Engineering Department, Worcester Polytechnic Institute, Worcester, MA 01609, USA.}}}
\date{}

\begin{document}
\maketitle

\begin{abstract}
High-dimensional partial differential equations (PDEs) arise in diverse scientific and engineering applications but remain computationally intractable due to the curse of dimensionality. Traditional numerical methods struggle with the exponential growth in computational complexity, particularly on hypercubic domains, where the number of required collocation points increases rapidly with dimensionality. Here, we introduce \textit{Anant-Net}, an efficient neural surrogate that overcomes this challenge, enabling the solution of PDEs in high  dimensions. Unlike hyperspheres, where the internal volume diminishes as dimensionality increases, hypercubes retain or expand their volume (for unit or larger length), making high-dimensional computations significantly more demanding. Anant-Net efficiently incorporates high-dimensional boundary conditions and minimizes the PDE residual at high-dimensional collocation points. To enhance interpretability, we integrate Kolmogorov–Arnold networks into the Anant-Net architecture. We benchmark Anant-Net's performance on several linear and nonlinear high-dimensional equations, including the Poisson, Sine-Gordon, and Allen-Cahn equations, as well as transient heat equations, demonstrating high accuracy and robustness across randomly sampled test points from high-dimensional spaces. Importantly, Anant-Net achieves these results with remarkable efficiency, solving 300-dimensional problems on a single GPU within a few hours. We also compare Anant-Net's results for accuracy and runtime with other state-of-the-art methods. Our findings establish Anant-Net as an accurate, interpretable, and scalable framework for efficiently solving high-dimensional PDEs. The Anant-Net code is available at  \href{https://github.com/ParamIntelligence/Anant-Net}{https://github.com/ParamIntelligence/Anant-Net}.
\end{abstract}

\vspace{0.2cm}

 \begin{small}Keywords: \textit{High-dimensional PDEs}; \textit{Scalable and Interpretable Architecture}; \textit{Physics-Informed Neural Surrogates}; \\ \textit{Kolmogorov–Arnold networks (KAN)}.
\end{small}

\section{Introduction}
Physics-informed deep learning (PIDL) represents a rapidly advancing framework that integrates known governing physical laws, typically formulated as PDEs, into the training process of deep neural networks. In contrast to conventional data-driven models that rely solely on observational data, PIDL incorporates physical constraints to guide learning, thereby enhancing generalization, reducing data dependence, and improving interpretability. This synthesis of physics and deep learning has demonstrated broad applicability in solving forward and inverse problems across scientific and engineering domains, particularly in scenarios involving limited, noisy, or deceptive data. Key methodologies under the PIDL umbrella include physics-informed neural networks (PINNs) \cite{raissi2019physics,abbasi2025challenges,jagtap2022deepLL,abbasi2025history}, which embed PDE constraints via automatic differentiation; sparse identification of nonlinear dynamics (SINDy) \cite{brunton2016discovering,fasel2021sindy}, which infers governing equations by promoting sparsity in learned representations; and physics-informed neural operators \cite{li2020fourier,lu2021learning,wang2021learning, peyvan2024riemannonets,goswami2024learning}, which approximate solution operators across function spaces to model families of PDEs. These approaches are particularly well-suited for high-dimensional problems, where traditional numerical solvers suffer from the curse of dimensionality. 

High-dimensional PDEs are integral to various scientific and engineering domains, including quantum mechanics, financial mathematics, and optimal control. Their solutions provide crucial insights into complex, multi-scale phenomena that cannot be accurately captured using lower-dimensional approximations. However, solving these equations efficiently remains a significant challenge due to the \textit{curse of dimensionality}, the exponential growth in computational complexity and data requirements as the number of dimensions increases. This issue is particularly pronounced in numerical approximations, making high-dimensional PDEs computationally intractable with conventional methods.
In this work, we introduce \textit{Anant-Net} (with \textit{Anant} meaning infinite in \textit{Sanskrit}). While the method does not address infinite-dimensional problems, it is designed to tackle finite but high-dimensional PDEs. We believe that Anant-Net provides an efficient and scalable approach for solving such problems. The Anant-Net is designed to mitigate the curse of dimensionality and enable the efficient solution of high-dimensional PDEs critical to real-world applications. Our approach provides an efficient framework for handling large datasets, whose size typically scales exponentially with dimensionality. While purely data-driven models might seem a viable alternative, they often struggle with excessive data requirements, making them impractical for high-dimensional problems. To address this, we leverage physics-informed neural surrogates, incorporating the underlying PDE as an additional constraint. This integration significantly reduces the dependency on large datasets, offering a more scalable and computationally efficient framework for solving high-dimensional PDEs.
\begin{figure}[h]
\centering
\centering
\includegraphics[trim={0cm 0cm 0cm 0cm},clip=true,scale=0.35]{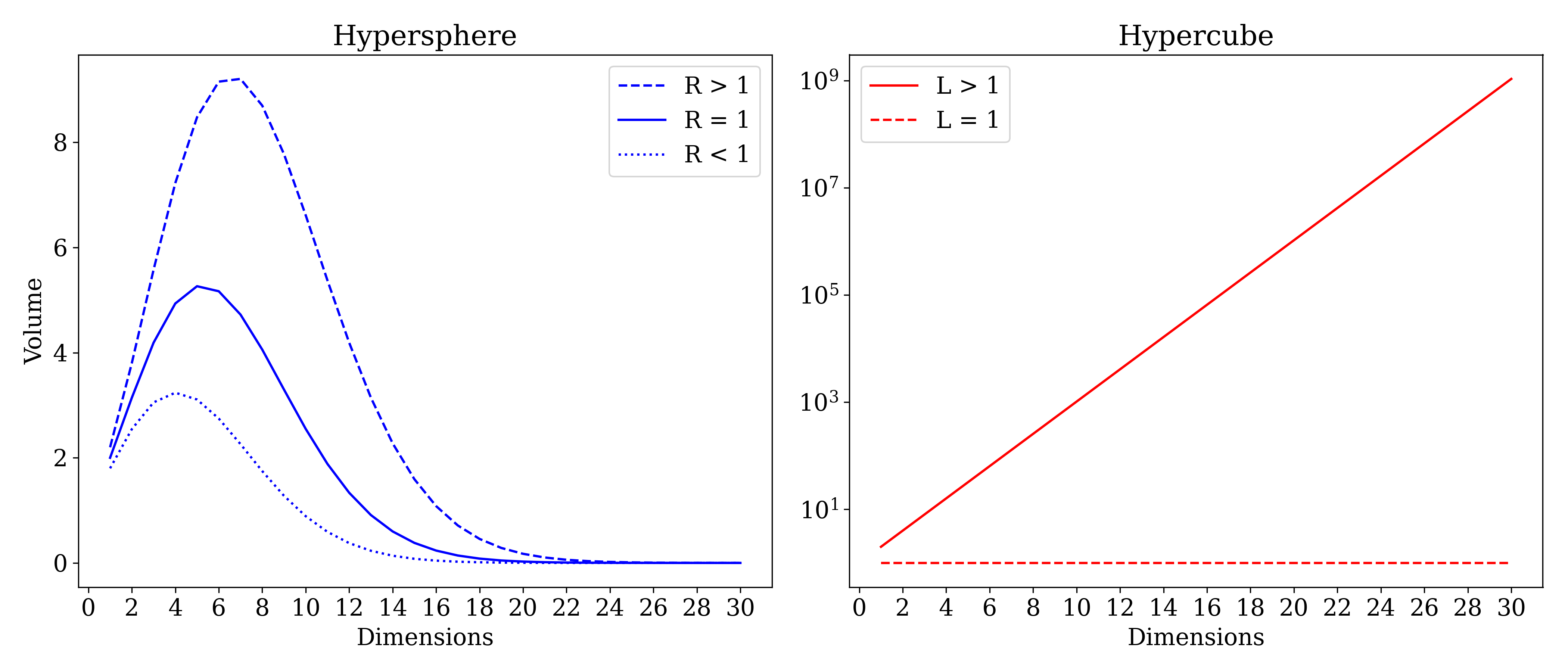}
\caption{Variation of the volume of a hypersphere/hypercube in high-dimensional spaces, highlighting the motivation for Anant-Net in addressing high-dimensional PDEs in hypercube domain. Here, $R$ denotes the radius of the hypersphere, and $L$ represents the edge length of the hypercube.}
\label{fig:volume}
\end{figure}
In high-dimensional spaces, the volume of a hypersphere decreases as dimensionality increases when the radius $R>1, <1, = 1$, eventually approaching zero. Conversely, the volume of a hypercube grows exponentially with dimension if the edge length $L>1$, while it remains constant at $L=1$; see figure \ref{fig:volume}. Since the volume of a hypersphere shrinks with increasing dimensions, enforcing data constraints exactly within this region makes the problem more easier. In contrast, for a hypercube with $L > 1$, the exponentially growing volume poses significant computational challenges, making it difficult to efficiently solve high-dimensional PDEs. 
\begin{table}[h!]
\begin{center}
\begin{tabular}{c | c | c | c}  
 \hline
 \textbf{Range} & \textbf{21 Dims.} & \textbf{99 Dims.} & \textbf{300 Dims.}\\ [0.05ex] 
 \hline\hline
 $[0,1]$ ($L=1$)& 0.4 $\pm$ 2.17e-03 & 0.7 $\pm$ 2.67e-03  & 0.5 $\pm$ 3.44e-03\\
 $[-1,1]$ ($L>1$) & 1.49 $\pm$ 1.49e-02 & 7.09 $\pm$ 3.71e-02 & 6.00 $\pm$ 3.81e-02\\
 \hline
\end{tabular}
\end{center}
\caption{Anant-Net: Mean and standard deviation (computed over 10 random realizations) of the \textbf{\% relative $L_2$ testing error} on randomly selected test points for the high-dimensional Poisson equation.}
\label{tab1:weak_strong_runtime}
\end{table}
Table \ref{tab1:weak_strong_runtime} presents the mean and standard deviation of percentage relative $L_2$ error (testing error) of Anant-Net evaluated on randomly selected test points for the high-dimensional Poisson equation. These testing errors show that Anant-Net can accurately solve high-dimensional PDEs.

\subsection{Related Work}
In the context of high-dimensional PDEs, two primary factors determine the efficacy of any method based on scientific machine learning: (1) the sampling strategy for data or collocation points, and (2) the computation of gradients for residual losses via automatic differentiation, which are used as soft constraints in training physics-informed models. It is evident that a significant number of existing methods represent a trade-off between these factors, which ultimately influences their scalability and suitability for solving real-world problems of practical relevance.  While various methods exist for solving high-dimensional PDEs, this work focuses on neural surrogates that not only leverage available data but also incorporate physical laws, ensuring the consistency of the solutions with underlying governing principles.  In recent years, the challenge of solving high-dimensional PDEs has been addressed by several researchers through the use of neural surrogate methods. Han et al. \cite{han2018solving} introduced a deep neural network framework that reformulates high-dimensional PDEs as backward stochastic differential equations. However, their approach relied on approximating the gradient of the solution using a neural network, which introduced scalability issues, particularly in scenarios involving long-time integration. This limitation was subsequently addressed by Raissi et al. \cite{raissi2018forward}, who proposed an alternative method that directly models the solution of the PDE with a neural network, rather than its gradient. The efficacy of this improvement was demonstrated through its application to the Black-Scholes-Barenblatt and Hamilton-Jacobi-Bellman equations in 100 dimensions. Hu et al. \cite{hu2024tackling} introduced a stochastic dimensions gradient descent (SDGD) framework based on physics-informed neural networks to solve high-dimensional PDEs on hypersphere. This method stochastically samples the dimensions of the PDE and computes the residuals only in the selected sample dimensions, rather than performing the computation across all dimensions. Their approach significantly reduced the computational burden on hardware, leading to notable speed-ups and improved accuracy compared to other contemporary methods for high-dimensional PDEs. However, their method bypassed the need for data or boundary point sampling by directly enforcing boundary conditions within the model, particularly for hypersphere, where the internal volume approaches zero as the number of dimensions increases (see Figure \ref{fig:volume}). In the similar line of work, Hu et al.\cite{hu2024tackling2} proposed Monte Carlo fractional PINN (MC-fPINN) and its extension, MC-tfPINN for tempered fractional PINN (an extension to fPINN \cite{pang2019fpinns}) to solve extremely high-dimensional problems. Further, an alternate approach was proposed by the same authors \cite{hu2024score} extended fractional PINN to solve high-dimensional Fokker-Planck-Lévy (FPL) equations in high-dimensions. Adaptive training methods have also been proposed as supplementary strategies to enhance the accuracy of physics-informed neural network models for high-dimensional PDEs \cite{zeng2022adaptive}. 

Wang et al. \cite{wang2024solving} introduced tensor neural networks for solving high-dimensional PDEs, wherein the independent input features are distributed across sub-networks to construct a low-rank basis representation of the input space. The solution is then mapped as a superposition of these low-dimensional basis vectors. The advantages of their method were demonstrated through experiments for both homogeneous and non-homogeneous boundary value problems up to 20 dimensions. Despite its superior accuracy, the method faces scalability limitations as the number of sub-networks increases with dimension, leading to significant computational burdens on hardware. More recently, De et al. \cite{de2025approximation} applied physics-informed extreme learning machines to high-dimensional PDEs, framing the training of random neural networks as a least-squares problem to determine the unknown weights in the final layer, while keeping the weights in the previous hidden layers fixed. Their approach achieved both superior accuracy and low computational cost for high-dimensional problems, with training demonstrated for up to 100 dimensions. Unlike automatic differentiation, their method used finite differences to compute gradients for high-dimensional PDEs. Among many strategies that were proposed to tackle the curse of dimensionality, one of the most promising strategy is to avoid the discretizations in space/time or using a grid-based format for high-dimensional data. To this end,  Darbon et al. \cite{darbon2021some} proposed two frameworks based on deep neural networks to approximate the viscosity solutions of the Hamiltonian Jacobi PDEs under certain reasonable assumptions on the PDE. Richter \cite{richter2021solving} et al. proposed a tensor train format for parabolic PDEs wherein the in-built orthogonality of their proposed format enabled fast and efficient optimization in stochastic backward formulations of the PDE. Meng et al. \cite{meng2022sympocnet} proposed an architecture that fuses SympNet with PINNs to tackle curse of dimensionality in constrained optimal multi-agent path planning problems. They investigated various formulations of the loss function to arrive at the one with least violation and later post-processed with pseudo-spectral method to achieve spectral accuracy for their path trajectories. In a similar line of work, Varghese et al.\cite{varghese2025sympgnns} SympNet was extended to Symplectic Graph Neural Networks (SympGNNs) that can effectively handle system identification in high-dimensional Hamiltonian systems such as a 2000-particle molecular dynamics simulation in a two-dimensional Lennard-Jones potential.


Several studies have leveraged the weak formulation of PDEs to train deep learning models, particularly for high-dimensional problems. A pioneering approach in this direction is the Deep Ritz method \cite{yu2018deep}, which utilizes the variational formulation of PDEs by minimizing an energy functional. This method, designed to address high-dimensional PDEs, employs a ResNet-based architecture and is trained using stochastic gradient descent. It systematically samples data from both the interior and the boundary of the domain, distinguishing itself from approaches such as that of Hu et al. \cite{hu2024tackling}, which circumvent boundary data sampling. Another method utilizes the variational formulation or the weak form of the PDE was proposed by Samaniego et al. \cite{samaniego2020energy}. However, their method was limited to solving low dimensional PDEs and couldn't be extended for the cases discussed henceforth in this paper. Building on this framework, the Weak Adversarial Network (WAN) \cite{zang2020weak} was introduced to solve high-dimensional PDEs by reformulating the problem of finding a weak solution as an operator norm minimization task. In contrast to conventional numerical methods such as finite difference schemes, WAN demonstrated computational efficiency, stability, and the ability to mitigate the curse of dimensionality. The methodology employs generative adversarial network (GAN) framework, where the weak solution and test functions are parameterized by deep neural networks, termed the primal and adversarial networks, respectively. More recently, the WAN framework has been extended through the introduction of XNODE-WAN \cite{oliva2022towards}, which replaces the primal network with a Neural Ordinary Differential Equation (Neural ODE) model. This enhancement has resulted in improved computational speed and accuracy, particularly for time-dependent high-dimensional PDEs. These approaches systematically sample data across both the interior and the boundary of the computational domain to enhance the robustness of their solutions.

\subsection{Our Contributions}
The main contributions of this work are as follows:
\begin{itemize}
\item We propose the Anant-Net architecture, a scalable and efficient neural architecture for solving high-dimensional PDEs defined on hypercubic domains. The network is designed to maintain high accuracy while addressing the computational challenges inherent to high-dimensional settings, particularly for hypercubes with side length $L > 1$. The Anant-Net framework leverages tensor product structures to compute PDE solutions across the specified batch-size dimensions and performs dimension-wise sweeps over the full batch to handle high-dimensional problems. This tensor product formulation enables the efficient processing of large volumes of training data and collocation points, which is important for scaling to high dimensions.

\item We identify data sampling as an important and resource-intensive aspect when solving high-dimensional PDEs. To address this, we propose a sampling-based strategy that reveals the relationship between accuracy, speedup, and the volume of boundary/data points as the PDE dimension increases. Moreover, we develop an efficient framework to handle large datasets, enabling scalable training and evaluation.

\item To mitigate the high cost associated with computing derivatives in high-dimensional spaces, Anant-Net avoids exhaustive differentiation. Instead, it strategically samples a subset of dimensions and applies automatic differentiation selectively, thereby preserving the complex structure of the governing PDE while significantly reducing computational complexity.

\item We evaluate the performance of Anant-Net by reporting test accuracy on randomly sampled points from high-dimensional space after training. Specifically, we present the mean and variance of the test errors. Anant-Net demonstrates the ability to solve PDEs in up to 300 dimensions within a few hours on a single GPU, highlighting its scalability and computational efficiency.

\item We also introduce an interpretable variant of Anant-Net based on the Kolmogorov–Arnold Network (KAN), called as Anant-KAN, enabling further insights into the learned solution structures in high-dimensional settings. To the best of our knowledge, this is the first application of KAN to high-dimensional problems.

\item We benchmark Anant-Net against several state-of-the-art methods, reporting comparative results in terms of runtime as well as the mean and variance of relative testing error (\%), and observe that Anant-Net consistently outperforms the existing methods.
\end{itemize} 

The paper is organized as follows: Section 2 provides a detailed description of the Anant-Net architecture, including the methodologies employed for organizing training data and incorporating collocation points. Furthermore, it introduces the interpretable KAN-based variant of Anant-Net, referred to as Anant-KAN. Section 3 presents computational results for a representative set of high-dimensional boundary and initial value problems, encompassing both linear and nonlinear, as well as steady-state and transient cases. A comparative analysis against existing state-of-the-art architectures is also provided. To this end, Section 4 highlights the key findings of this work and provides a detailed discussion of the advantages and limitations of the proposed methodology.

\section{Methodology: Anant-Net Architecture}
Separable physics-informed neural networks (SPINNs) \cite{cho2024separable} were recently proposed as an architectural framework for solving high-dimensional PDEs. In this approach, the independent features of an $d$-dimensional PDE are distributed across $n$-body subnetworks. Leveraging the structure of the architecture, SPINNs employ Einstein summation to efficiently combine the outputs of the $n$ independent subnetworks to construct the overall solution field. This strategy has been shown to perform exceptionally well in terms of both accuracy and computational efficiency for moderate-dimensional problems.
However, the method exhibits limitations in scalability, as its performance degrades for problems with dimensionality greater than five. This is primarily due to the computational burden associated with naively applying Einstein summation over a increasing number of subnetworks, which becomes intractable as dimensions increases. Despite this limitation, a key strength of the SPINN framework lies in its ability to process a large volume of training data and collocation points, achieving high throughput without incurring additional computational or memory overhead, even in the absence of specialized hardware.
\begin{figure}[htpb]
\centering
\centering
\includegraphics[trim={0cm 0cm 0cm 0cm},clip=true,scale=0.72]{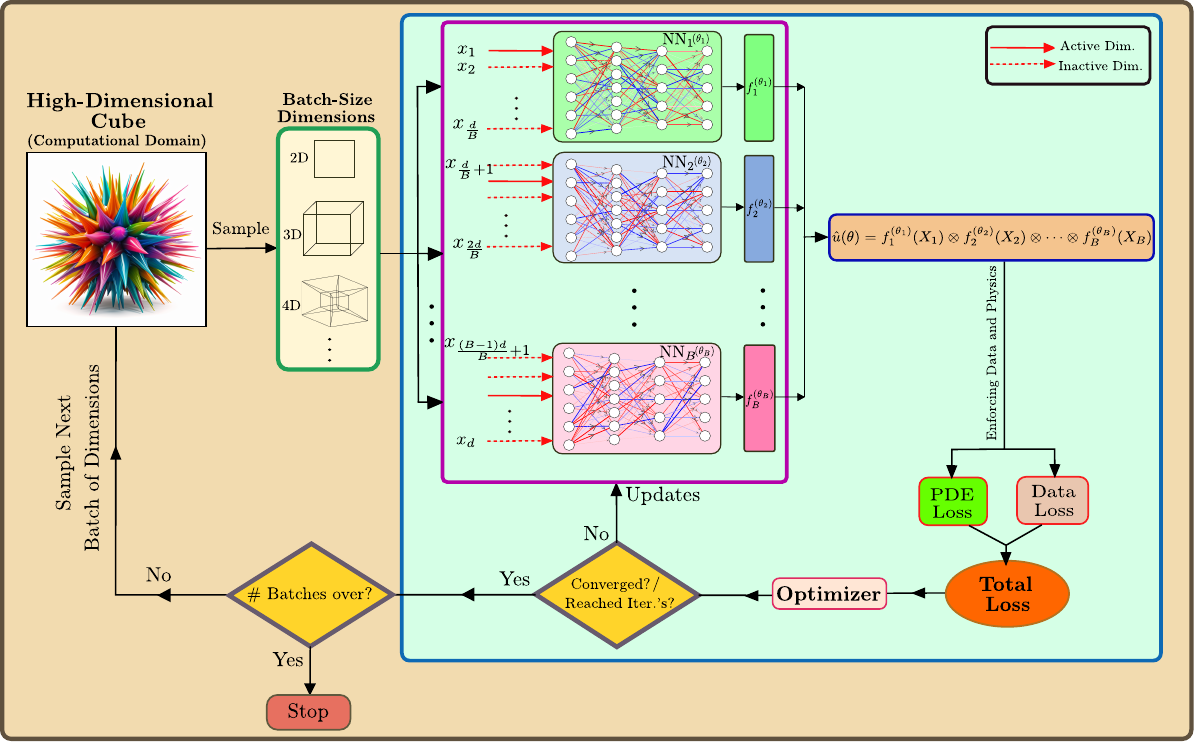}
\caption{Schematic representation of general Anant-Net architecture with batch-size $B$, highlighting its core components and computational flow for solving high-dimensional PDEs.  We partition the $d$-dimensional input space into $B$ equal segments, each of size $d/B$. The first body network receives the subset ${x_1, x_2, \dots, x_{\frac{d}{B}}}$. The second body network processes ${x_{\frac{d}{B} + 1}, \dots, x_{\frac{2d}{B}}}$, while the third network processess ${x_{\frac{2d}{B} + 1}, \dots, x_{\frac{3d}{B}}}$, and so on. The final (i.e., $B^\text{th}$) body network is assigned the remaining inputs ${x_{\frac{(B-1)d}{B} + 1}, \dots, x_d}$.}
\label{fig:AnantN}
\end{figure}
The primary objective of the proposed architecture is to enable the scalable solution of high-dimensional PDEs. As discussed in the previous section, several prominent network architectures that report strong performance on high-dimensional PDE benchmarks often circumvent this challenge by embedding boundary conditions directly into the training process, thereby avoiding the explicit sampling of interior and boundary data points.
In contrast, the present work addresses this limitation by introducing the Anant-Net architecture capable of efficiently processing large volumes of training and collocation data, a necessary condition for robust performance at scale. Our method not only accommodates the data-intensive demands inherent to high-dimensional problems but also consistently achieves high accuracy on randomly chosen test data for different problems.

Figure \ref{fig:AnantN} shows the general architecture of Anant-Net. The architecture comprises a number of body networks equal to the batch size. Let 
$u$ denotes the solution of a $d$-dimensional PDE, which is the unknown to be approximated by Anant-Net in a $d$-dimensional hypercube. For brevity, we consider a batch size of three, the Anant-Net prediction of the solution field, denoted by $\hat{u}_{\theta}$ , is expressed as:
\begin{equation}\label{AN}
\hat{u}_{\theta}(X_1, X_2, X_3) =  f^{(\theta_1)}_{1}(X_1) \otimes f^{(\theta_2)}_{2}(X_2) \otimes f^{(\theta_3)}_{3}(X_3) = \sum_{j}^{emb} f^{(\theta_1)}_{1,j}(X_1)f^{(\theta_2)}_{2,j}(X_2)f^{(\theta_3)}_{3,j}(X_3),
\end{equation}
where $\otimes$ represents the tensor product notation and the inputs are partitioned as $X_1 = \{x_1, \dots, x_{d/3}\}$, $X_2 = \{x_{d/3+1}, \dots,  x_{2d/3}\}$ , and $X_3 = \{x_{2d/3+1}, \dots, x_d\}$, such that $X_1 \cap X_2 \cap X_3 = \emptyset$. The indices $p \in \mathrm{D}_1 = \{1,2,\ldots,d/3\}, q\in \mathrm{D}_2 = \{d/3+1,2,\ldots,2d/3\},  r\in \mathrm{D}_3 = \{2d/3+1,2,\ldots,d\}$ , corresponding to the number of batch elements, are stochastically sampled from the set of active dimensions, i.e.,  $\{x_p, x_q, x_r\} \subset X =  X_1 \cup X_2 \cup X_3 \in \mathbb{R}^d$, which sweeps over all dimensional indices.  The network parameters are given by $\theta = \bigcup \theta_i, \ \forall i $ (all body networks), where $f_i^{(\theta_i)}$ denotes the output vector of the $i^{\text{th}}$ body network (a feed-forward neural network), which can be given as $$f_i^{(\theta_i)}(X_i) = \Psi_L\circ\Psi_{L-1}\circ\dots\circ\Psi_1(X_i),$$ where $\Psi_\ell(Z) = \sigma_\ell(W_\ell Z + b_\ell)$, with $\sigma_\ell$, $W_\ell$, and $b_\ell$ denoting the activation function, weight matrix, and bias vector at layer $\ell$, respectively, and $L$ is the total number of hidden-layers. Here $f^{(\theta_i)}_{i,j}$ represents $j^{\text{th}}$ component of the embedding vector $f^{(\theta_i)}_i \in \mathbb{R}^{emb}$. Although there is no strict ordering required for partitioning the dimension set $\mathrm{D} = \mathrm{D}_1 \cup \mathrm{D}_2 \cup \mathrm{D}_3$, we adopt an ascending-order partitioning strategy for all experiments presented in this paper for the sake of consistency. The number of partitioned subsets is arbitrary and can be increased by proportionally increasing the number of body networks in the Anant-Net architecture. However, for brevity, we fix the total number of partitioned sets; and correspondingly, the number of body networks, to three in all experiments discussed herein. 
Algorithm \ref{Alg1} gives the details of training data and collocation points for the Anant-Net architecture. It is also important to note that modifying the number of body networks by changing the number of partitioned dimension subsets must be reflected in the data loader, as described in Algorithm \ref{Alg1}.

\begin{algorithm}[H]
\caption{Training Data and Collocation points loader for Anant-Net architecture.}
1: Consider a $d$-dimensional PDE where $ \mathrm{D} = \{1, 2, 3, \dots d\}$ is the set of all the dimensional indices

2: Partitioning the dimensions into $B = \{p, q, r\}$ sets (equal to the batch size), corresponding to the $B=3$ body networks.

3: $\{p, q, r\}$ are the stochastically sampled \textit{active} dimensions (the dimension across which the solution has a non-zero gradient) and $\mathrm{D}$\textbackslash $\{p, q, r\}$ are \textit{inactive} dimensions.

\Repeat{Total number of batches ($d/B$) are covered}{
     Sample $\{p, q, r\}$ randomly
     
     Yield $\mathcal{G}_{\{p, q, r\}}^{\text{data}}, ~\mathcal{G}_{\{p, q, r\}}^{\text{coll.}},~ u_{\{p, q, r\}}^{\text{data}},~  \forall \hspace{0.2cm} p,q,r \in \mathrm{D}$
     
     (Here 
     $\mathcal{G}_{data}^{pqr}$ represents grid of data points and $\mathcal{G}_{coll}^{pqr}$ represents grid of collocation points)
}
\label{Alg1}
\end{algorithm}

\noindent
For a general PDE of the form as shown below,
\begin{equation}\label{eq:PDEProb}
\mathcal{L}u = f \text{ in } \Omega^d \text{ and } \mathcal{B}u = g \text{ on } \partial\Omega^d,    
\end{equation}
where, 
$u$ denotes the solution of a $d$-dimensional PDE, which is the unknown, $\mathcal{L}$ and $\mathcal{B}$ are operators acting on the interior, $\Omega$ and boundary, $\partial\Omega$ respectively. $\mathcal{B}u = g$ is the boundary constraint. The above equation can be recast as an equivalent optimization problem with the following loss function:
\begin{equation}\label{eq:LossFun1}
\mathcal{J}(\theta) = \lambda_r\|\mathcal{L}\hat{u}_{\theta} - f\|_2^2 + \lambda_b\|\mathcal{B}\hat{u}_{\theta} - g\|_2^2,
\end{equation}
where, $\lambda_r$ and $\lambda_b$ are the weights on PDE residual loss and boundary loss respectively, which are suitably chosen to avoid any gradient imbalance between the two objectives while training the model for high dimensional PDEs; see \cite{wang2022and} and \cite{anagnostopoulos2024residual} for more details. $\hat{u}_{\theta}$ is the output of the Anant-Net that approximate the solution of a PDE. Algorithm \ref{Alg2} shows the details of Anant-Net training procedure. Appendix \ref{appendix:c} provides a detailed description of all the symbols used in this work.

\begin{algorithm}[H]
\caption{Anant-Net training procedure}
1: In general, for $B$-batch size, the prediction of the solution field ($\hat{u}_{\theta}$) via Anant-Net architecture is given by:
\begin{equation*}\label{AN1}
\hat{u}_{\theta}(X_1, X_2, \ldots, X_B) =  f^{(\theta_1)}_1(X_1) \otimes f^{(\theta_2)}_2(X_2) \otimes ~\cdots~ \otimes f^{(\theta_B)}_B(X_B).
\end{equation*}
For $B=3$, the active dimension indices $p, q, r$, corresponding to the batch size, are stochastically sampled such that \textbf{all dimensions are swept without repetition within a single epoch}.

\vspace{0.18cm}
 
\hrulefill
\vspace{0.18cm}
  
\Repeat{i $\leq$ Total number of batches ($d/B$)}{
    2: Sample $\mathcal{G}_{\{p, q, r\}}^{\text{data}}$, $\mathcal{G}_{\{p, q, r\}}^{\text{coll.}}$, and the corresponding dimension set by using Algorithm \ref{Alg1}

    3: Compute data loss ($\mathcal{J}_{\text{data}}$) :
    $$\mathcal{J}_{\text{data}}\left(\mathcal{G}_{\{p, q, r\}}^{\text{data}}, u^{\text{data}};\theta \right) =  \left\|\hat{u}_{\theta} \left(\mathcal{G}_{\{p, q, r\}}^{\text{data}} \right) - u^{\text{data}} \left(\mathcal{G}_{\{p, q, r\}}^{\text{data}}\right) \right\|_2^2.$$

    4: Compute PDE residual loss ($\mathcal{J}_{\text{Res}}$) on the sampled (active) dimensions $\{p, q, r\}$ within a hypercube:
    $$\mathcal{J}_{\text{Res}}\left(\mathcal{G}_{\{p, q, r\}}^{\text{coll.}}; \theta \right) = \biggl\| \sum_{k \in \{p, q, r\}} \mathcal{L}_k \hat{u}_{\theta}(x_1, x_2, \dots x_d) - f(x_1, x_2, \dots x_d)\biggr\|_2^2.$$
    Note that, since the PDE loss is computed only over the active dimensions, 
    all inactive dimensions do not contribute towards the gradient computation since they represent fixed locations in the computational domain.

    5: Compute total loss ($\mathcal{J}_{\text{Total}}$):
    $$\mathcal{J}_{\text{Total}} \left(\mathcal{G}_{\{p, q, r\}}^{\text{data}}, \mathcal{G}_{\{p, q, r\}}^{\text{coll.}}; \theta \right) = \lambda_b \cdot \mathcal{J}_{\text{data}}\left(\mathcal{G}_{\{p, q, r\}}^{\text{data}}, u^{\text{data}};\theta \right) + \lambda_r \cdot \mathcal{J}_{\text{Res}}\left(\mathcal{G}_{\{p, q, r\}}^{\text{coll.}}; \theta \right).$$

    6: Compute gradient of the total loss, update the parameters based on the average gradient using the suitable optimizer.
}
\label{Alg2}
\end{algorithm}


\subsection{Anant-Net’s Data Structures: Balancing Training Efficiency and Testing Flexibility}
For the Anant-Net architecture, it is important to highlight the non-trivial differences between the training and testing procedures. During training, the data structure follows a tensor-array format, as described in Algorithm \ref{Alg1}. In this format, only $B$ selected dimensions are active, meaning they exhibit active gradients, while the remaining $(d - B)$ dimensions remain fixed in an arbitrary $d$-dimensional problem. This tensor-based format offers enhanced memory efficiency, which helps prevent memory bottlenecks even in problems with extremely large dimensionality. Furthermore, the tensor-based training framework enables the data to be handled as hypercubes, fully leveraging the Anant-Net architecture described in Algorithm \ref{Alg2}.
However, using a tensor-array structure during the testing phase would render the model overly rigid, thereby limiting its applicability to more practical scenarios involving randomly selected dimensions. To address this, the testing phase employs a distinct approach, wherein the data structure adopts a vector-array format, consisting of randomly sampled testing points (and not a grid). This format allows unrestricted selection of dimensions from an arbitrary $d$-dimensional space during inference.

\subsection{Theoretical Analysis of Computational Complexity}
Here, we consider space complexity as the function of the number of trainable parameters in a given model architecture. For a fully-connected neural network of depth, $D$ (layers) and width, $W$ (neurons per layer), the space complexity of the model can be approximated as $\mathcal{O}(W^2D)$ assuming no bias terms. However, for a separable neural network (such as SPINN\cite{cho2024separable}) with $B$ body networks, the space complexity is given by $\mathcal{O}(B\tilde{W}^2\tilde{D})$ where $\tilde{D} < D \text{ and } \tilde{W} < W$. To ensure a fair comparison, we assume $\mathcal{O}(B\tilde{W}^2\tilde{D}) \approx \mathcal{O}(W^2D)$ thereby keeping the representation capacity approximately the same across both architectures. However, for such separable architecture, the space complexity becomes exceedingly large as the complexity scales linearly with the dimension, $d$ of the PDE since $B = d$ leading to inefficiencies for high-dimensional problems. In comparison, Anant-Net also follows the separable architecture paradigm, with space complexity $\mathcal{O}(B\tilde{W}^2\tilde{D})$, but importantly with $B << d$. As a result, the complexity of Anant-Net does not scale with dimensionality of the PDE, making it significantly more scalable.

We now analyze the time complexity of the proposed architecture. Let $N$ denote the number of collocation points sampled from a $d$-dimensional space for solving a $d$-dimensional PDE. Typically, $N$ is much larger than $d$. A standard fully connected neural network exhibits a time complexity of $\mathcal{O}(N^d)$ for such problems. In contrast, separable architectures, such as SPINN \cite{cho2024separable}, reduce the computational burden to $\mathcal{O}(Nd)$ by decomposing the multi-dimensional PDE into separable lower-dimensional components, thereby enabling more efficient training and inference.
However, a key limitation of separable formulations becomes evident in extremely high-dimensional settings, where the linear growth in complexity with respect to the dimension $d$ remains prohibitive. The Anant-Net architecture addressed this challenge by tacking the high-dimensions batch-wise,  achieving a time complexity of $ \mathcal{O}\left(NB\right)$, where  $B \ll d $ is a batch size.
Therefore, the choice of $B$ emerges as an important hyperparameter that governs the trade-off between computational efficiency and model expressivity. Selecting an small value of $B$ (smaller than $d$) allows Anant-Net to maintain favorable scalability properties, especially in high-dimensional regimes. Importantly, the selection of $B$ may also be influenced by hardware constraints and data availability, both of which play an important role in determining the architecture's practical effectiveness.

\subsection{Universal Approximation Theorem: Through the Lens of Anant\texorpdfstring{-}{-}Net}
\label{sec:UAT}
The Universal Approximation Theorem (UAT) provides a foundational guarantee for neural networks, stating that a feedforward neural network with a single hidden layer containing a sufficient number of neurons and a non-linear activation function can approximate any continuous function on a compact domain to arbitrary accuracy. This result highlights the theoretical expressivity of neural architectures and their applicability to a wide range of function approximation tasks, including the solution of PDEs.

In the context of scientific machine learning, UAT plays an important role in justifying the use of neural networks as surrogates or solution ans\"atze for PDEs. In this section, we analyze the universal approximation capabilities of the proposed Anant-Net architecture, specifically in the context of high-dimensional PDEs. Our goal is to illustrate, both theoretically and empirically, that Anant-Net retains sufficient expressivity to approximate complex solution manifolds that arise in such problems. To facilitate compact and efficient notation in our analysis, we adopt the Einstein summation convention in formulating the Anant-Net ansatz, rather than relying on explicit tensor product notation ($\otimes$).

\begin{theorem}
(Anant-Net) Let X, Y be compact subsets of $\mathbb{R}^d$. With $u \in L^2(X \times Y).$ Then, for arbitrary $\varepsilon > 0$ we can find a sufficiently large $r > 0$ (which is the size of the embedding layer for all body networks) and neural networks $f_j \triangleq f_{1,j}^{(\theta_1)}$ and $g_j \triangleq f_{2,j}^{(\theta_2)}$ such that,

\begin{equation}
\frac{1}{|\hat{\mathcal{B}}|}\sum_{k=1}^{|\hat{\mathcal{B}}|}\biggl\|u^k - \sum_{j=1}^{r}f_jg_j\biggr\|_{L^2(X \times Y)} < \varepsilon, 
\end{equation}
\begin{equation}
f_j = f_j(\hat{\mathcal{B}}_k^1,X \text{\textbackslash} \hat{\mathcal{B}_k^1)} \hspace{1cm}\text{and}\hspace{1cm} g_j = g_j(\hat{\mathcal{B}}_k^2, Y\text{\textbackslash}  \hat{\mathcal{B}}_k^2), 
\end{equation}
\\
where, $\hat{\mathcal{B}} = \{\hat{\mathcal{B}}_1,\hat{\mathcal{B}}_2, \dots \hat{\mathcal{B}}_k\}$ is the set of sets consisting of active dimensions where at any given step-$k$, $\hat{\mathcal{B}}_k = \{\hat{\mathcal{B}}_k^1, \hat{\mathcal{B}}_k^2\} = \{x_p, x_q\}$ is the set of active dimensions, $\{X $\textbackslash $ \hat{\mathcal{B}}_k^1\}$ and $\{Y $\textbackslash $ \hat{\mathcal{B}}_k^2\}$ is the set of inactive dimensions. Here, $X = \{x_1,x_2,\dots x_p \dots x_{d/2}\}$, $Y = \{x_{d/2+1},\dots x_q \dots x_d\}$. 
\end{theorem}
The proof is given in APPENDIX \ref{appendix:a}.
 
\subsection{Effect of Preconditioning on Training Dynamics}
In this section, we investigate the role of preconditioning in enhancing the training of the Anant-Net for high-dimensional PDEs. Ryck et al. \cite{de2023operator} established that linearly transforming the model parameters is effectively equivalent to preconditioning the loss gradient via multiplication with a positive definite matrix. Their findings demonstrated the utility of such preconditioning in accelerating convergence for a range of low-dimensional PDEs (typically in one or two spatial dimensions). Building on this foundation, the present study extends the theoretical framework to high-dimensional settings and demonstrates the effectiveness of gradient preconditioning in improving training performance when employing the Anant-Net architecture. It is important to acknowledge, however, that these benefits are accompanied by an increase in computational cost due to the cost associated with constructing and applying the preconditioning matrix. Consider a \textit{linear} Anant-Net architecture with 3 body networks with the \textit{ansatz} bearing the following form given by equation \eqref{AN},
\begin{equation*}
\hat{u}_{\theta}(X_1, X_2, X_3) =  f_1^{(\theta_1)}(X_1) \otimes f_2^{(\theta_2)}(X_2) \otimes f_3^{(\theta_3)}(X_3),
\end{equation*}
\\
where  $f_k^{(\theta_k)}(X_k) = \sum_{i=1}^{n_k} w_{2,k}^i(w_{1,k}X_k + b_1) = \sum_{i=1}^{n_k} w_{2,k}^i\phi_i(X_k), ~ k = 1,2,3.$ The parameter $ w^i_{j,k}$ denotes the weight associated with the $i^{\text{th}} $ neuron in the $ j^{\text{th}}$ layer of the $k^{\text{th}} $ body network.
In the above formulation, the basis functions are implicitly represented by the hidden layers of the body networks within the Anant-Net architecture, while the corresponding coefficients are extracted from the final output layer. This representation is analogous to the framework proposed by Cyr et al. \cite{cyr2020robust}, wherein a generic deep neural network is interpreted as a linear combination of learned basis functions. We consider the continuous formulation of the loss function, which serves as the analog of the discrete loss representation given in Equation \eqref{eq:LossFun1}, corresponding to the same PDE defined in Equation  \eqref{eq:PDEProb}.
\begin{equation}\label{eq:LossFun2}
    \mathcal{J}(\theta) = \frac{1}{2}\int_{\Omega} |\mathcal{L}\hat{u}_{\theta}(x) - f(x)|^2dx + \frac{\lambda}{2}\int_{\partial\Omega}|\hat{u}_{\theta}(x) - g(x)|^2 d\Omega,
\end{equation}
where, $x\in X \subset \mathbb{R}^d$.
In this formulation, the Anant-Net architecture is approximated by a linear model (without any nonlinear activation functions), thereby yielding the linear representation presented above. For clarity and brevity, we restrict our analysis to the Anant-Net comprising three body networks, each containing two hidden layers.  The resulting linear model is trained using standard gradient descent optimization. In its simplest form, the gradient descent update rule is given by  $\theta_{r+1} = \theta_r - \eta \nabla L(\theta_r),$ where $ \theta_r $ denotes the set of parameters at the  $r^{\text{th}}$ iteration, and $\eta$ is the learning rate.

Given the inherent non-convexity of the loss function, an important aspect that makes analytical investigation particularly challenging, De Ryck et al. \cite{de2023operator} propose an alternative formulation to simplify the analysis of gradient descent dynamics. Specifically, by fixing the parameters at a given iteration $r$, the Taylor series expansion of the network ansatz about the initialization point $\theta_0$ takes the following form:
$$\hat{u}(x; \theta_r) = \hat{u}(x;\theta_0) + \nabla_{\theta}\hat{u}(x;\theta_0)^\mathsf{T}(\theta_r - \theta_0) + \frac{1}{2}(\theta_r - \theta_0)^{\mathsf{T}}H_r(\theta_r - \theta_0),$$
where $H_r$ is the hessian computed at intermediate parameters between $0$ and $r$. Here, for brevity a new notation $\phi_i(x) \triangleq \partial_{\theta_i}u(x;\theta_0)$ is introduced with $\mathcal{L}\phi_i \in L^2(\Omega)$. The above formulation is substituted in Equation \ref{eq:LossFun2} that leads to the following simplification:
\begin{equation}
\theta_{r+1} = \theta_r - \eta\nabla L(\theta_r) = (I - \eta\mathbb{A})\theta_r + \eta(\mathbb{A}\theta_0 + \mathbb{B}) + \eta \epsilon_r.
\end{equation}
Here, $ \epsilon_r \in \mathbb{R}^n $ encapsulates the higher-order terms neglected in the Taylor series expansion, thereby isolating the leading-order contributions retained in the resulting expression for the \textit{simplified} gradient descent formulation.
$$\mathbb{A}_{i,j} = \langle \mathcal{L}\phi_i , \mathcal{L}\phi_j \rangle_{L^2{(\Omega)}} + \lambda\langle \phi_i , \phi_j \rangle_{L^2{(\Omega)}}  \hspace{1cm} \mathbb{A} \in \mathbb{R}^{n\times n},$$
$$ \mathbb{B}_i = \langle f - \mathcal{L}u_{\theta_0}, \mathcal{L}\phi_i \rangle_{L^2{(\Omega)}} + \lambda\langle u - u_{\theta_0}, \phi_i\rangle_{L^2{(\Omega)}} \hspace{1cm} \mathbb{B} \in \mathbb{R}^n.$$
In the case where the matrix $ \mathbb{A} $ is invertible, let $ \kappa(\mathbb{A}) $ denote its condition number. To achieve a specified error tolerance $\epsilon$,  the number of gradient descent iterations required satisfies the relation $N(\epsilon) \propto 1/\ln(1 - c/\kappa(\mathbb{A})) $ for some constant $0 < c < 1$. Consequently, as $ \mathbb{A}$ becomes increasingly ill-conditioned, \textit{i.e.}, $\kappa(\mathbb{A}) \to \infty$, the number of iterations $ N(\epsilon)$ diverges, indicating a deterioration in convergence rate. To address this challenge in the training of physics-informed machine learning models, we employ an explicit preconditioning strategy for the gradients, as detailed below.
\begin{equation}\label{eg:PrecondHess}
\widehat{\theta}_{r+1} = \widehat{\theta}_r - \eta \mathbb{PP}^\mathsf{T}\nabla L(\widehat{\theta}_r),  
\end{equation}
where $\widehat{\theta}_r := \mathbb{P} \theta_r $ and $\mathbb{P}\mathbb{P}^\mathsf{T} $ is a symmetric positive definite matrix. It follows from the preceding discussion that such a linear transformation of the parameters is mathematically equivalent to preconditioning the gradient. In this work, we adopt a preconditioning strategy grounded in (quasi-)Newton optimization methods, wherein the gradients are modified using curvature information derived from the loss landscape. Specifically, (quasi-)Newton approaches leverage both the gradient and an approximation of the Hessian of the loss function to accelerate convergence during optimization.
\begin{equation}\label{eg:Hess}
\widehat{\theta}_{r+1} = \widehat{\theta}_r - \eta H^{-1}\nabla L(\widehat{\theta}_r).  
\end{equation}

\begin{align*}
    \text{Hessian:}\hspace{0.5cm}H &= \partial_{\theta_i}\partial_{\theta_j} \mathcal{J}(\theta) \nonumber \\
    & = \int_{\Omega} (\mathcal{L}\partial_{\theta_i}u_{\theta}(x))\cdot \mathcal{L}\partial_{\theta_j}u_{\theta}(x) dx + \lambda \int_{\partial\Omega}\partial_{\theta_i}u_{\theta}(x)\cdot \partial_{\theta_j}u_{\theta}(x) dx \nonumber \\
    & = \int_{\Omega} (\mathcal{L}\phi_i(x)\cdot \mathcal{L}\phi_j(x) dx + \lambda \int_{\partial\Omega}\phi_i(x)\cdot \phi_j(x) dx = \langle \mathcal{L}\phi_i , \mathcal{L}\phi_j \rangle_{L^2{(\Omega)}} + \lambda\langle \phi_i , \phi_j \rangle_{L^2{(\Omega)}} \nonumber \\
    & = \mathbb{A}_{i,j}    
\end{align*}
By comparing Equation \eqref{eg:PrecondHess} and Equation \eqref{eg:Hess} it can be concluded that using a (quasi-)Newton method  implicitly preconditions the gradient since Hessian of the loss function is equivalent to $\mathbb{A}$.

\begin{figure}[ht]
\centering
{
\centering
\includegraphics[trim = 0 1cm 0 0, scale=0.45]{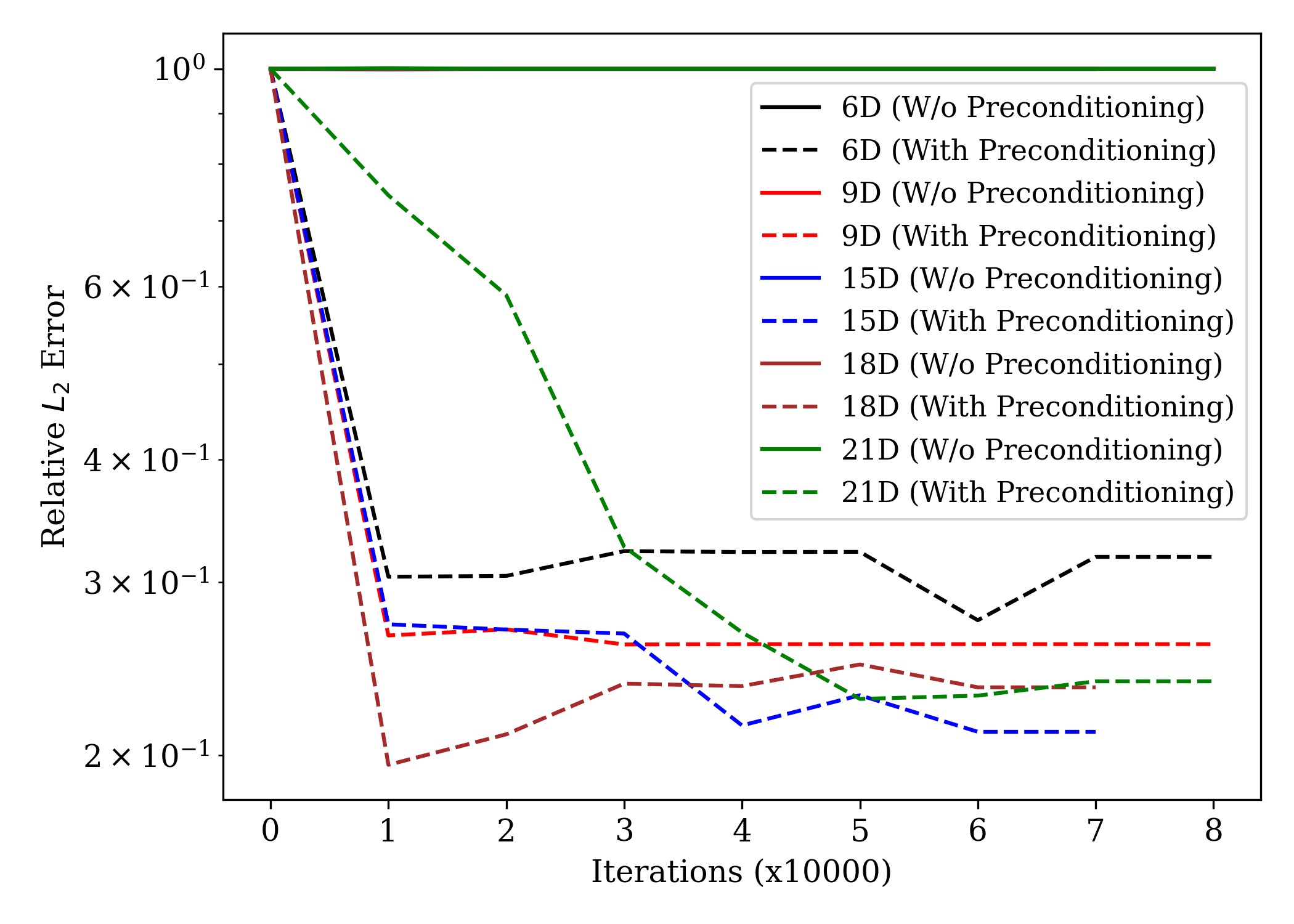}
}
\hspace{0.25cm}
{
\centering
\includegraphics[trim = 0 1cm 0 0, scale=0.45]{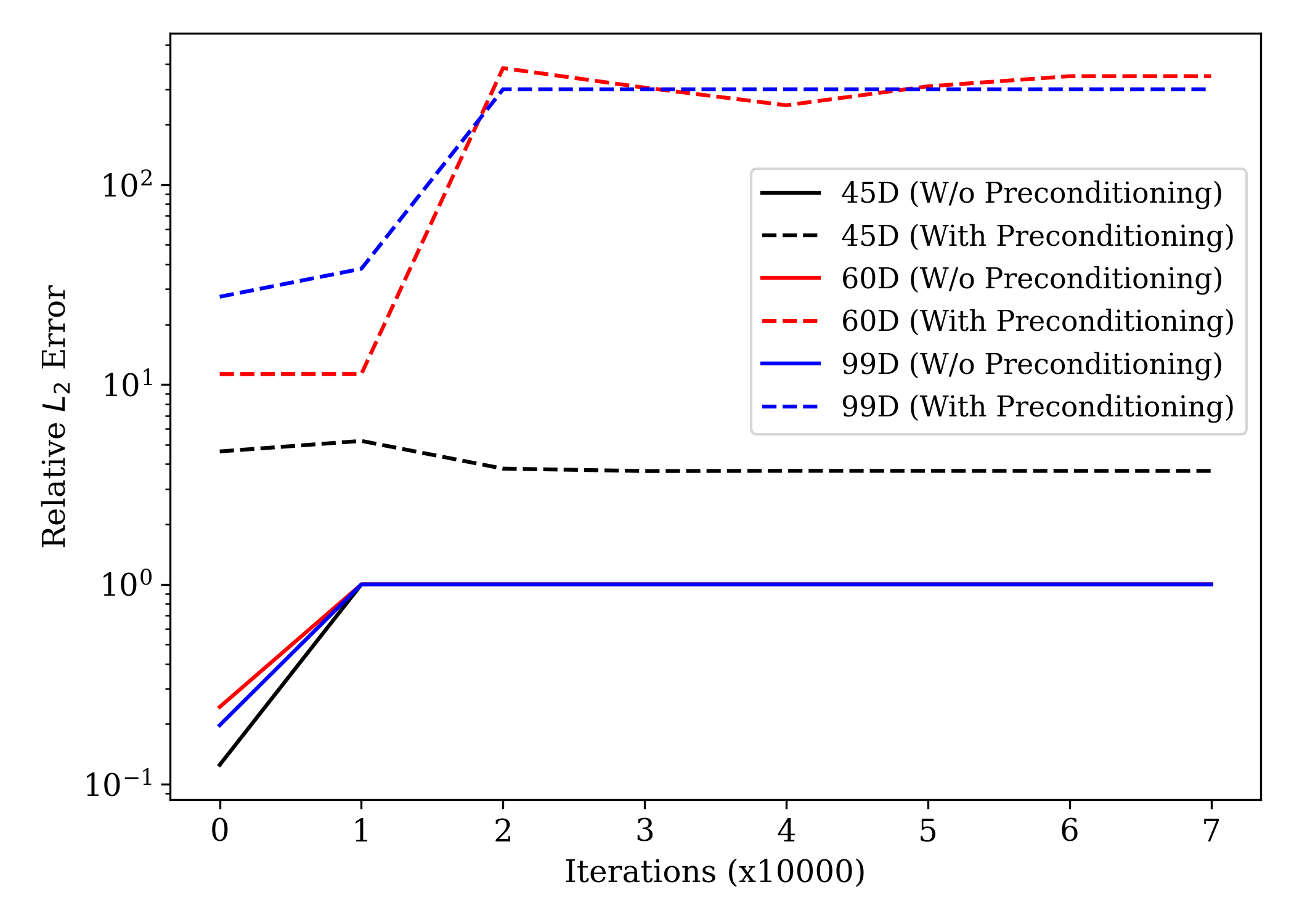}
}
\caption{Effect of Preconditioning on High-Dimensional Poisson Equation Training: Training results using (quasi-)Newton optimization for 6D, 9D, 15D, 18D, and 21D show overlapping performance without preconditioning. For 45D and higher (right), preconditioning offers no improvement, due to the linear architecture. Note that these experiments were performed on a T4 GPU with 15 GB RAM.}
\label{fig:precond}
\end{figure}
\noindent
While preconditioning has demonstrated clear advantages in accelerating convergence within the framework of physics-informed deep learning, prior investigations have been primarily constrained to low-dimensional settings (typically $\leq$ 3 dimensions). In this work, we extend these findings to higher-dimensional PDEs by leveraging the proposed Anant-Net architecture. In particular, we employ a \textit{linear} version of Anant-Net to validate theoretical predictions through empirical experiments that align closely with the theoretical analysis presented in this section. The corresponding results are illustrated in Figure \ref{fig:precond}. For clarity, we define \textit{high-dimensional} as dimensionalities greater than 3 and up to 21, and \textit{extremely high-dimensional (X-high)} as dimensionalities exceeding 21, as referenced in the figure. Within the high-dimensional regime, we observe that the theoretical benefits of preconditioning are preserved under the linear Anant-Net architecture. Specifically, preconditioning the gradients via Newton-based optimization yields stable training dynamics and favorable convergence behavior; see Figure \ref{fig:precond} (left) . In contrast, omitting preconditioning causes the model to converge to suboptimal local minima, with negligible performance gains throughout training. However, in the X-high dimensional regime, the linear Anant-Net architecture fails to show any tangible benefit from preconditioning, as evidenced in Figure \ref{fig:precond} (right). This breakdown of theoretical alignment in extremely high dimensions can be attributed to the inherent limitations of the linear two-layer network, which appears insufficient for capturing the complexity of the solution landscape in such settings. Furthermore, increasing the network's depth or width does not overcome this limitation in the absence of nonlinearities. Nonetheless, as we elaborate in the Results section, these limitations can be effectively addressed by incorporating nonlinear activation functions, thereby enabling the resolution of high-dimensional PDEs. From a practical standpoint, to avoid the computational burden typically associated with second-order optimizers, we implement gradient preconditioning via (quasi-)Newton methods as a secondary optimization step. This step follows an initial training phase using a standard stochastic gradient-based optimizer.

\subsection{Anant-Net with Interpretable KAN}
\begin{figure}[h]
\centering
{
\includegraphics[trim = {4cm 1.5cm 4cm 1.5cm}, scale=0.3]{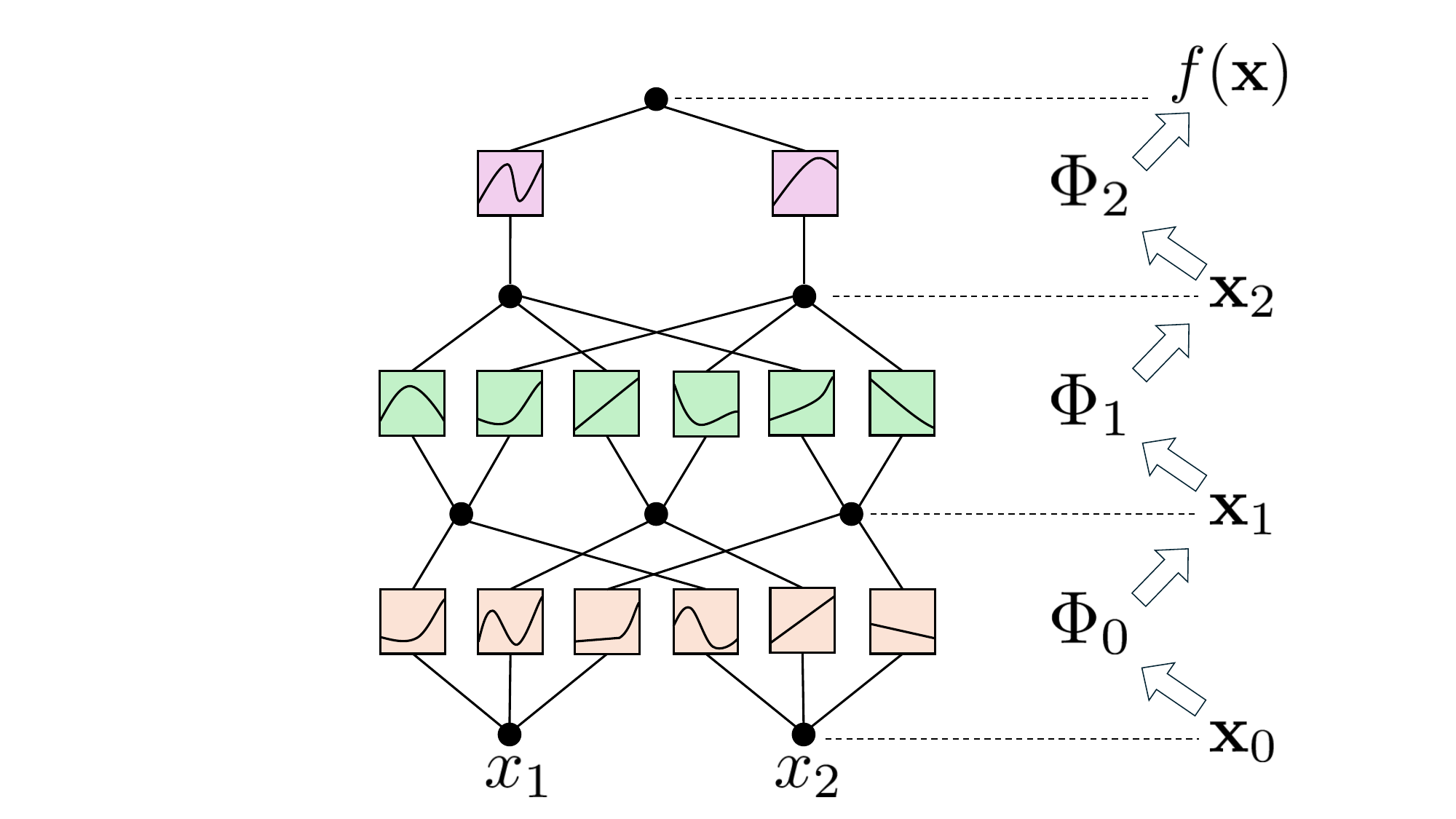}
}
\caption{A schematic of a 3-layer Kolmogorov–Arnold Network (KAN) is shown, with input $\mathbf{x}_0 = \{x_1, x_2\}$ and output $f(\mathbf{x}) = (\Phi_2 \circ \Phi_1 \circ \Phi_0)(\mathbf{x}_0)$. Replacing the feedforward neural networks (body networks) in the Anant-Net architecture (as shown in Figure \ref{fig:AnantN}) with KAN-based body networks yields the Anant-KAN architecture.}
\label{fig:KAN}
\end{figure}

\noindent
Kolmogorov–Arnold neural networks (KANs), recently introduced by Liu et al. \cite{liu2024kan}, extend the classical Kolmogorov–Arnold representation theorem, which was originally formulated for shallow networks with limited hidden neurons, to modern deep learning architectures. By integrating backpropagation-based training, the KAN framework generalizes naturally to networks with an arbitrary number of layers. In this formulation, a KAN with $L$ layers is expressed as a composition of $L$ basis functions (see Fig. \ref{fig:KAN}):
$$
f(\mathbf{x}) = (\Phi_{L-1} \circ \Phi_{L-2} \circ \dots \circ \Phi_1 \circ \Phi_0)(\mathbf{x}) ,   
$$
where $f(\mathbf{x})$ is a multivariate continuous function with $\mathbf{x}\in\mathbb{R}^d$ and functionality of basis function $\Phi$ in each layer can be defined as follows:
$$
\mathbf{x}_{l+1} = \Phi_l \mathbf{x}_{l},
$$
where $\mathbf{x}_l \in \mathbb{R}^{n_l}$ and $\mathbf{x}_{l+1} \in \mathbb{R}^{n_{l+1}}$ denote the input and output of the $l$-th layer, respectively, in a KAN architecture with an arbitrary number of $L$ layers, and $n_l$ represents number of neurons in the $l $-th layer. The transformation $\Phi_l $ associated with the $l $-th layer can then be expressed in matrix form as follows:

$$
\Phi_l = \begin{bmatrix}
\phi_{l,1,1}(\cdot) & \phi_{l,1,2}(\cdot) & \dots & \phi_{l,1,n_l}(\cdot)\\
\phi_{l,2,1}(\cdot) & \phi_{l,2,2}(\cdot) & \dots & \phi_{l,2,n_l}(\cdot)\\
\vdots & \vdots & \ddots & \vdots\\
\phi_{l,n_{l+1},1}(\cdot) & \phi_{l,n_{l+1},2}(\cdot) & \dots & \phi_{l,n_{l+1},n_l}(\cdot)\\
\end{bmatrix}.
$$

\noindent
In both the original formulation \cite{liu2024kan} of the Kolmogorov–Arnold representation theorem and its recent extension (\cite{10763509}, \cite{Sidharth2024ChebyshevPK}, \cite{shukla2024comprehensive}), KANs were constructed using spline basis functions equipped with internal parameterizations. Two key hyperparameters govern the spline basis in this context: the polynomial order $k$ and the grid size $G $. These parameters play a central role in the performance of KANs. Notably, Theorem 2.1 in Liu et al.  \cite{ liu2024kan} establishes that the approximation error for any target function depends solely on $G $ and $k $, and remains independent of the input dimension $d $.  More recently, several alternative basis functions have been proposed to enhance KANs, particularly for applications involving the solution of PDEs. In this work, we examine some of the most effective basis functions through a series of experiments. To the best of our knowledge, this study presents the first application of KANs to the solution of high-dimensional PDEs.
Therefore, the efficiency of KANs in function approximation is fundamentally determined by the choice of basis functions $\Phi $. In this work, we investigate a range of basis formulations for KANs and systematically evaluate their performance in approximating the solution spaces of high-dimensional PDEs, using the Anant-Net architecture as a representative framework. KANs are particularly useful for their inherent interpretability, an attribute typically absent in standard multilayer perceptrons (MLPs). They are capable of learning both compositional structures and univariate functional mappings, often resulting in superior performance compared to MLPs. Furthermore, KANs have been shown to exhibit enhanced parametric efficiency, achieving comparable or improved predictive accuracy with significantly fewer trainable parameters than their MLP counterparts. Despite these advantages, several practical challenges remain unresolved. In particular, the selection of critical hyperparameters, such as grid size and spline order, is problem-dependent and often non-trivial, necessitating further investigation to enable more robust and automated deployments of KAN architectures.
\\

Recently, KANs have been employed for solving PDEs. Shukla et al. \cite{shukla2024comprehensive} introduced the PIKAN and DeepOKAN architectures, while Wang et al. \cite{wang2024kolmogorov} proposed Kolmogorov–Arnold Informed Neural Networks, which replace conventional MLP-based frameworks with KAN-based representations. Their study demonstrated that KANs, when coupled with lower-order orthogonal polynomials instead of the originally proposed spline-based basis functions, exhibit superior performance across a range of problems. Despite this success, some models using KAN with orthogonal polynomials experienced instability during training, though the cause of these divergences remains unclear. Furthermore, their work was primarily limited to low-dimensional PDEs.  In a similar vein, PIKAN was extended to a separable framework, leading to the development of SPIKAN by Jacob et al. \cite{jacob2024spikans}. The advantages of this separable framework were demonstrated on specific problems, such as the 2D Helmholtz equation, with the maximum dimensionality considered in their work being $\leq 3$. It is important to note that KAN has the potential to address the curse of dimensionality (see Theorem 2.1 in Liu et al. \cite{liu2024kan}), which asserts that the approximation error of KAN for a $d$-dimensional function $f$ is independent of $d$. We aim to investigate the immediate benefits of this property by extending Anant-Net to solve high-dimensional PDEs. To the best of our knowledge, the advantages of KAN for high-dimensional PDEs have not yet been fully explored, and we seek to investigate these benefits through the Anant-KAN (Anant-Net with KAN) architecture.

\subsubsection{Computational Complexity for Anant-KAN}
In this context, we consider space complexity as a function of the number of trainable parameters in a given model architecture for KANs. For a KAN with depth $D$ (number of layers) and width $W$  (number of nodes per layer), the space complexity of the model can be approximated as $\mathcal{O}(W^2 D (G + k))$, which simplifies to $ \mathcal{O}(W^2 D G)$ for a spline basis with grid size $G$ and spline order $k$. However, for a separable KAN (such as SPIKAN \cite{jacob2024spikans}) with $B $ body networks, the space complexity is given by $\mathcal{O}(B \tilde{W}^2 \tilde{D} G) $, where $\tilde{D} < D $ and $ \tilde{W} < W $. To ensure a fair comparison, we assume that $ \mathcal{O}(B \tilde{W}^2 \tilde{D} G) \approx \mathcal{O}(W^2 D G) $, thereby maintaining approximately the same representation capacity across both architectures. However, in such a separable architecture, the space complexity scales linearly with the dimensionality $d $ of the PDE, since $B = d $, leading to inefficiencies for high-dimensional problems. In contrast, Anant-KAN also follows the separable architecture paradigm, with space complexity $\mathcal{O}(B \tilde{W}^2 \tilde{D} G) $ using a spline basis with grid size $G $, but crucially with $B \ll d $. As a result, the complexity of Anant-KAN does not scale with the dimensionality $d $ of the PDE, making it significantly more scalable. It is important to note that the computational time complexity of Anant-KAN is similar to that of Anant-Net, which is $\mathcal{O}(NB) $, with $B \ll d $ for solving high-dimensional PDEs.

\section{Results}
This section presents a series of results demonstrating the capability of Anant-Net in solving high-dimensional PDEs, both linear and nonlinear, as well as steady-state and transient, defined over hypercubic domains with side length  $ L > 1$. Specifically, we focus on the Poisson equation, sine-Gordon equation, and Allen-Cahn equation. A key challenge in such settings is the enforcement of boundary conditions, where boundary sampling significantly contributes to the complexity of the problem as the dimensionality increases. To evaluate the performance of the proposed architecture, we use the percentage relative $L_2$ error, defined as:
$$(\%) \text{ Relative } L_2 \text{ Error } = \frac{\big\| u_{\text{pred}} - u_{\text{exact}}\big\|_2}{\big\|u_{\text{exact}}\big\|_2} \times100.$$
Uncertainty quantification is performed by reporting the test error over 10 independent random realizations, with data points sampled from the high-dimensional domain. Unless otherwise specified, all experiments are executed on a T4 GPU with 15 GB RAM. To this end, we systematically compare the performance of Anant-Net with other state-of-the-art methods that also employ boundary sampling strategies, rather than enforcing boundary conditions directly. The weights for the individual loss terms is fixed for all the cases: $\lambda_{r} = 1$ and $\lambda_{b} = 15$ in Equation \ref{eq:LossFun1}. We use Xavier initialization to initialize the trainable parameters of the model for all the cases. For training the model, we use both AdamW and L-BFGS optimizers with learning rates, $1e-03$ and $1e-02$ respectively. 
 
In addition to the architectural hyperparameters in deep neural networks, several algorithmic hyperparameters must be tuned to achieve optimal performance with the Anant-Net architecture. These include, the number of body networks ($B$), grid resolution for collocation points ($N_C$), grid resolution for boundary/data points ($N_B$), number of collocation grids ($\#CG$), number of boundary/data grids ($\#BG$), and the sampling frequency ($S_{\text{FREQ}}$). In a way, the total number of iterations for training the model is approximately equal to $S_{\text{FREQ}} \times \#CG$ and the training is stopped when the maximum number of iterations is reached. The number of body networks ($B$) directly influences the architecture of Anant-Net. For example, when $B = 3$, the architecture consists of three body networks that process 3D cubes sampled from the high-dimensional space. Similarly, when $B = 4$, four body networks are used to process 4D tesseracts from the high-dimensional domain. The sampling frequency ($S_{\text{FREQ}}$) directly affects the training process by determining how frequently a new set of dimensions is sampled during training. If $S_{\text{FREQ}}$ is too low, the model spends less time interacting with the newly sampled subspace, leading to poor generalization. Conversely, if $S_{\text{FREQ}}$ is too high, the model may overfit due to prolonged exposure to the same subspace. Finally, the grid resolution for collocation points ($N_C$), grid resolution for boundary/data points ($N_B$), number of collocation grids ($\#CG$), and number of boundary/data grids ($\#BG$) determine the size and structure of the dataset handled by Anant-Net. We will employe $B=3$ for all computational experiments unless stated otherwise.

\begin{figure}[ht]
\centering
{
\includegraphics[trim = 0cm 1.5cm 0cm 2cm, clip, scale=0.5]{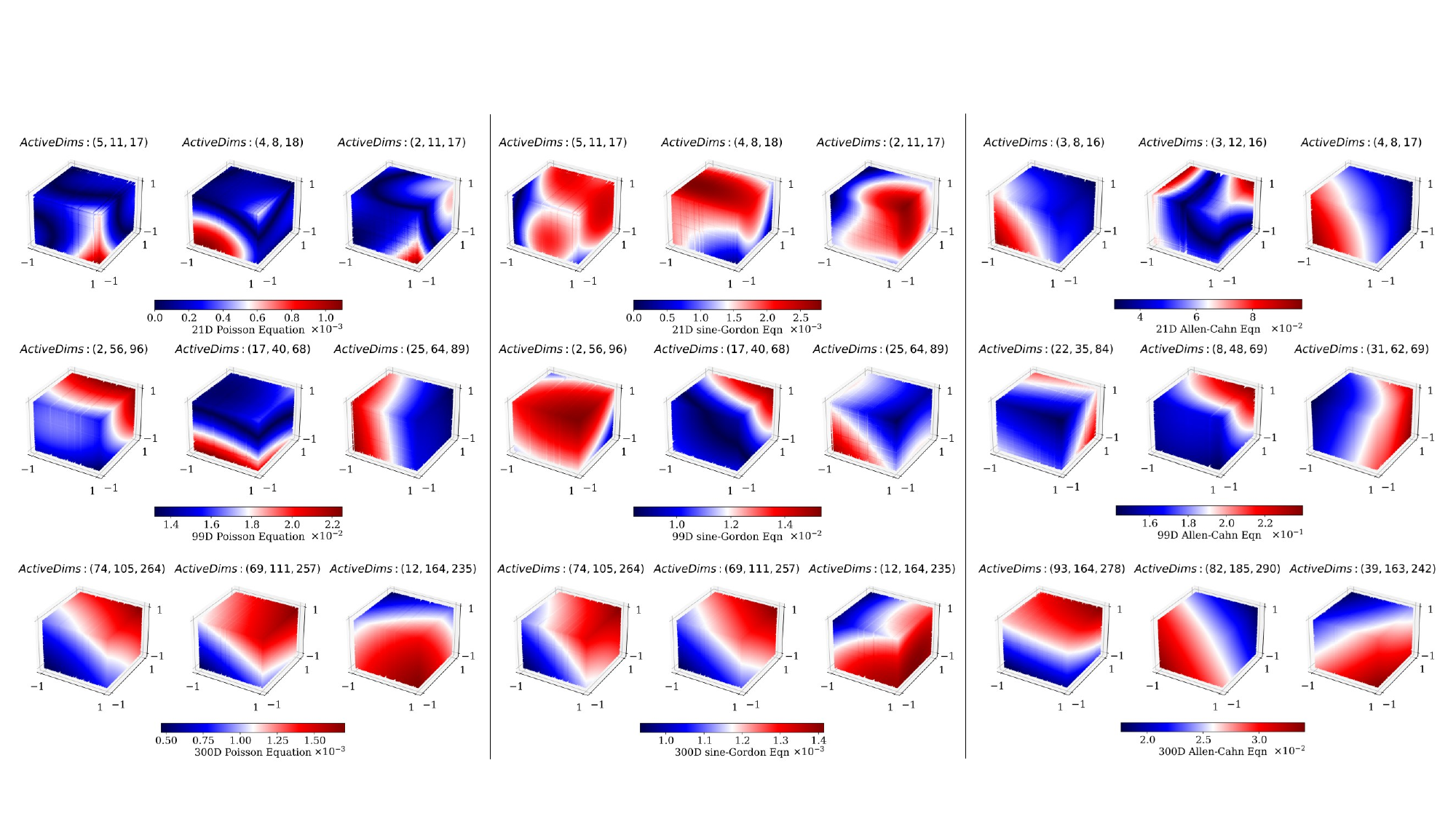}
}
\caption{Point-wise absolute testing error on randomly selected 3D cubes from 21D, 99D, and 300D cases for the (\textit{left}) Poisson equation, (\textit{center}) sine-Gordon equation, and (\textit{right}) Allen–Cahn equation. Here, "$\mathsf{ActiveDims}$: $\{p, q, r\}$" denotes the active dimensions in the $d$-dimensional space where gradients are non-zero, and $\mathrm{D}$\textbackslash$\{p, q, r\}$ represents the inactive dimensions. $\mathrm{D} = \{1, 2, \dots, d\}$ denotes the set of all dimensions.} 
\label{fig:absoluteerrorEE}
\end{figure}

\subsection{High-dimensional Poisson's Equation}
\label{sec:Poisson}
The high-dimensional Poisson’s equation is a fundamental PDE that arises in various scientific and engineering disciplines, particularly in modeling steady-state processes such as electrostatics, heat conduction, and gravitational potentials in complex systems. In high-dimensional settings, such as those encountered in quantum mechanics and financial mathematics, the computational complexity of solving the equation increases exponentially. 
Here, we consider Poisson's equation of the following form:
\begin{equation}
\label{eq:Poisson}
   -\Delta u(x) = f(x), x\in\Omega, 
\end{equation}
\begin{equation}
\label{eq:PoissonBoundary}
    u(x) = g(x), x \in \partial\Omega,
\end{equation}
\noindent
where $\Omega \in [-1, 1]^d$ and the exact solution is assumed to be of the form:

$$u_{\text{exact}}(x) = \biggl(\frac{1}{d}\sum_{i=1}^d x_i\biggl)^2 + \sin\biggl(\frac{1}{d}\sum_{i=1}^d x_i\biggl),$$
and
$$f(x) = \frac{1}{d}\biggl(\sin\biggl(\frac{1}{d}\sum_{i=1}^d x_i\biggl) - 2\biggl).$$

\noindent

\begin{table}[H]
\begin{center}
\begin{tabular}{||c | c | c | c | c||}  
 \hline
 Dimension & Architecture & Optimizer & $\#$Collocation & $\#$Data \\ [0.05ex] 
 \hline\hline
 21D & $\bigl[7, 64, 64, 10\bigr] \times 3$ & Adam+L-BFGS & 38416 & 6912\\ 
 \hline
 99D & $\bigl[33, 64, 64, 40\bigr] \times 3$ & Adam+L-BFGS & 90552 & 10800\\
 \hline
 300D & $\bigl[100, 128, 128, 128\bigr] \times 3$ & Adam+L-BFGS & 274400 & 32400\\
 \hline
\end{tabular}
\end{center}
\caption{Anant-Net architecture used for solving high-dimensional Poisson equation}
\label{tab:arch-poisson}
\end{table}

\begin{figure}[H]
\centering
{\label{fig:capparatus}
\centering
\includegraphics[width=0.3\linewidth]{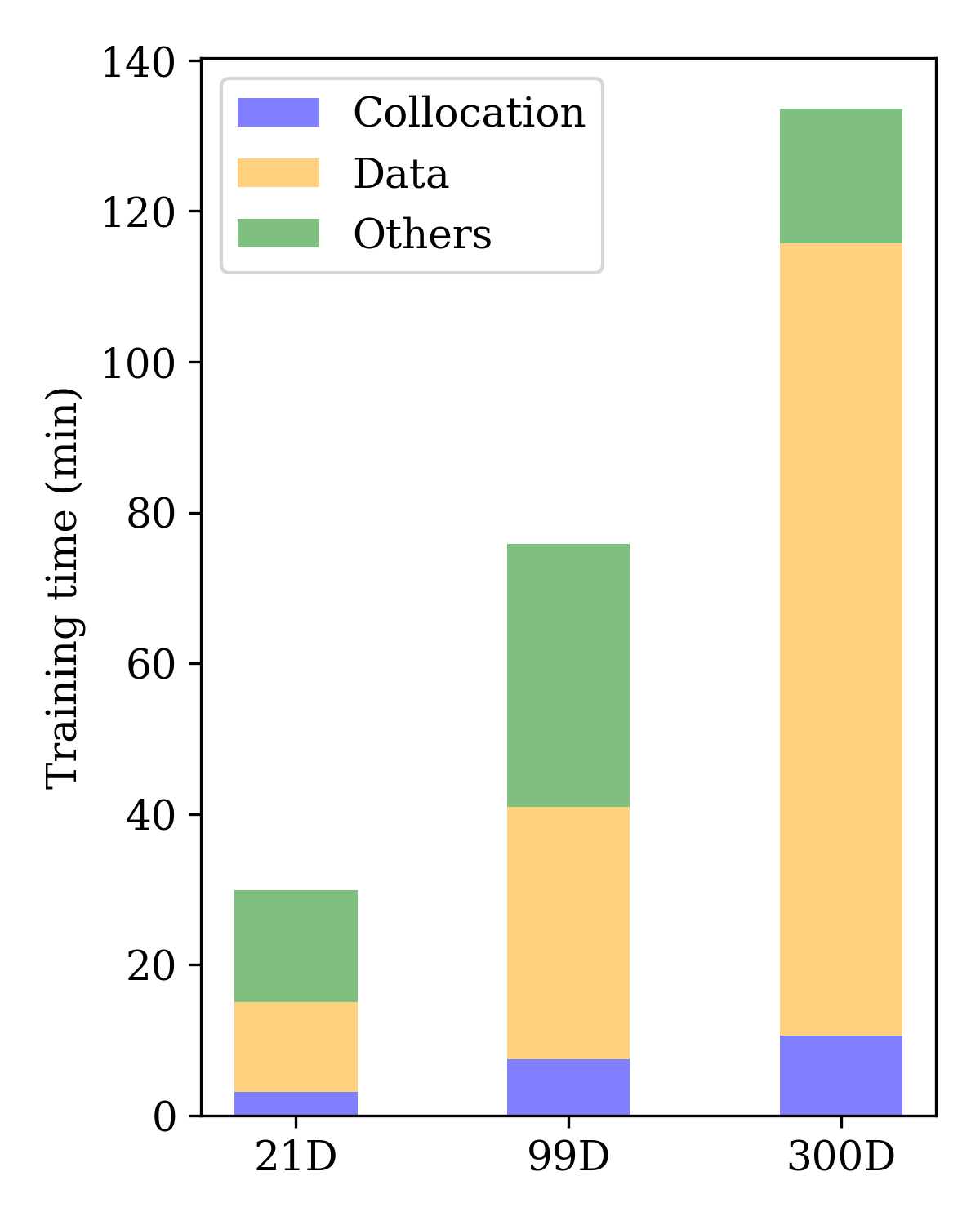}
}
\hspace{0.5cm}
\centering
{\label{fig:cdiagram}
\centering
\includegraphics[width=0.3\linewidth]{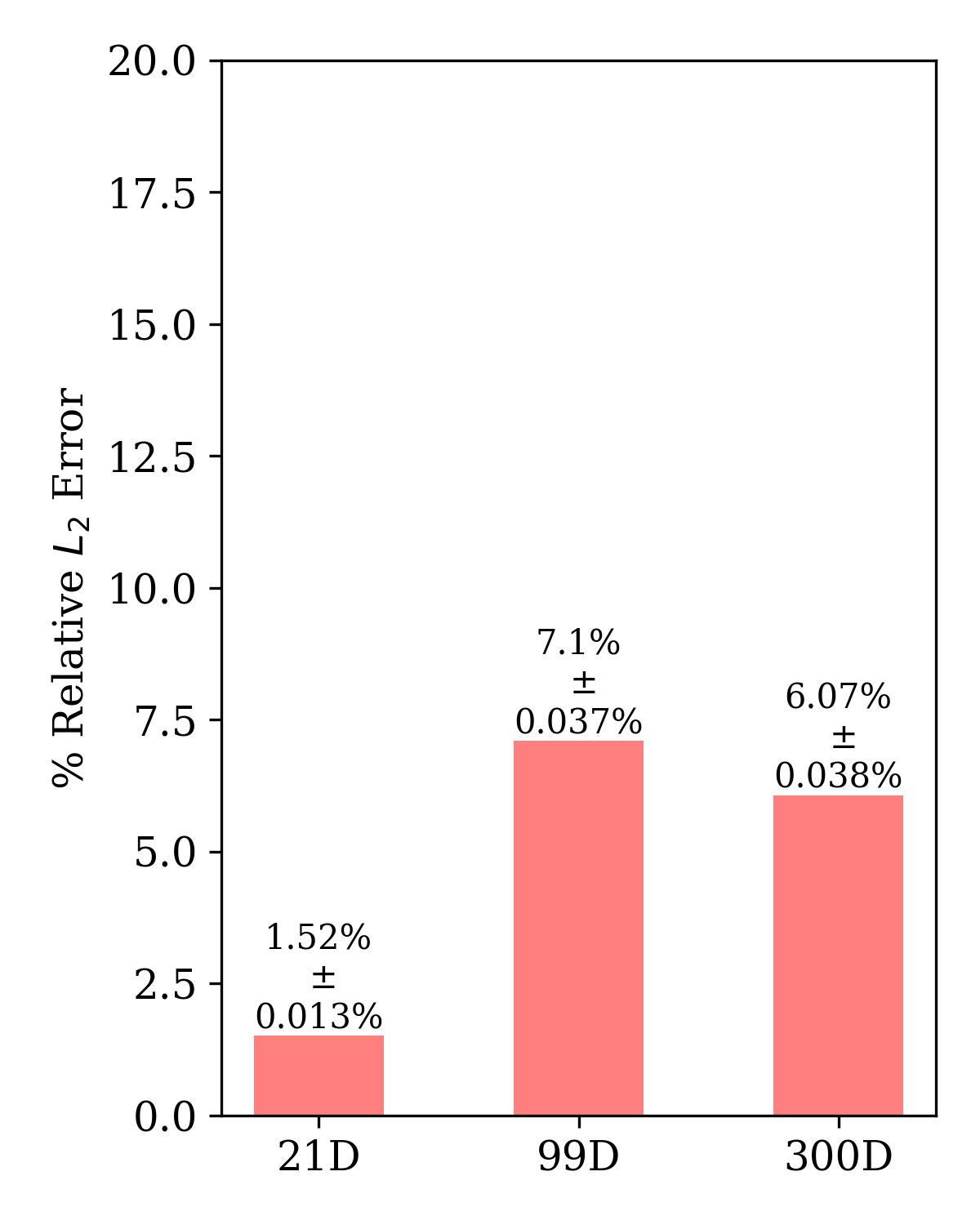}
}
\caption{Poisson Equation: (Left) Training time profiling; (Right) mean testing error with standard deviation, computed over 10 random realizations (seeds), for the 21D, 99D, and 300D cases.}
\label{fig:Poisson_AnantNet_Acc_Rtime}
\end{figure}

\noindent
In this section, we solve the high-dimensional Poisson’s equation in 21D, 99D, and 300D. For all cases, we use the Adam optimizer to train Anant-Net, followed by L-BFGS to further fine-tune the learning process. The benefits of preconditioning the gradients were discussed in previous sections, where we empirically validated the theory using (quasi-)Newton optimization with a linear version of Anant-Net. However, in this section, Anant-Net incorporates a non-linear activation function ($\tanh$) to solve the high-dimensional Poisson’s equation. Table \ref{tab:arch-poisson} provides further details on the specific hyperparameters used. As suggested by the curse of dimensionality, the number of collocation and data samples required to solve high-dimensional PDEs on a hypercube grows exponentially with dimension. This poses a major challenge for existing methods, including state-of-the-art PINNs such as NTK-based PINNs \cite{wang2022and}, self-adaptive PINNs \cite{mcclenny2023self}, etc., which struggle with scalability due to the computational burden of large sample volumes. Anant-Net addresses this issue efficiently. Table \ref{tab:arch-poisson} summarizes the total number of collocation and boundary points used for solving Poisson’s equation in 21D, 99D, and 300D. The sample sizes are controlled via hyperparameters: $N_C$, $\#CG$ for collocation points, and $N_B$, $\#BG$ for boundary points. Figure \ref{fig:Poisson_AnantNet_Acc_Rtime} (left) indicates that boundary data sampling is the primary contributor to training time across all dimensions, highlighting a need to optimize this component. Meanwhile, Figure \ref{fig:Poisson_AnantNet_Acc_Rtime} (right) shows a mean relative $L_2$ testing error along with the standard deviation, evaluated on randomly selected test points from the high-dimensional space. Figure \ref{fig:absoluteerrorEE} (left) shows the absolute point-wise error on randomly selected 3D cubes from the 21D, 99D, and 300D cases. More details on the hyperparameters used for training is explained in APPENDIX \ref{appendix:b}.

\subsection{High-dimensional sine-Gordon Equation}
\label{sec:Sine-Gordon}
The high-dimensional sine-Gordon equation is a nonlinear PDE  that generalizes the classical sine-Gordon model to multiple spatial dimensions. It serves as a canonical model in mathematical physics, with relevance to field theory, nonlinear wave propagation, and condensed matter systems. In higher dimensions, the equation admits complex solution structures, including multidimensional solitons and breathers, whose computation is impeded by both the nonlinearity and the high-dimensional nature of the problem. We consider sine-Gordon equation of the following form:
$$ -\Delta u(x) + \sin(u(x)) = f(x), x\in\Omega, $$
$$ u(x) = g(x), x \in \partial\Omega, $$

\noindent
where $\Omega \in [-1, 1]^d$ and the exact solution is assumed to be of the form:

$$u_{\text{exact}}(x) = \biggl(\frac{1}{d}\sum_{i=1}^d x_i\biggl)^2 + \sin\biggl(\frac{1}{d}\sum_{i=1}^d x_i\biggl),$$
and
$$f(x) = \frac{1}{d}\biggl(\sin\biggl(\frac{1}{d}\sum_{i=1}^d x_i\biggl) - 2\biggl) + \sin(u_{\text{exact}}(x)).$$

\begin{table}[H]
\begin{center}
\begin{tabular}{||c | c | c | c | c||}  
 \hline
 Dimension & Architecture & Optimizer & $\#$Collocation & $\#$Data \\ [0.05ex] 
 \hline\hline
 21D & $\bigl[7, 64, 64, 10\bigr] \times 3$ & Adam+L-BFGS & 54880 & 12960 \\ 
 \hline
 99D & $\bigl[33, 64, 64, 40\bigr] \times 3$ & Adam+L-BFGS & 109760 & 14040 \\
 \hline
 300D & $\bigl[100, 64, 64, 128\bigr] \times 3$ & Adam+L-BFGS & 301840 & 34560 \\
 \hline
\end{tabular}
\end{center}
\caption{Anant-Net architecture used for solving high-dimensional sine-Gordon equation}
\label{tab:arch-sg}
\end{table}

\begin{figure}[H]
\centering
{
\centering
\includegraphics[width=0.3\linewidth]{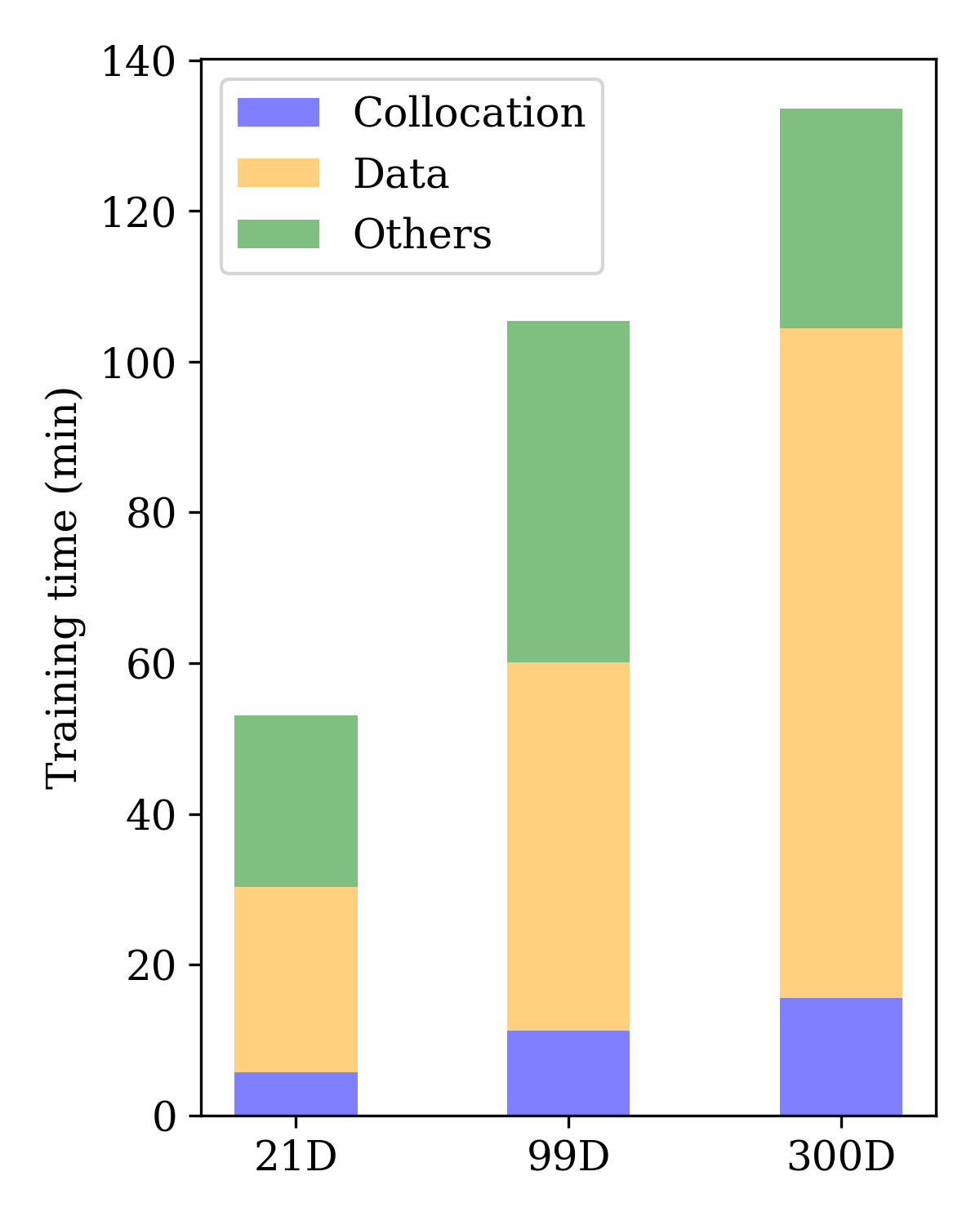}
}
\hspace{0.5cm}
\centering
{
\centering
\includegraphics[width=0.3\linewidth]{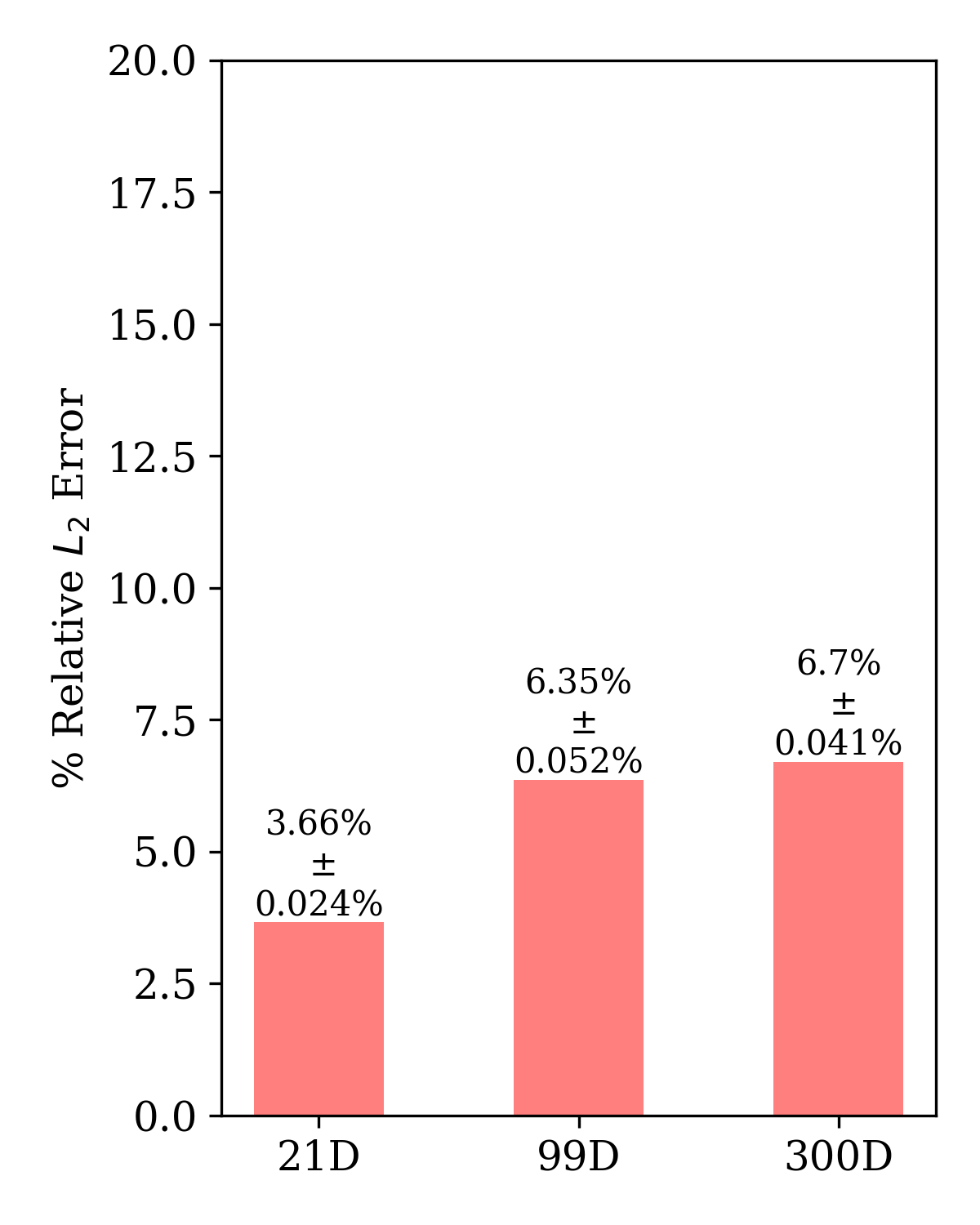}
}
\caption{Sine-Gordon Equation: (Left) Training time profiling; (Right) mean testing error with standard deviation, computed over 10 random realizations (seeds), for the 21D, 99D, and 300D cases.}
\label{fig:SG_AnantNet_Acc_Rtime}
\end{figure}

\noindent
As opposed to the previous section, where Anant-Net was applied to a linear PDE, we now solve a nonlinear PDE; the high-dimensional sine-Gordon equation in 21D, 99D, and 300D. Except for the 300D case, the architectural and optimization hyperparameters remain consistent with the previous setup. Due to the nonlinearity of the sine-Gordon equation, the number of collocation and data samples is slightly higher compared to the Poisson equation. Details of the data sampling are provided in Table \ref{tab:arch-sg}. This increase is justified, as the model must account for both the growing hypercube volume with dimension and the added complexity introduced by nonlinearity. Additionally, for the sine-Gordon equation, we employ a layer-wise adaptive activation function of the form proposed in \cite{jagtap2020locally}. Figure \ref{fig:SG_AnantNet_Acc_Rtime} (left) shows that boundary data sampling remains the primary contributor to training time, consistent with the findings from the previous section. Figure \ref{fig:SG_AnantNet_Acc_Rtime} (right) shows a mean relative $L_2$ testing error along with the standard deviation, evaluated on randomly selected test points from the high-dimensional space. The testing error increases with the dimensionality. Figure \ref{fig:absoluteerrorEE} (middle) shows the absolute point-wise error on randomly selected 3D cubes from the 21D, 99D, and 300D cases. More details on the hyperparameters used for training is explained in APPENDIX \ref{appendix:b}.

\subsection{High-dimensional Allen-Cahn Equation}
\label{sec:Allen-Cahn}
The high-dimensional Allen–Cahn equation is a nonlinear reaction-diffusion PDE that models phase separation phenomena in multi-component systems. Initially formulated to describe the motion of anti-phase boundaries in binary alloys, its extension to high-dimensional domains has garnered significant interest for capturing complex interfacial dynamics and pattern formation in physical, chemical, and biological systems. We consider Allen-Cahn equation of the following form:
$$ -\Delta u(x) + u(x) - u(x)^3 = f(x), x\in\Omega, $$
$$ u(x) = g(x), x \in \partial\Omega, $$
\noindent
where $\Omega \in [-1, 1]^d$ and the exact solution is assumed to be of the form:
$$u_{\text{exact}}(x) = \biggl(\frac{1}{d}\sum_{i=1}^d x_i\biggl)^2 + \sin\biggl(\frac{1}{d}\sum_{i=1}^d x_i\biggl),$$
and
$$f(x) = \frac{1}{d}\biggl(\sin\biggl(\frac{1}{d}\sum_{i=1}^d x_i\biggl) - 2\biggl) + u_{\text{exact}}(x) - u_{\text{exact}}(x)^3.$$
\begin{table}[H]
\begin{center}
\begin{tabular}{||c | c | c | c | c||}  
 \hline
 Dimension & Architecture & Optimizer & $\#$Collocation & $\#$Data \\ [0.05ex] 
 \hline\hline
 21D & $\bigl[7, 64, 64, 10\bigr] \times 3$ & Adam+L-BFGS & 38416 & 8640 \\ 
 \hline
 99D & $\bigl[33, 64, 64, 40\bigr] \times 3$ & Adam+L-BFGS & 264000 & 36450 \\
 \hline
 300D & $\bigl[100, 64, 64, 128\bigr] \times 3$ & Adam+L-BFGS & 274400 & 43200 \\
 \hline
\end{tabular}
\end{center}
\caption{Anant-Net architecture used for solving high-dimensional Allen-Cahn equation}
\label{tab:arch-ac}
\end{table}

\begin{figure}[H]
\centering
{
\centering
\includegraphics[width=0.3\linewidth]{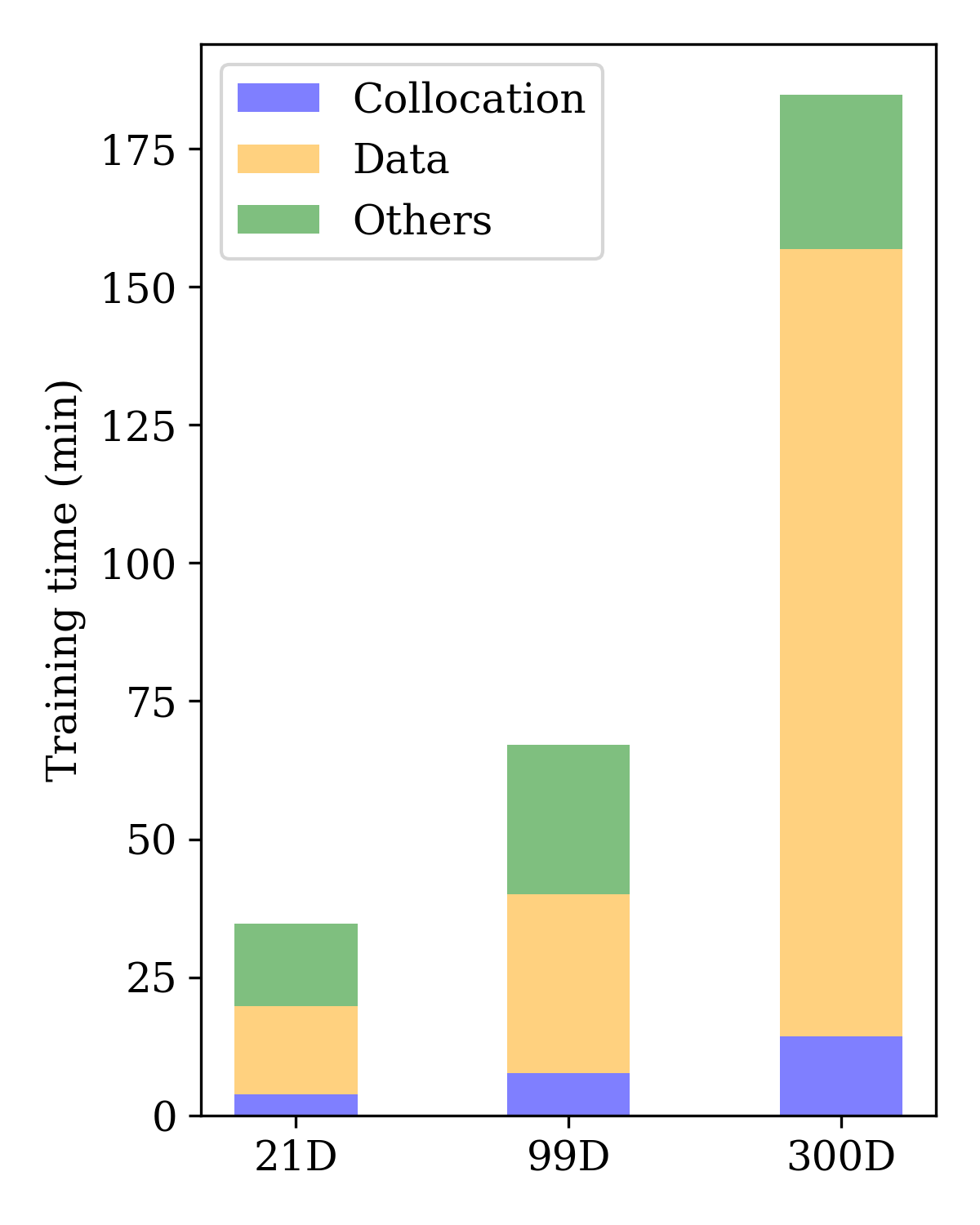}
}
\hspace{0.5cm}
\centering
{
\centering
\includegraphics[width=0.3\linewidth]{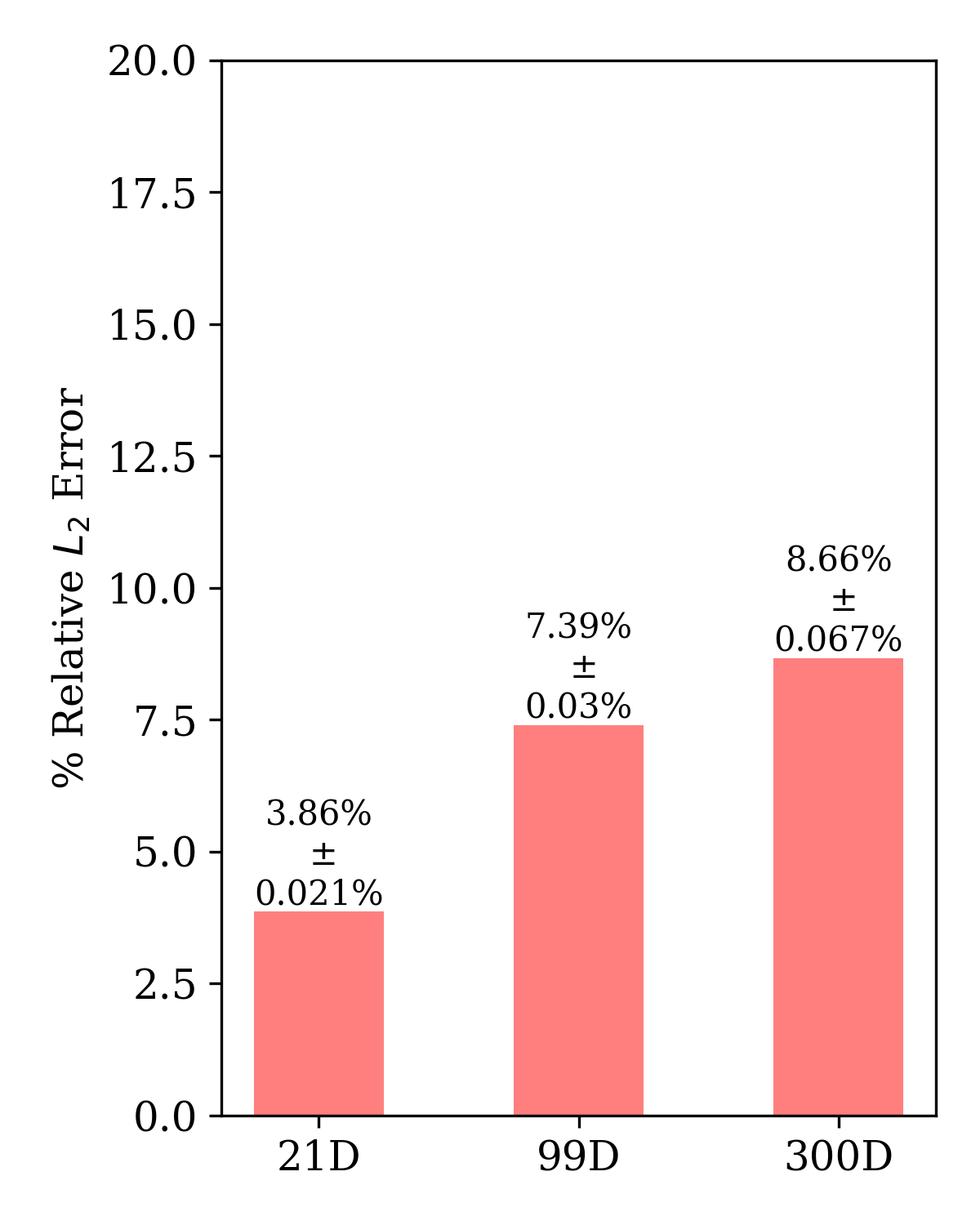}
}
\caption{Allen-Cahn Equation: (Left) Training time profiling; (Right) mean testing error with standard deviation, computed over 10 random realizations (seeds), for the 21D, 99D, and 300D cases.}
\label{fig:AC_AnantNet_Acc_Rtime}
\end{figure}
\noindent
The high-dimensional Allen–Cahn equation is solved in 21D, 99D, and 300D with minimal modifications to the architecture and optimization hyperparameters used in the previous sections. Details on hyperparameters and data sampling are provided in Table \ref{tab:arch-ac}. Although we again address a nonlinear PDE, in this case we employ a fixed nonlinear activation function ($\sin$) without adaptive parameters. As illustrated in Figure \ref{fig:AC_AnantNet_Acc_Rtime} (left), the performance trends align with earlier observations: boundary data sampling remains the dominant contributor to the overall computational cost during training. An exception is observed in the same figure, where the training times for the 21D and 99D cases are notably lower compared to similar problems discussed in previous sections. This reduction in runtime is likely attributable to the fewer training iterations required to achieve comparable accuracy for the Allen–Cahn equation when using Anant-Net. Figure \ref{fig:AC_AnantNet_Acc_Rtime} (right) shows the mean and standard derivation of relative $L_2$ testing error for 21D, 99D, and 300D cases. Figure \ref{fig:absoluteerrorEE} (right) shows the absolute point-wise error on randomly selected 3D cubes from the 21D, 99D, and 300D cases. Additional details on the number of training iterations used across all high-dimensional PDEs are provided in the APPENDIX \ref{appendix:b}.

\subsection{Effect of Boundary Data Sampling, Collocation Points Sampling, and Batch Size on Anant-Net Performance}
The optimality of boundary data sampling reflects a fundamental trade-off between model accuracy and computational efficiency. While a large volume of boundary data can improve prediction accuracy, it often incurs substantial computational costs during training. Conversely, reducing the volume of boundary data lowers computational demand but may compromise model accuracy. Striking a balance between these two factors is particularly important when solving high-dimensional PDEs. In this section, we examine the sensitivity of Anant-Net’s accuracy to varying volumes of boundary data in high-dimensional PDE problems. Although test accuracy may also depend on factors such as network architecture, the quantity and quality of boundary data are expected to have a significant influence. Results are presented for both linear and nonlinear non-homogeneous high-dimensional PDEs using clean data.
\begin{figure}[ht]
\centering
{
\centering
\includegraphics[width=0.24\linewidth]{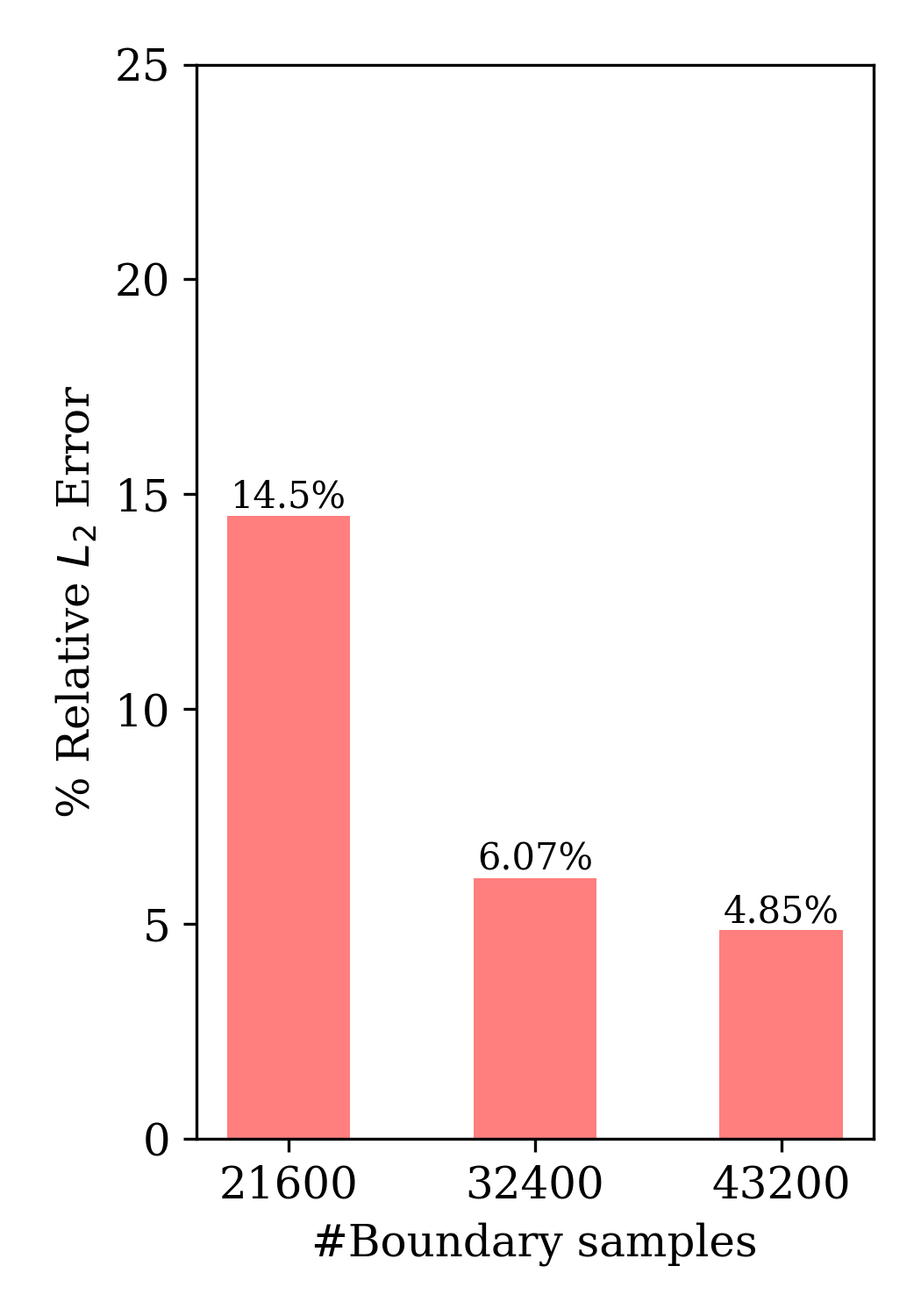}
\includegraphics[width=0.22\linewidth]{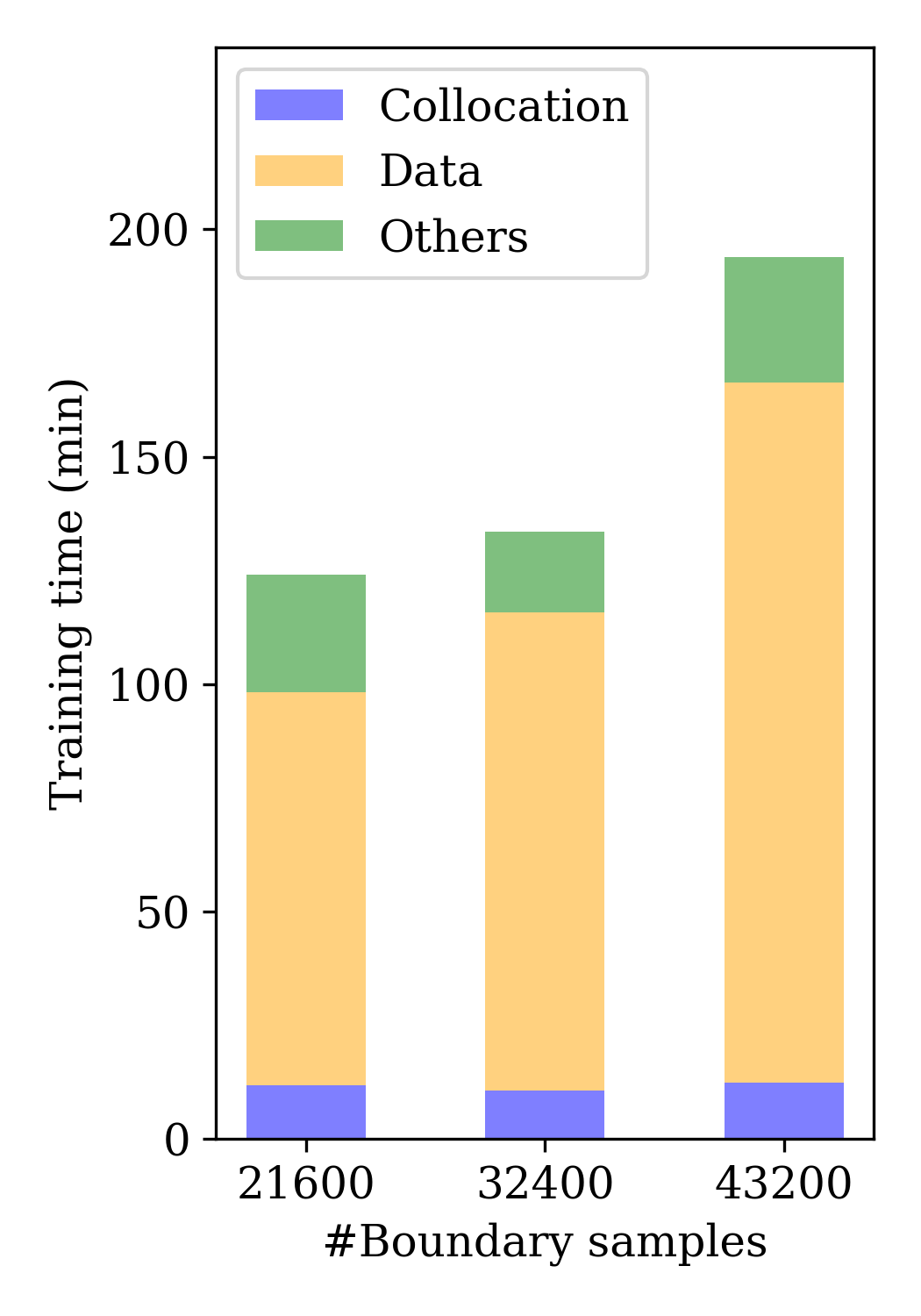}
}
\hspace{0.4cm}
\centering
{
\centering
\includegraphics[width=0.24\linewidth]{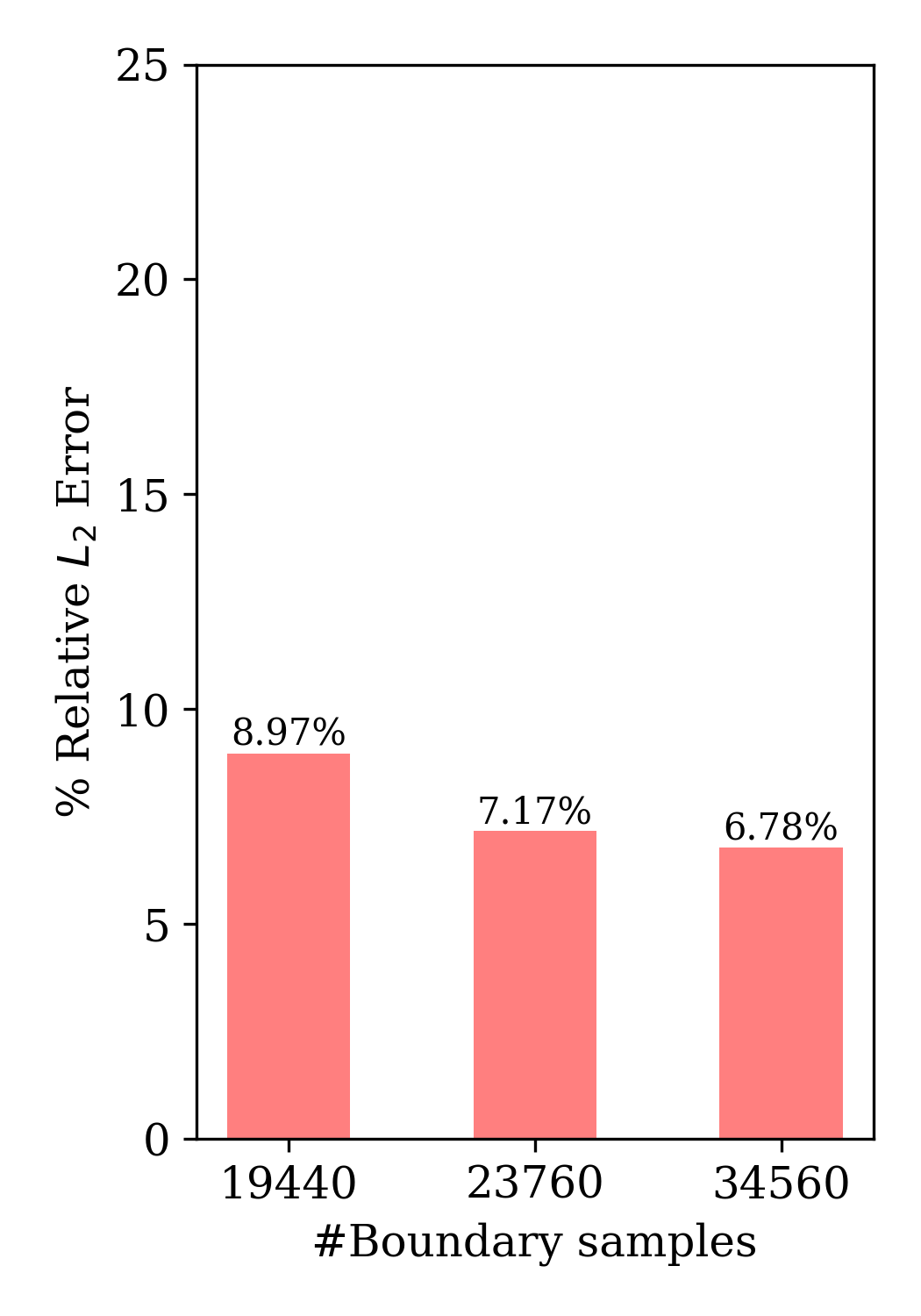}
\includegraphics[width=0.22\linewidth]{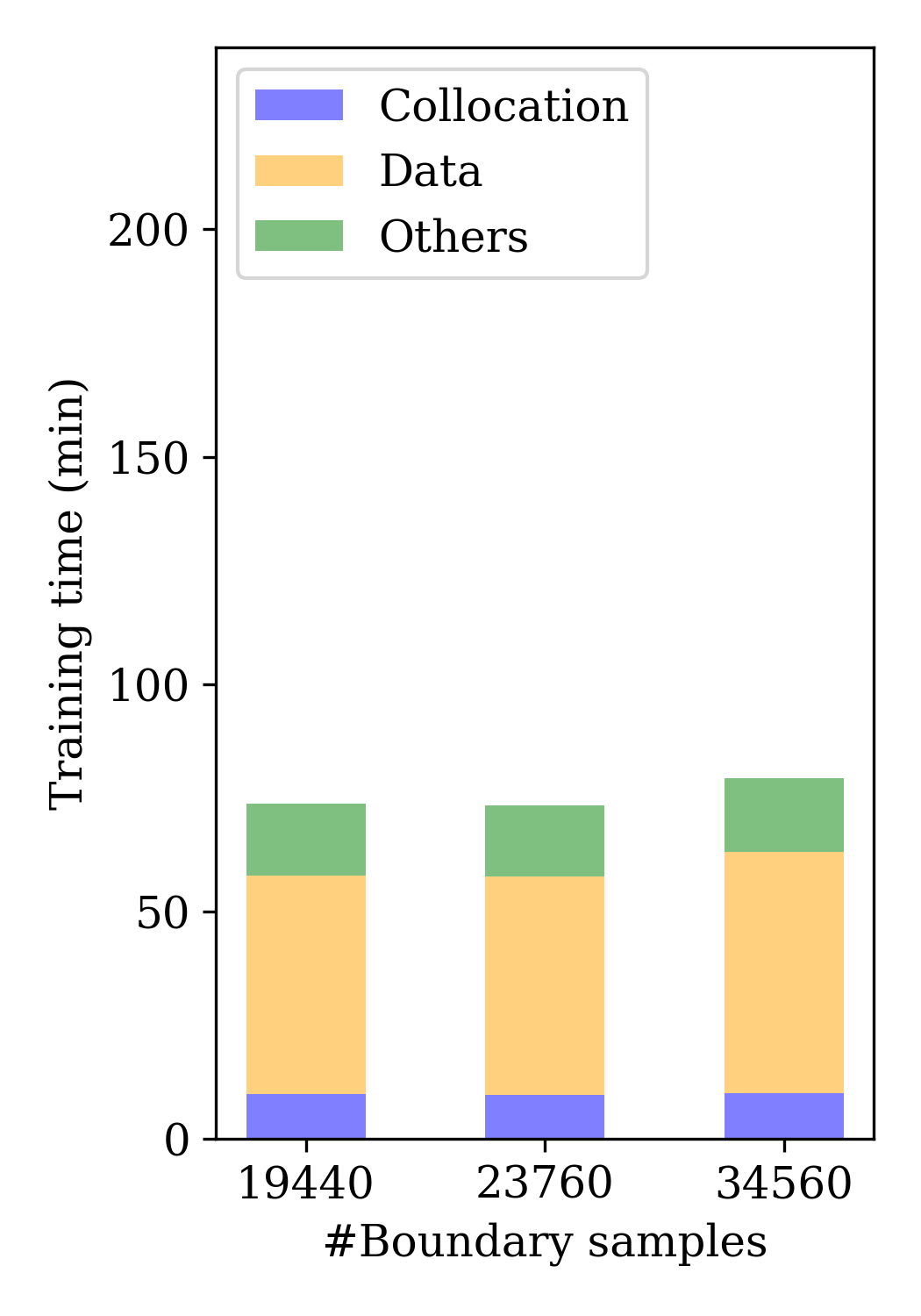}
}
\caption{Testing accuracy on random samples and training run-time for 300D (Left) Poisson equation and (Right) sine-Gordon equation.}
\label{fig:boundata-acc-300D}
\end{figure}
In particular, we consider the 300-dimensional Poisson and sine-Gordon equations. In this study, the volume of boundary data is varied while keeping the volume of collocation points, network architecture, and all other hyperparameters fixed for Anant-Net. Profiling of the training time (Fig. \ref{fig:boundata-acc-300D}) confirms that variations in boundary data volume are the primary contributors to changes in computational cost. As shown in Fig. \ref{fig:boundata-acc-300D}, for both linear and nonlinear high-dimensional PDEs, the test accuracy, evaluated on randomly sampled test points, initially improves with increased boundary data but tends to plateau beyond a certain threshold. For the 300-dimensional Poisson and sine-Gordon equations, training is performed using an initial sweep with the Adam optimizer followed by L-BFGS. For the linear high-dimensional Poisson equation, a clear trend is observed wherein increased boundary data enhances model accuracy, albeit at the cost of longer training times (Fig. \ref{fig:boundata-acc-300D}). A similar pattern is evident for the nonlinear high-dimensional sine-Gordon equation. These results indicate that, regardless of the underlying PDE type, increasing the boundary data volume improves the predictive accuracy of Anant-Net, but at the expense of higher computational cost.

\begin{figure}[ht]
\centering
{
\centering
\includegraphics[width=0.24\linewidth]{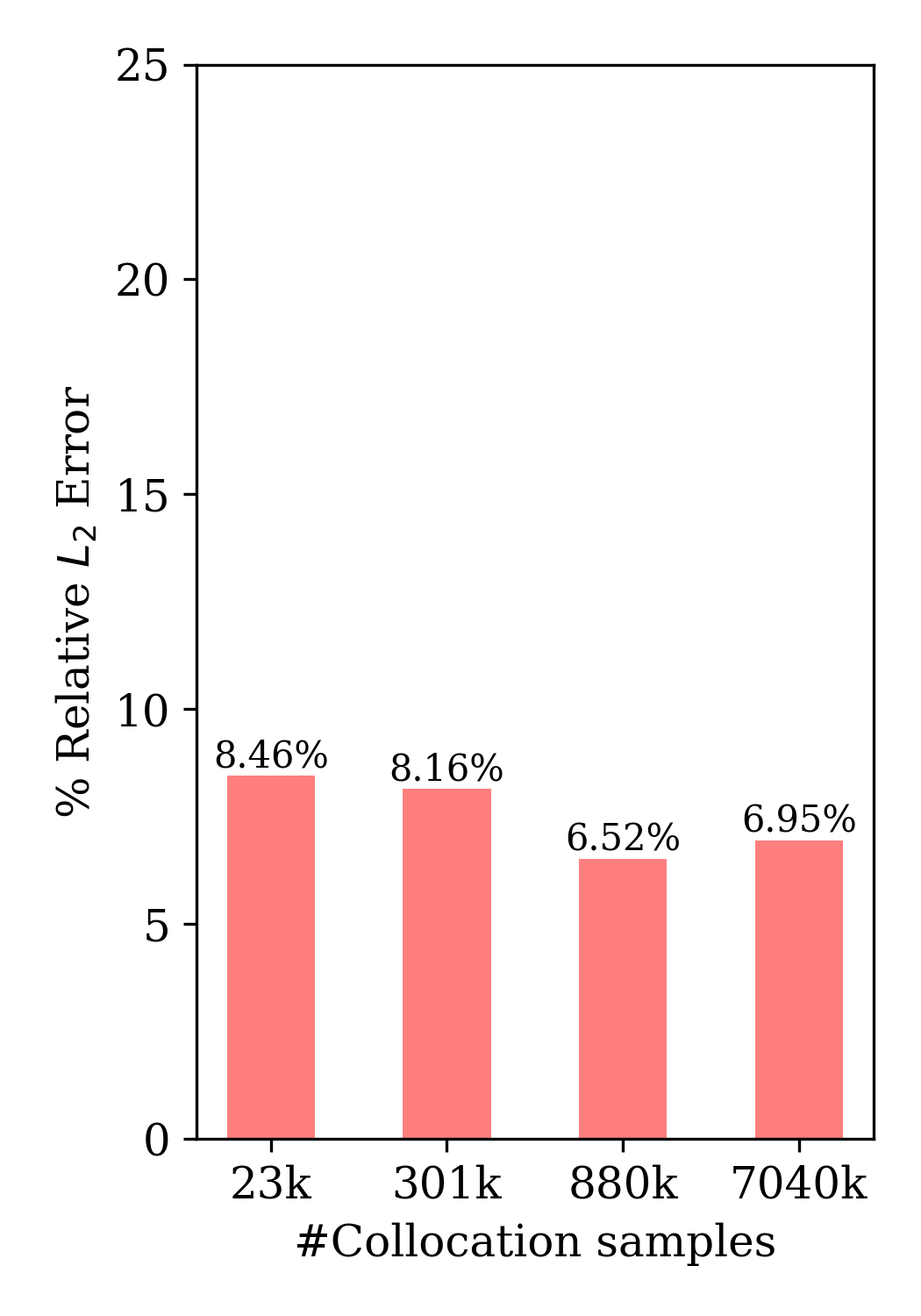}
\includegraphics[width=0.22\linewidth]{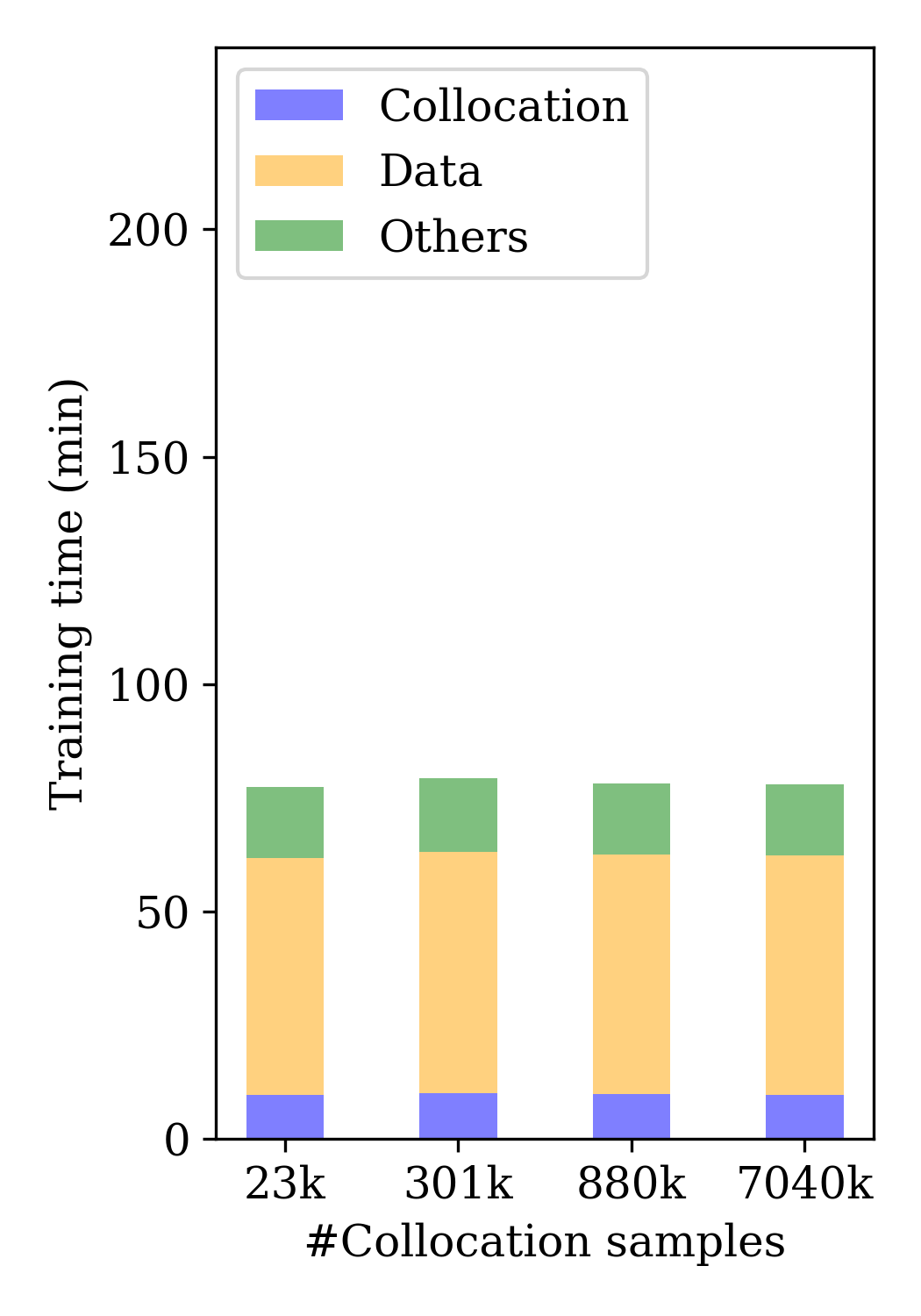}
}
\hspace{0.4cm}
\centering
{
\centering
\includegraphics[width=0.24\linewidth]{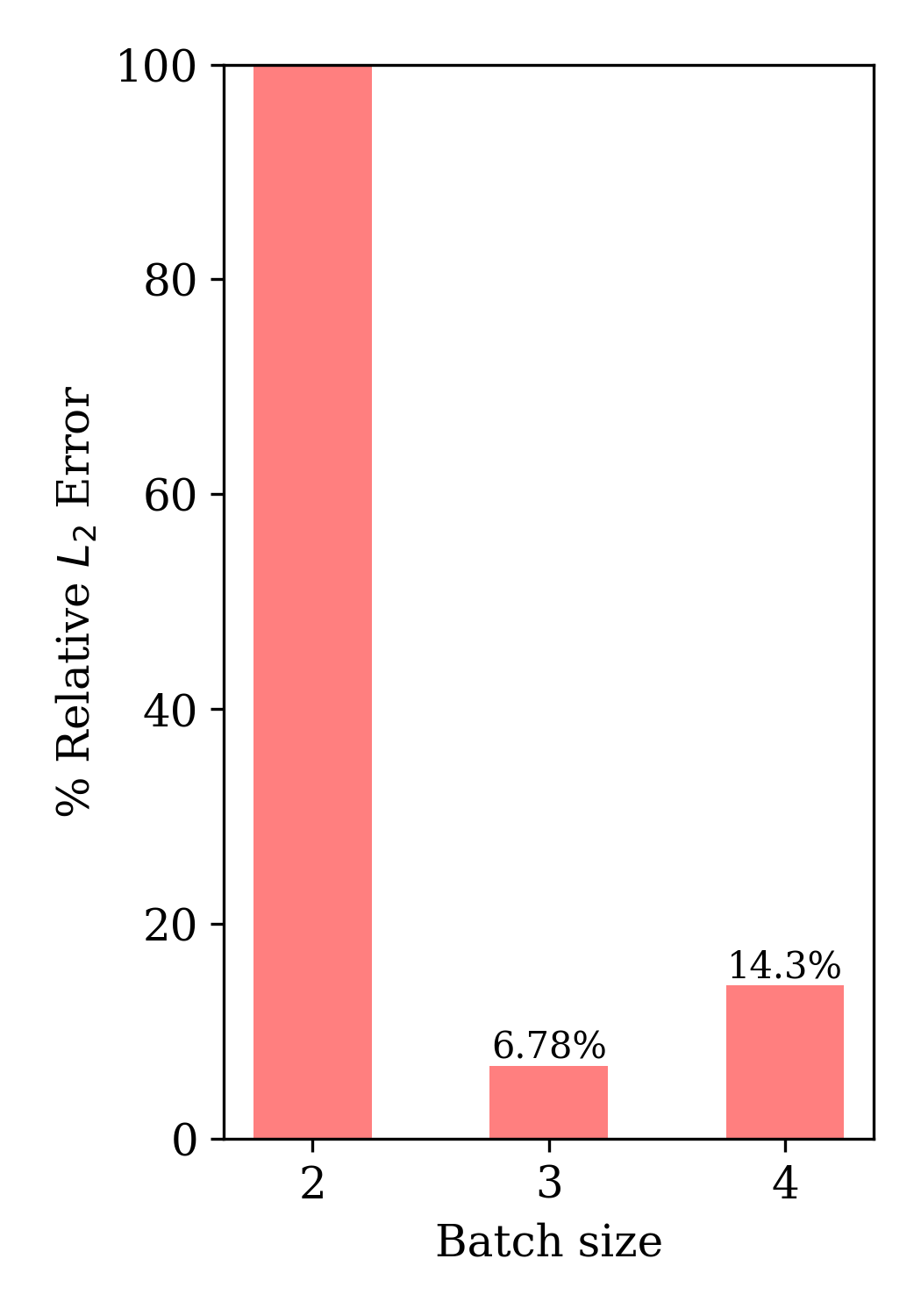}
\includegraphics[width=0.22\linewidth]{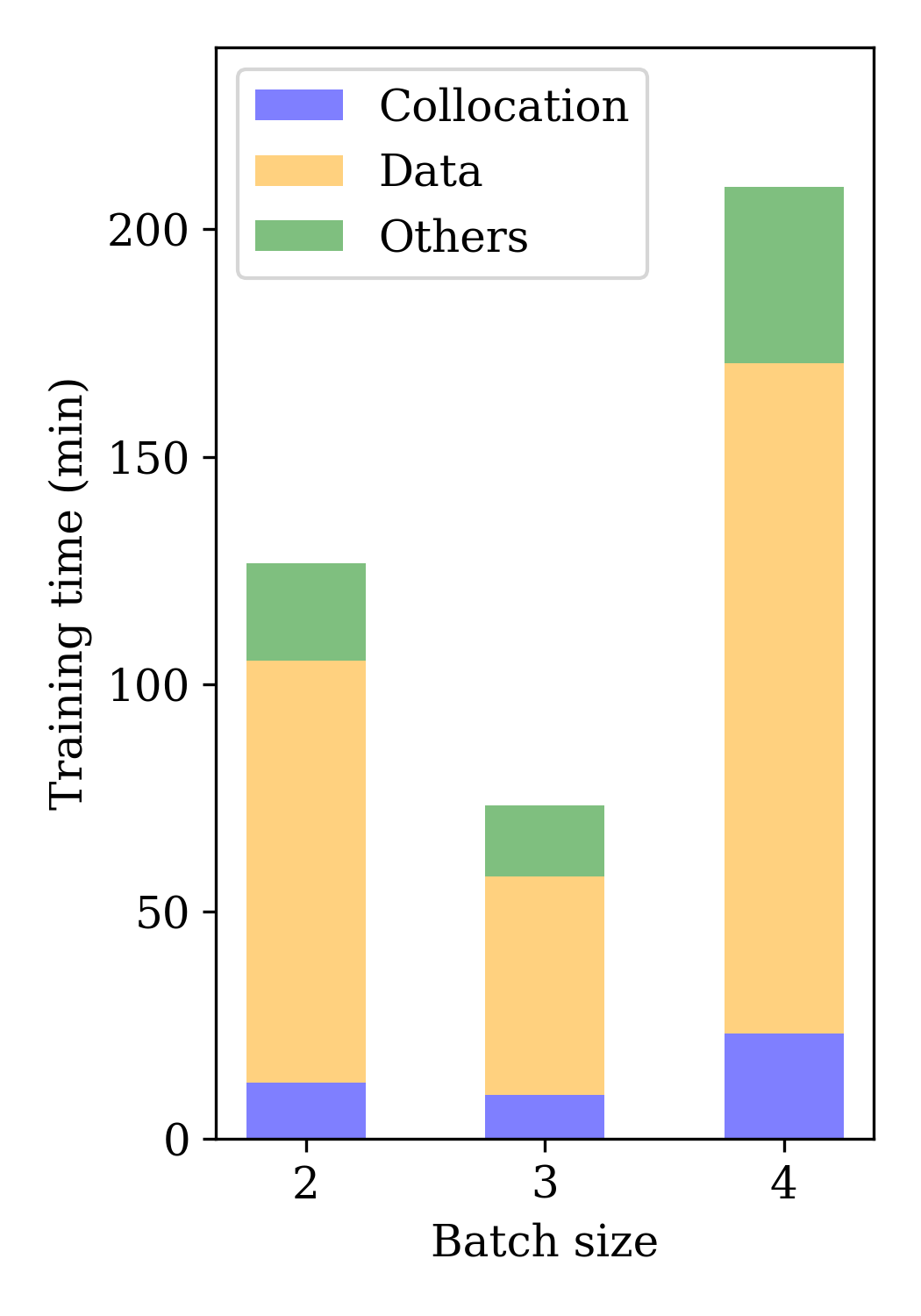}
}
\caption{Sensitivity of (Left) \textit{collocation grid density} ($\#CG \times (N_C)^B$) and (Right) \textit{batch size} ($B$) on testing accuracy and training run-time for 300D sine-Gordon equation while keeping all the other hyperparameters (such as the sampling frequency, depth/width of the body networks etc.) fixed.}
\label{fig:coll-acc-300D}
\end{figure}
\noindent
The sensitivity of hyperparameters are further studied from the perspective of collocation grids and batch size ($i.e.$ number of body networks). This is shown in the Figure \ref{fig:coll-acc-300D}. As the number of collocation samples increases, the accuracy of Anant-Net increases with negligible changes in the training time as shown in Figure \ref{fig:coll-acc-300D} (Left). Note that number of collocation samples is increased by increasing the resolution of the collocation grids ($N_C$) while keeping the number of collocation grids ($\#CG$) constant. The increase in the resolution of collocation grids is seamlessly handled by the tensor product operation without any memory bottleneck while keeping the GPU below its saturation limit leading to constant runtime as shown in the Figure \ref{fig:coll-acc-300D} (Left). Figure \ref{fig:coll-acc-300D} (Right) shows the sensitivity of batch size ($B$) on the performance of Anant-Net while keeping all the other hyperparameters unchanged and maintaining almost similar number of collocation and data samples across all the cases. Here, we recognize batch size = 3 as the optimal batch size for the 300D sine-Gordon equation and reducing the batch size to 2 showcases dramatic drop in the accuracy of Anant-Net on randomly selected testing data samples. This shows that batch size = 2 does not have sufficient capacity to approximate the solution to a 300D non-linear partial differential equation. However, when batch size = 4, although the accuracy is substantially better than batch size = 2, we observe sufficient drop in the accuracy of Anant-Net compared to batch size = 3. Although this seems like an anomaly at first, it is worthwhile to note that the accuracy of Anant-Net for batch size = 4 can be improved further by fine-tuning the fixed hyperparameters (such as the sampling frequency, depth/width of the body networks etc.). The difference in run-time observed across different values of batch size in Figure \ref{fig:coll-acc-300D} (Right) is representative of the difference in number of collocation grids ($\#CG$) and number of boundary data grids ($\#BG$) which are needfully changed in order to keep the total number of collocation samples ($\#CG\times(N_C)^B$) and total number of data samples ($\#BG\times(N_B)^B$) constant across the different cases of varying batch size ($B$).


\subsection{Anant-KAN Results}
\label{sec:Anant-KAN}
Recent studies have explored various basis functions for function approximation using KANs, targeting both continuous functions and PDE solutions. Table \ref{tab:KAN-basis0} highlights efficient bases for PDE solution spaces. We extend these bases to high-dimensional PDEs using our proposed architecture, Anant-Net. Theorem 2.1 in \cite{liu2024kan} shows that KAN approximation error depends only on grid size $G$ and spline order $k$, not on input dimension, suggesting its ability to overcome the curse of dimensionality; validated empirically in this work. While \cite{liu2024kan} emphasizes splines as universal bases via the Kolmogorov-Arnold theorem, alternative bases like Chebyshev polynomials have also shown promise. Chebyshev-KAN \cite{ss2024chebyshev} benefits from orthogonality and recursive computation, improving training stability and efficiency. However, instability was noted for low-dimensional PDEs in \cite{shukla2024comprehensive}, causing training errors to diverge. This issue was addressed by introducing Modified Chebyshev-KAN, which incorporates $\tanh$ activations in all but the final layer to suppress high-frequency instabilities.

\begin{table}[htbp]
\begin{center}
\begin{tabular}{||c | c | c | c||}  
 \hline
 Network & Method & Basis $(\phi)$ & Activation \\ [0.05ex] 
 \hline\hline
 Anant-KAN1 & KAN \cite{liu2024kan} & Spline & -\\ 
 \hline
 Anant-KAN2 & Chebyshev KAN \cite{ss2024chebyshev} & Chebyshev Polynomials & Tanh$^{(1)}$\\
 \hline
 Anant-KAN3 & Modified Chebyshev KAN \cite{shukla2024comprehensive} & Chebyshev Polynomials & Tanh$^{(2)}$\\
 \hline
 Anant-KAN4 & Fourier KAN \cite{10763509} & Fourier ($a.cos(kx) + b.sin(kx)$) & -\\
 \hline
\end{tabular}
\end{center}
\caption{Variants of KAN with different basis functions $(\phi)$. $^{(1)}$ represents tanh activation applied only at the input layer for normalization while $^{(2)}$ represents tanh activation applied to all layer except the last layer. Anant-KAN for different basis functions for $\bigl[7, 64, 64, 10\bigr] \times 3$ configuration of architecture.}
\label{tab:KAN-basis0}
\end{table}

\begin{figure}[ht]
\centering
{
\includegraphics[width=0.55\linewidth]{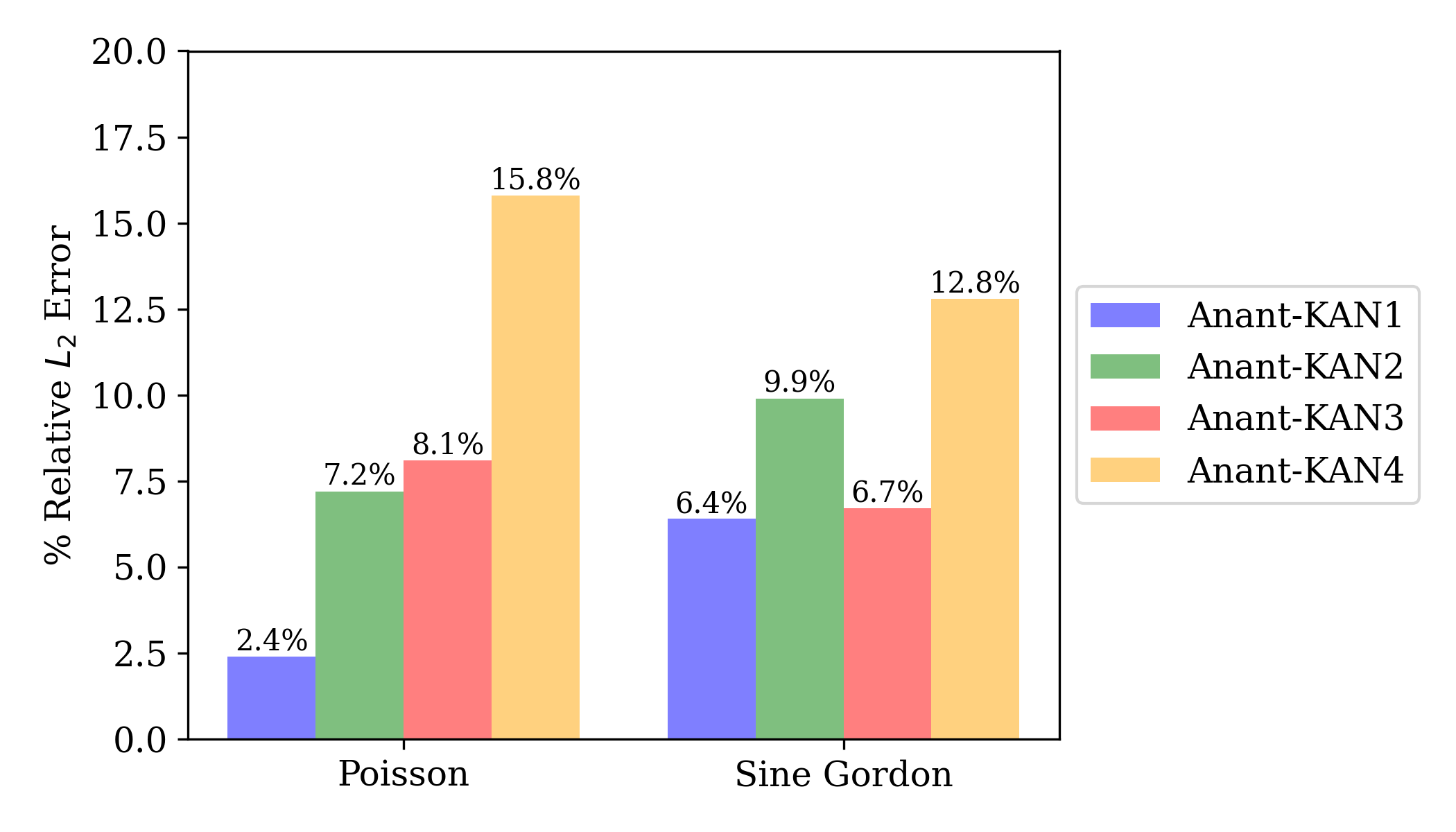}
}
\caption{Comparison of Anant-KAN with different basis functions for solving 21D Poisson equation and sine-Gordon equation for a fixed order of basis $(k) = 3$.}
\label{fig:kan_basis_accuracy}
\end{figure}

\noindent
In this section, we focus on enhancing the interpretability of our proposed architecture by leveraging the KAN, referred to here as Anant-KAN. This is the first application of KAN to high-dimensional PDEs. One of the primary advantages of KAN lies in its parameter efficiency, as demonstrated in Figure \ref{fig:kan_basis_accuracy}. However, despite its compact parameterization, KAN exhibits significant computational overhead, which persists regardless of the dimensionality of the PDE. Our computational experiments further corroborate this limitation, especially in high-dimensional settings.

\begin{figure}[ht]
\centering
{
\includegraphics[width=0.35\linewidth, trim={0 0.6cm 1cm 0}]{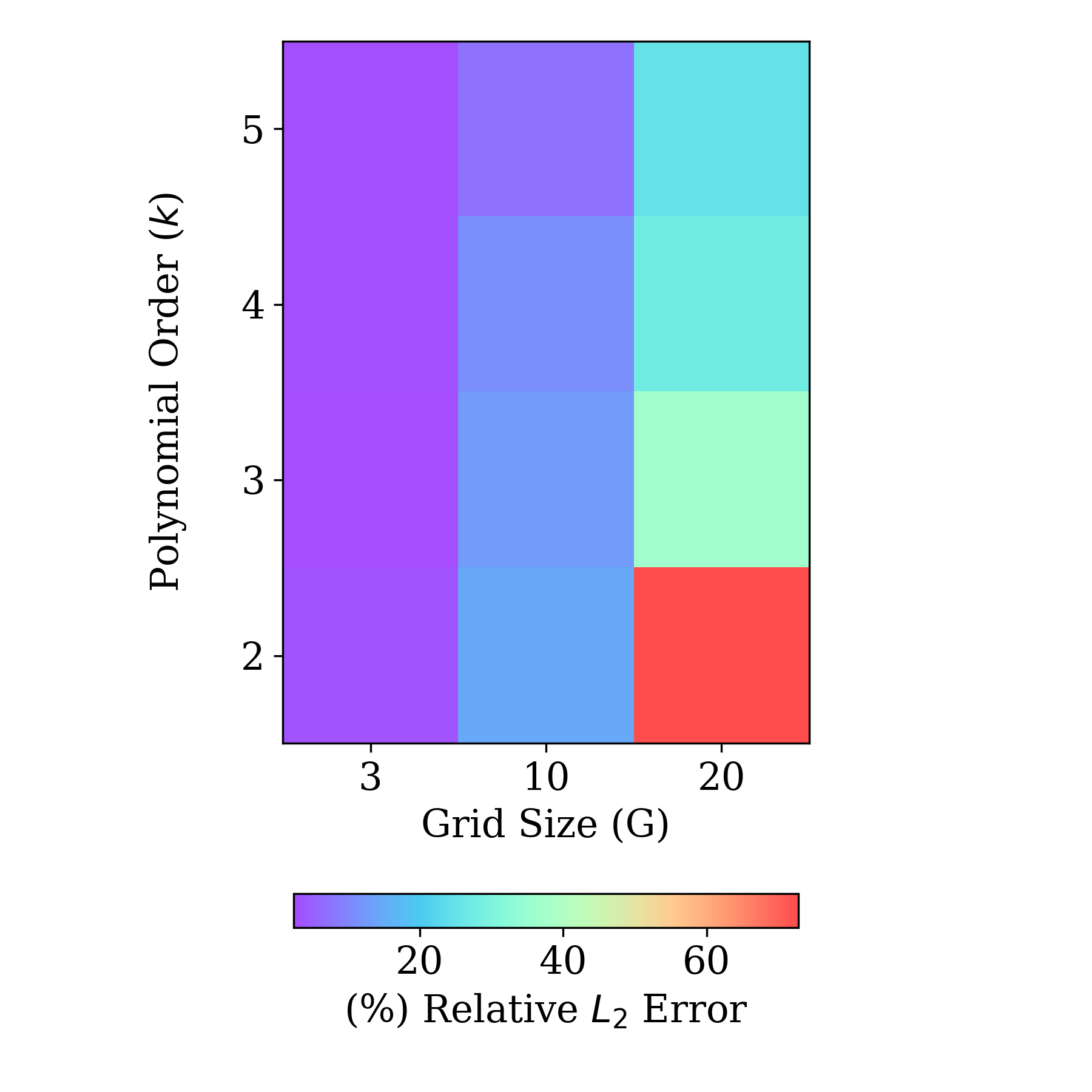}
}
\hspace{0.1cm}
{
\includegraphics[width=0.35\linewidth, trim={0 0.6cm 1cm 0}]{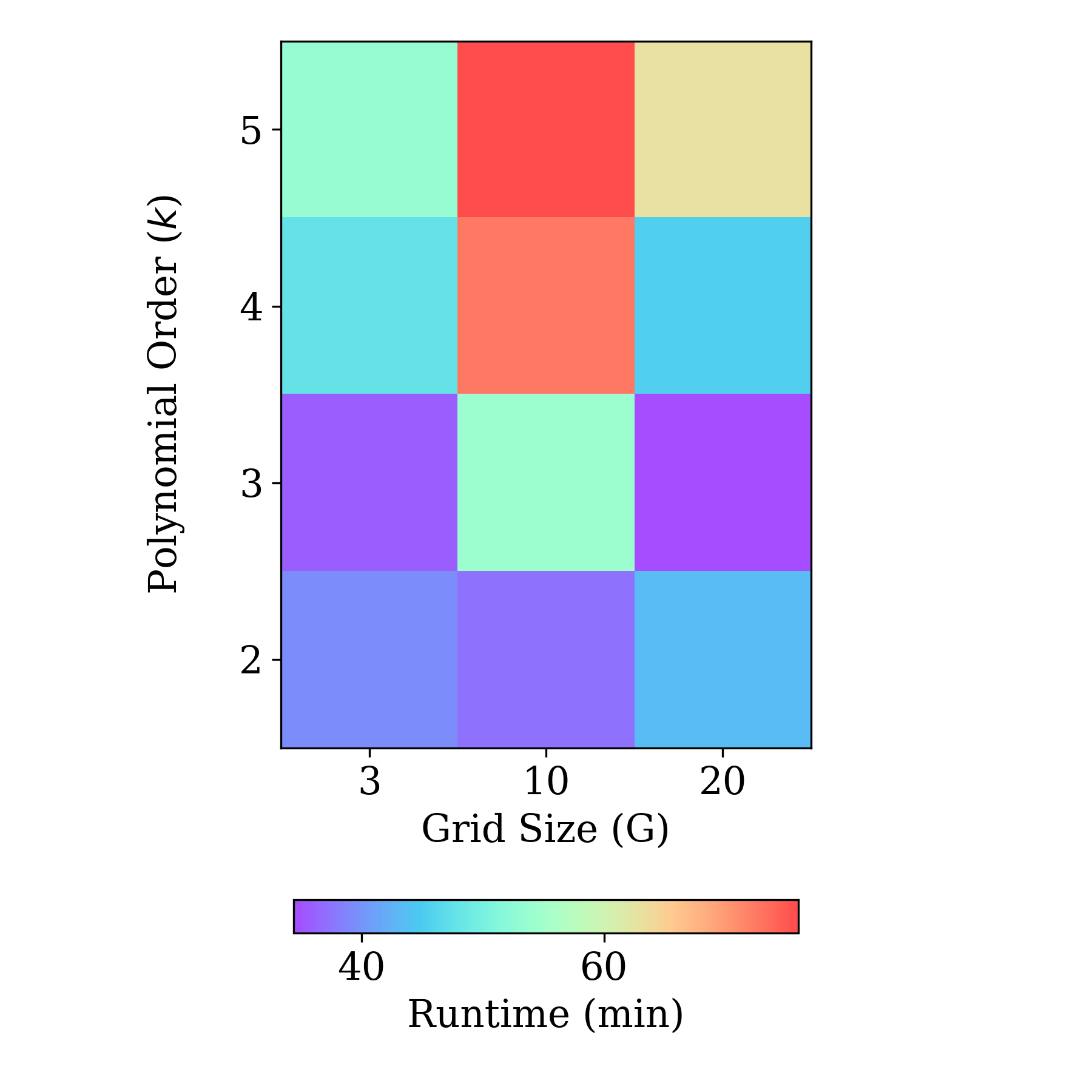}
}
\caption{Sensitivity of polynomial order ($k$) and grid size ($G$) on relative $L_2$ error and run-time for 21D Poisson equation using Anant-KAN with spline basis (Anant-KAN1).}
\label{fig:kan_sensitivity}
\end{figure}

\begin{figure}[ht]
\centering
{
\centering
\includegraphics[width=0.4\linewidth]{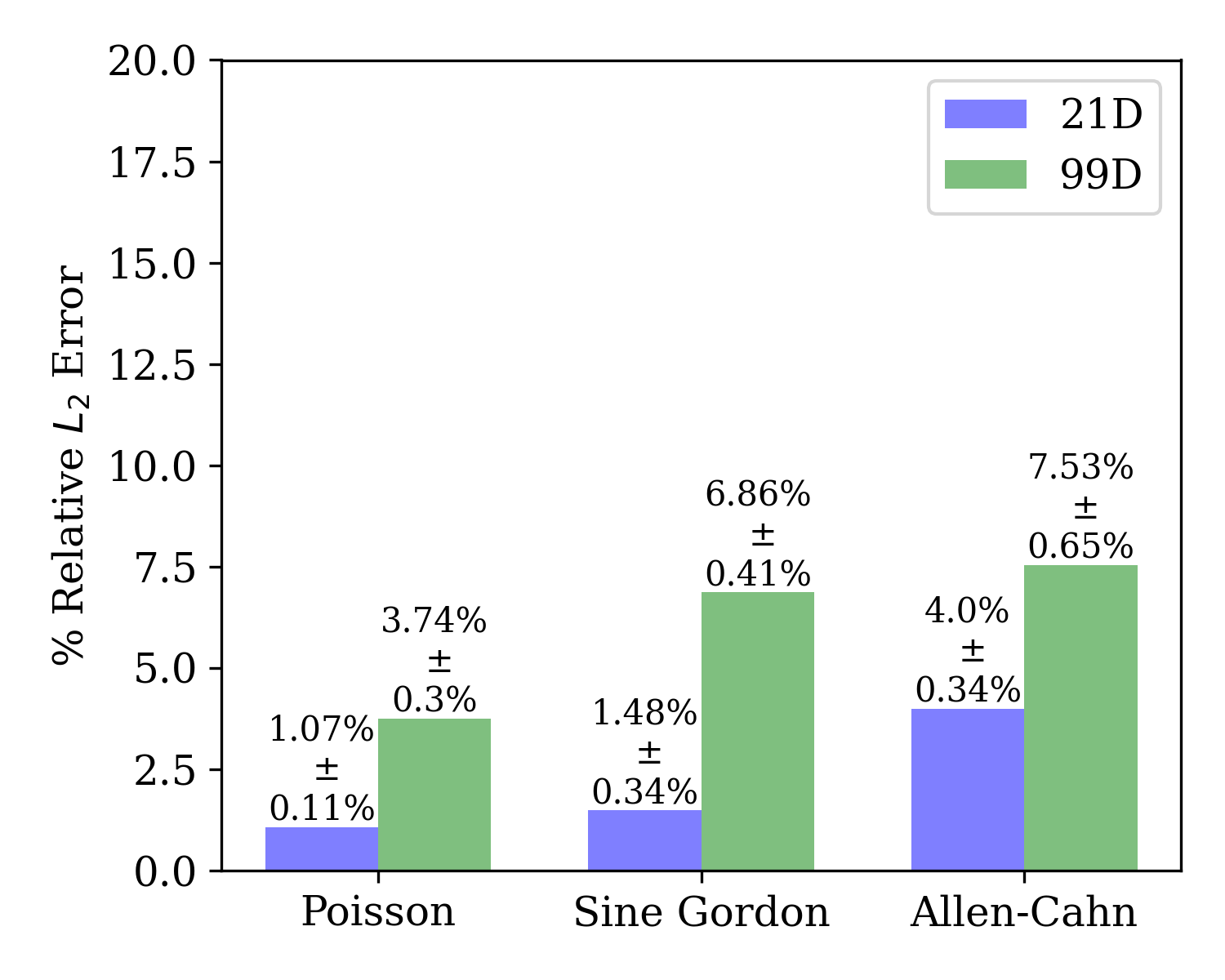}
}
\hspace{0.5cm}
{
\centering
\includegraphics[width=0.4\linewidth]{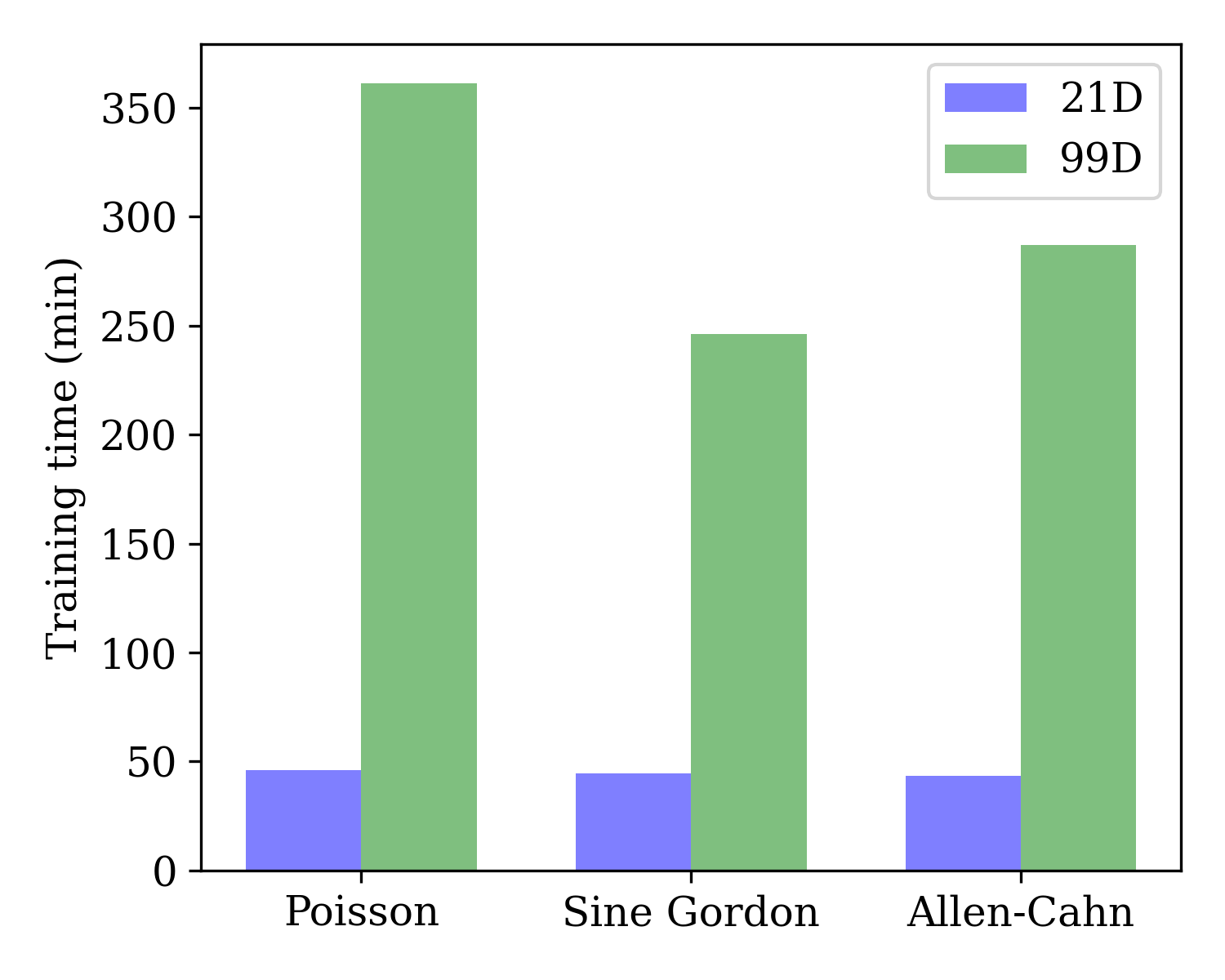}
}
\caption{Comparison of Anant-KAN with spline basis (order ($k$) = 2, grid size ($G$) = 5) for 21D and 99D Poisson equation, sine-Gordon equation, and Allen-Cahn equation using layer-wise adaptive-sin activation \cite{jagtap2020locally} for all body networks. Note that the mean relative $L_2$ error on random samples with standard deviation was obtained from 10 random realizations (seeds).}
\label{fig:kan_p_sg_compare_accuracy}
\end{figure}

The key findings of the Anant-KAN architecture are as follows:
\begin{itemize}
\item For both linear and non-linear PDEs, spline basis functions consistently demonstrate superior approximation capability in high-dimensional PDEs using Anant-KAN framework. This performance trend is clearly illustrated in Fig. \ref{fig:kan_basis_accuracy}.
\item Building on this observation, we further refine our investigation by identifying the optimal configuration of the spline basis, characterized by the pair $(k, G)$, denotes the polynomial order and the grid resolution, respectively. The sensitivity of Anant-KAN to variations in these parameters is presented in Fig. \ref{fig:kan_sensitivity}.
\item  In general, for a fixed value of grid size $G$, increasing the polynomial order $k$ improves the accuracy of the Anant-KAN. Also, accuracy becomes more sensitive to changes in $k$ at larger values of $G$.
\item  In contrast, runtime analysis reveals that increasing $k$ for any fixed $G$ incurs a substantial computational cost, as shown in the Fig. \ref{fig:kan_sensitivity} (right). Consequently, for high-dimensional PDEs, a parameter configuration of $k \leq 3$ and $G \leq 10$ makes an effective balance between accuracy and computational efficiency.
\item We apply the Anant-KAN architecture to solve high-dimensional PDEs using the configuration $k = 2$ and $G = 5$, coupled with additional hyperparameter tuning strategies such as adaptive activation functions \cite{jagtap2020adaptive,jagtap2020locally,jagtap2022deep, jagtap2023important}. The resulting improvements in predictive performance are shown in Fig. \ref{fig:kan_p_sg_compare_accuracy}.
\item Finally, we observe a distinct increase in relative $L_2$ error for the high-dimensional Poisson equation as the grid resolution $G$ increases. This phenomenon is likely attributable to overfitting, a hypothesis supported by earlier observations in \cite{liu2024kan}, where an abrupt rise in testing error occurred beyond a certain grid size, termed the \textit{interpolation threshold}. While this behavior was originally reported in low-dimensional problems, our study demonstrates its early onset in high-dimensional PDE contexts, suggesting the need for regularization or adaptive complexity control in such settings.
\end{itemize}

The high computational cost associated with KAN imposes a significant constraint on training strategies for Anant-KAN and similar architectures. In particular, the use of second-order optimization methods such as L-BFGS becomes computationally prohibitive for high-dimensional PDEs, necessitating reliance on first-order optimizers such as Adam. This limitation is clearly reflected in our computational experiments. Among the various basis functions evaluated, the spline basis consistently provides the most accurate representation of solutions to high-dimensional PDEs, as shown in Fig.~\ref{fig:kan_basis_accuracy}. On the other hand, the Fourier basis (Anant-KAN4) performs the poorest across both linear and non-linear PDEs. 

\begin{figure}[ht]
\centering
{
\includegraphics[trim = {0cm 5.5cm 0cm 3cm}, clip, scale=0.5]{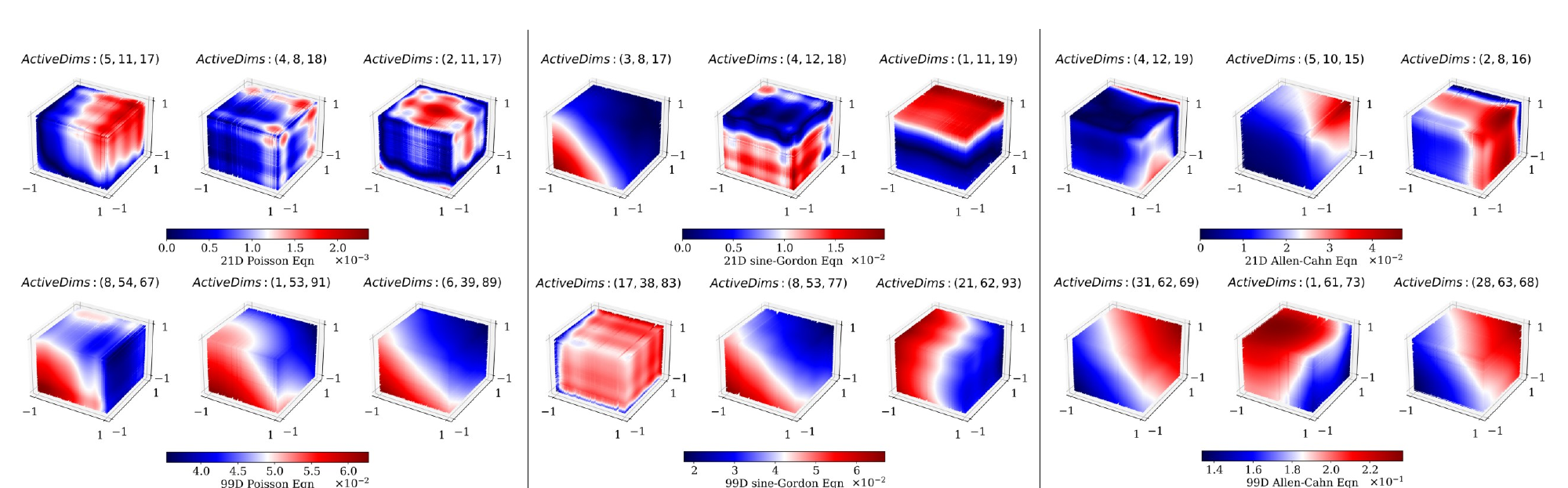}
}
\caption{Point-wise absolute testing error for random 3D hyper-surfaces of 21D, 99D (\textit{left}) Poisson equation (\textit{center}) sine-Gordon equation, (\textit{right}) Allen-Cahn equation. Here, "$\mathsf{ActiveDims}$: $(p, q, r)$" represents the active dimensions in the $d$-dimensional space where gradients are non-zero and $\mathrm{D} - \{p,q,r\}$ are inactive. Also, $\mathrm{D} = \{1,2,\dots,d\}$ is a set which constitutes all the dimensions.}  
\label{fig:AbsError_AnantKAN}
\end{figure}

The Anant-KAN architecture, which employes spline basis functions, is further enhanced through fine-tuning, enabling accurate solutions of both linear and nonlinear high-dimensional partial differential equations, as demonstrated in Figure \ref{fig:kan_p_sg_compare_accuracy}. These results strongly suggest that the proposed Anant-KAN architecture can achieve exceptional accuracy for a broad class of PDEs. The point-wise absolute error associated with Anant-KAN, evaluated over a randomly sampled set of dimensions from the high-dimensional space, is presented in Figure \ref{fig:AbsError_AnantKAN}. All experiments in this section are conducted on a high-RAM CPU system with a 50 GB RAM configuration. Nonetheless, the implementation can be further optimized for GPU acceleration to enable parallelism and scale Anant-KAN to even higher-dimensional problems.

\subsection{High-dimensional Transient Heat Equation}
\label{sec:Heat}
\noindent
The high-dimensional heat equation is a parabolic PDE that arises in various scientific and engineering disciplines, which governs unsteady processes such as thermal diffusion, mass diffusion and the like. Here, we consider the heat equation of the following form:
\begin{equation}
\label{eq:Heat}
   \partial_tu-\Delta u = f(x, t), (x,t)\in\Omega \times [0,T],
\end{equation}
\begin{equation}
\label{eq:HeatBoundary}
    u(x,t) = g(x,t), (x,t) \in \partial\Omega \times [0,T],
\end{equation}
\begin{equation}
\label{eq:HeatInitial}
    u(x,0) = h(x), x \in \Omega,
\end{equation}
\noindent
where $\Omega \in [-1, 1]^d$, $T = 1$ and the exact solution is assumed to be of the form:

$$u_{\text{exact}}(x) = \cos\biggl(\frac{1}{d}\sum_{i=1}^d x_i\biggl)\exp(-t),$$
and
$$f(x) = \bigl(\frac{1}{d} - 1 \bigr)\cos\biggl(\frac{1}{d}\sum_{i=1}^d x_i\biggl)\exp(-t),$$
$$g(x, t) = \cos\biggl(\frac{1}{d}\sum_{i=1}^d x_i\biggl)\exp(-t),$$
$$h(x) = \cos\biggl(\frac{1}{d}\sum_{i=1}^d x_i\biggl).$$
\noindent
\begin{table}[H]
\begin{center}
\begin{tabular}{||c | c | c | c | c||}  
 \hline
 Dimension & Architecture & Optimizer & $\#$Collocation & $\#$Data \\ [0.05ex] 
 \hline\hline
 20D & $\bigl[1, 64, 64, 15\bigr] \times 1, \bigl[10, 64, 64, 15\bigr] \times 2 $ & Adam & 54880 & 50000\\ 
 \hline
 100D & $\bigl[1, 64, 64, 50\bigr] \times 1, \bigl[50, 64, 64, 50\bigr] \times 2 $ & Adam & 164640 & 90000\\
 \hline
 300D & $\bigl[1, 64, 64, 160\bigr] \times 1, \bigl[150, 64, 64, 160\bigr] \times 2 $  & Adam & 411600 & 164000\\
 \hline
\end{tabular}
\end{center}
\caption{Anant-Net architecture used for solving high-dimensional heat equation. Note that $\#$Data represents the number of data points. In this case, $\#$Data $=$ Number of boundary samples $=$ Number of initial samples.}
\label{tab:arch-heat}
\end{table}

\begin{figure}[H]
\centering
{
\centering
\includegraphics[width=0.3\linewidth]{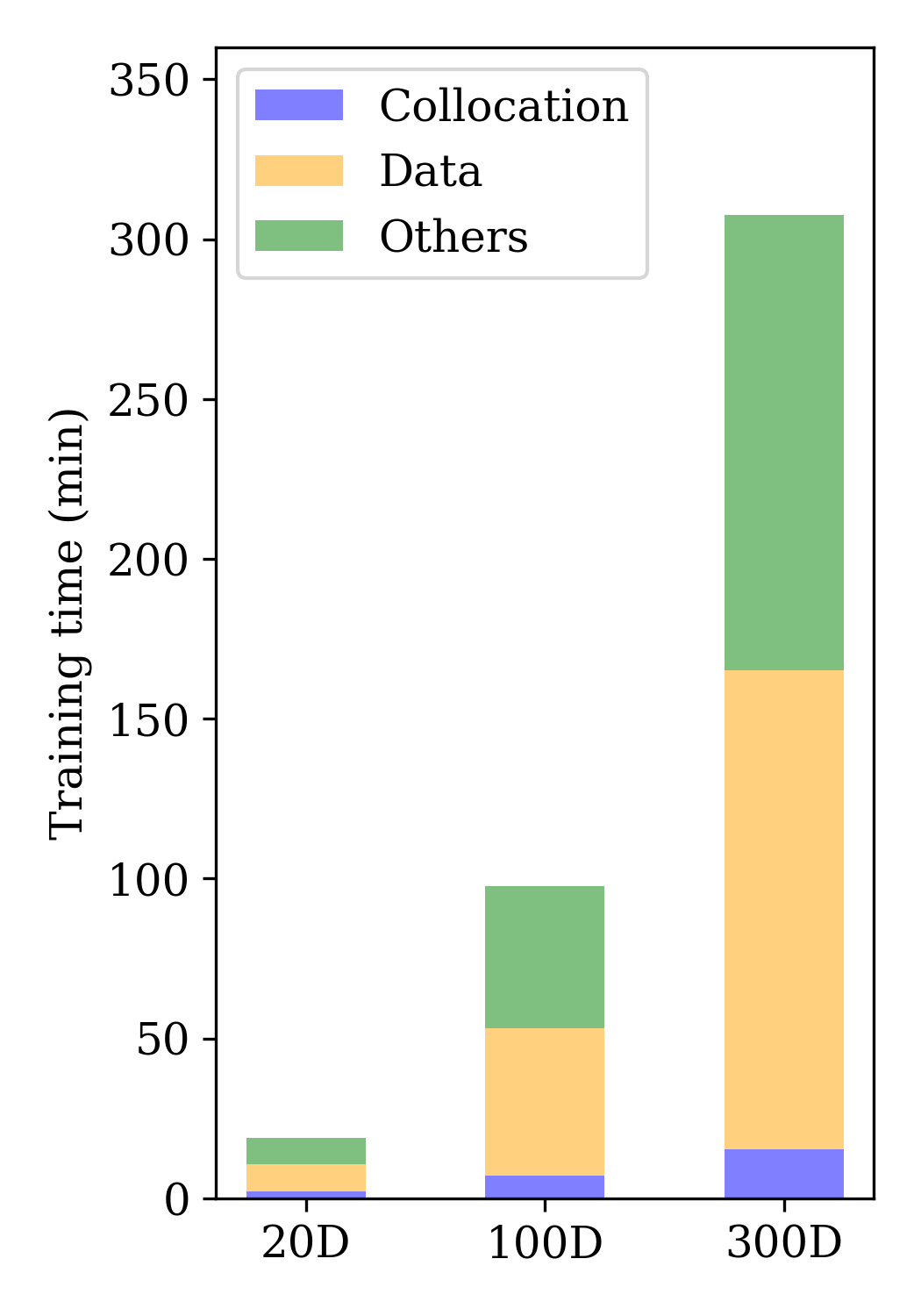}
}
\hspace{0.5cm}
\centering
{
\centering
\includegraphics[width=0.3\linewidth]{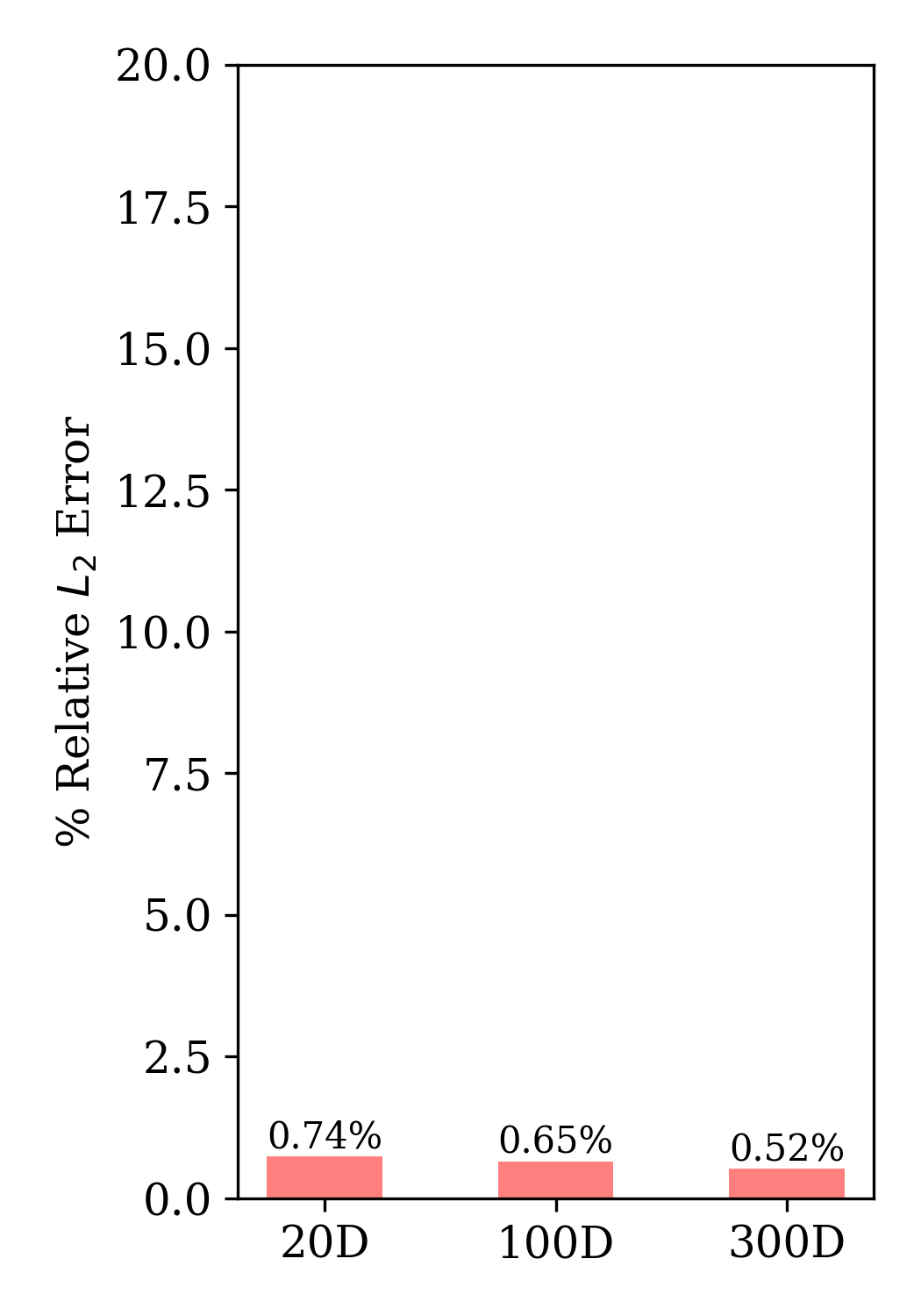}
}
\caption{Heat Equation: (Left) Training time profiling; (Right) mean testing error, for the 20D, 100D, and 300D cases with the additional time ($t$) dimension on T4 GPU.}
\label{fig:Heat_AnantNet_Acc_Rtime}
\end{figure}

\begin{figure}[h]
\centering
{
\centering
\includegraphics[width=\linewidth, trim={0cm, 1cm, 0cm, 1cm}]{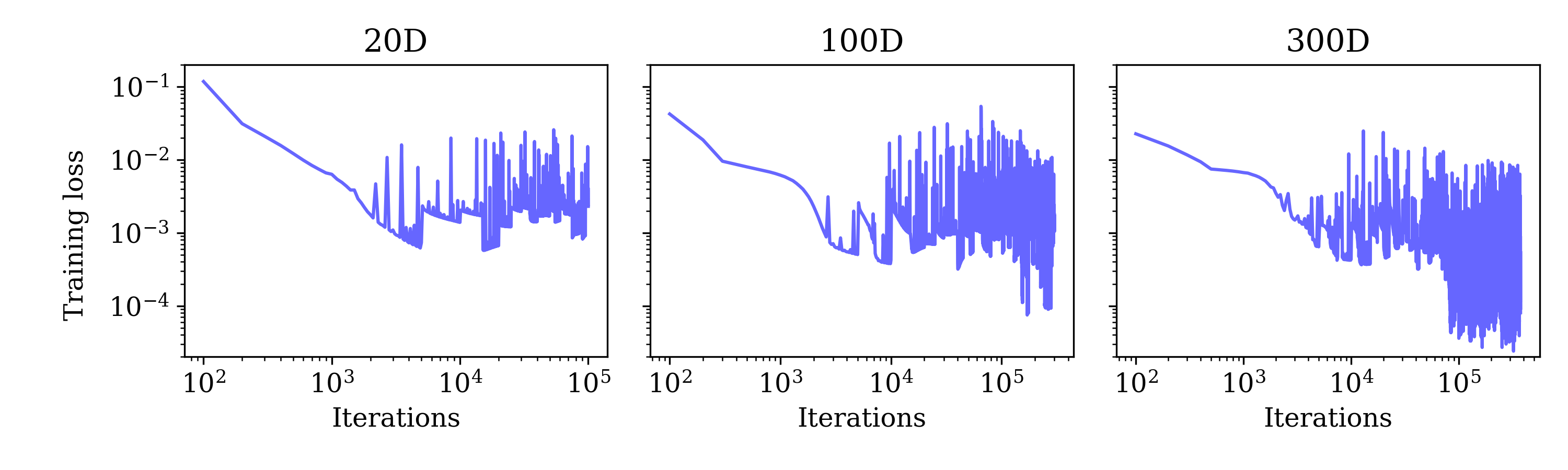}
}
\caption{Heat Equation: Training loss for the 20D, 100D, and 300D cases with the additional time ($t$) dimension on T4 GPU.}
\label{fig:Heat_AnantNet_Training_Loss}
\end{figure}
\noindent
In this case we employ layer-wise adaptive nonlinear activation function (sin) with trainable adaptive parameters for all body networks. We observe consistent scaling for run time and exceptional accuracy for Anant-Net while approximating the solution for the parabolic, high-dimensional, non-homogeneous heat equation for 20D, 100D and 300D. The performance of Anant-Net is shown in Figure \ref{fig:Heat_AnantNet_Acc_Rtime} and Table \ref{tab:arch-heat} provides more information about the architecture, number of collocation samples and data samples used for training. It is worthwhile to note that, unlike in the previous cases that dealt with a steady state high-dimensional PDE, here we deal with an unsteady high-dimensional PDE leading to the presence of the time derivative along with other spatial derivatives in the formulation of the partial differential equation. To this end, we assign an entire body network for learning the dynamics of the high-dimensional system or in other words, the time dimension ($t$) is always an active dimension for that body network. However, the other spatial dimensions ($x_i$) are distributed to the remaining body networks and these body networks will have simultaneous active and inactive dimensions at every iteration during the training process which are randomly sampled like in the previous sections. The convergence behavior for the heat equation for 20D, 100D and 300D are shown in Figure \ref{fig:Heat_AnantNet_Training_Loss}. More information about the architecture and hyperparameters are provided in APPENDIX \ref{appendix:b}.

\begin{figure}[H]
\centering
{
\centering
\includegraphics[width=0.3\linewidth, scale=0.45]{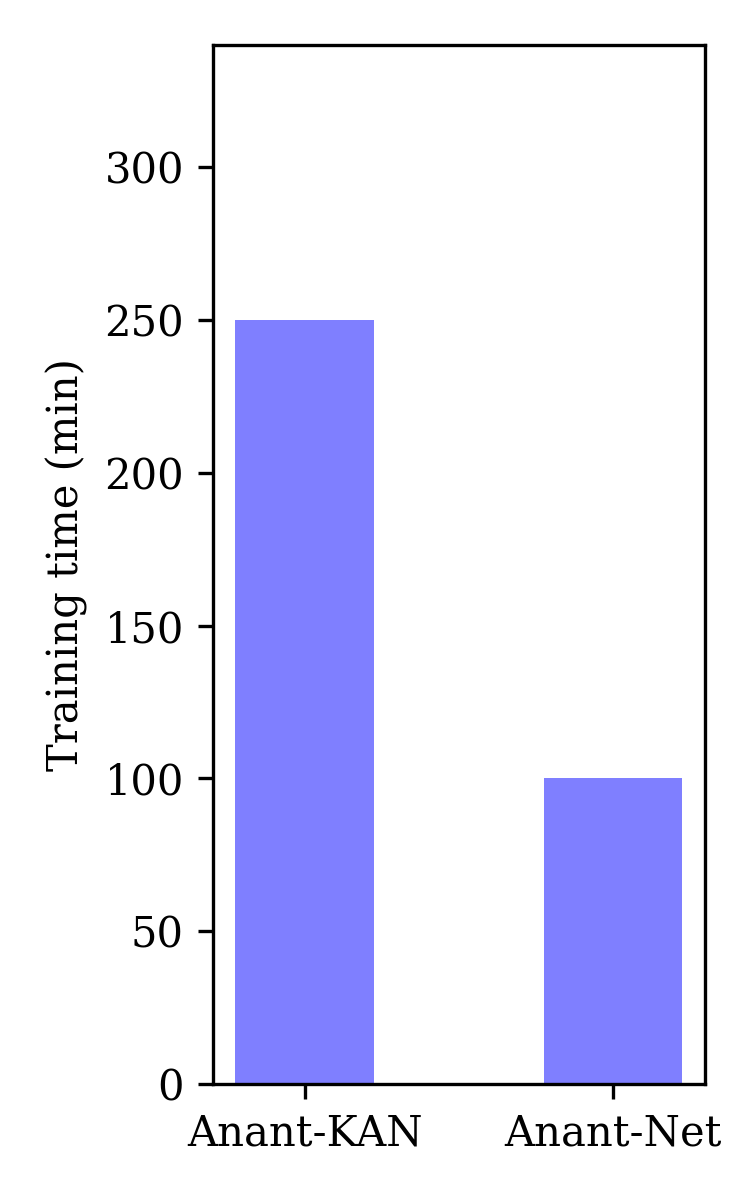}
}
\hspace{0.5cm}
\centering
{
\centering
\includegraphics[width=0.3\linewidth, scale=0.45]{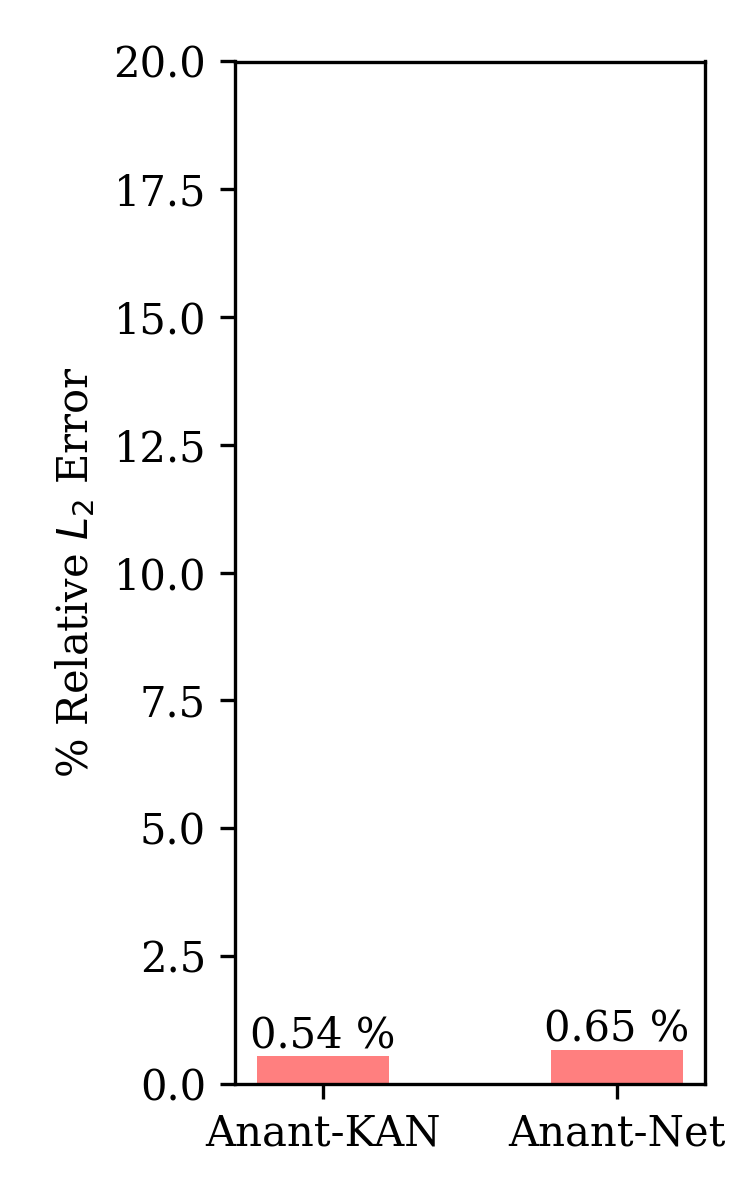}
}
\caption{Heat Equation: (Left) Training time profiling; (Right) mean testing error, for the 100D with the additional time ($t$) dimension for Anant-Net (on T4 GPU) and Anant-KAN (on High RAM CPU) architectures. For Anant-KAN, we use the spline basis with order of basis $(k) = 2$ and grid size $(G) = 5$.}
\label{fig:Heat_AKAN_ANET}
\end{figure}
\begin{figure}[H]
\centering
{
\includegraphics[trim = {5cm 5.5cm 5cm 3cm}, clip, scale=0.75]{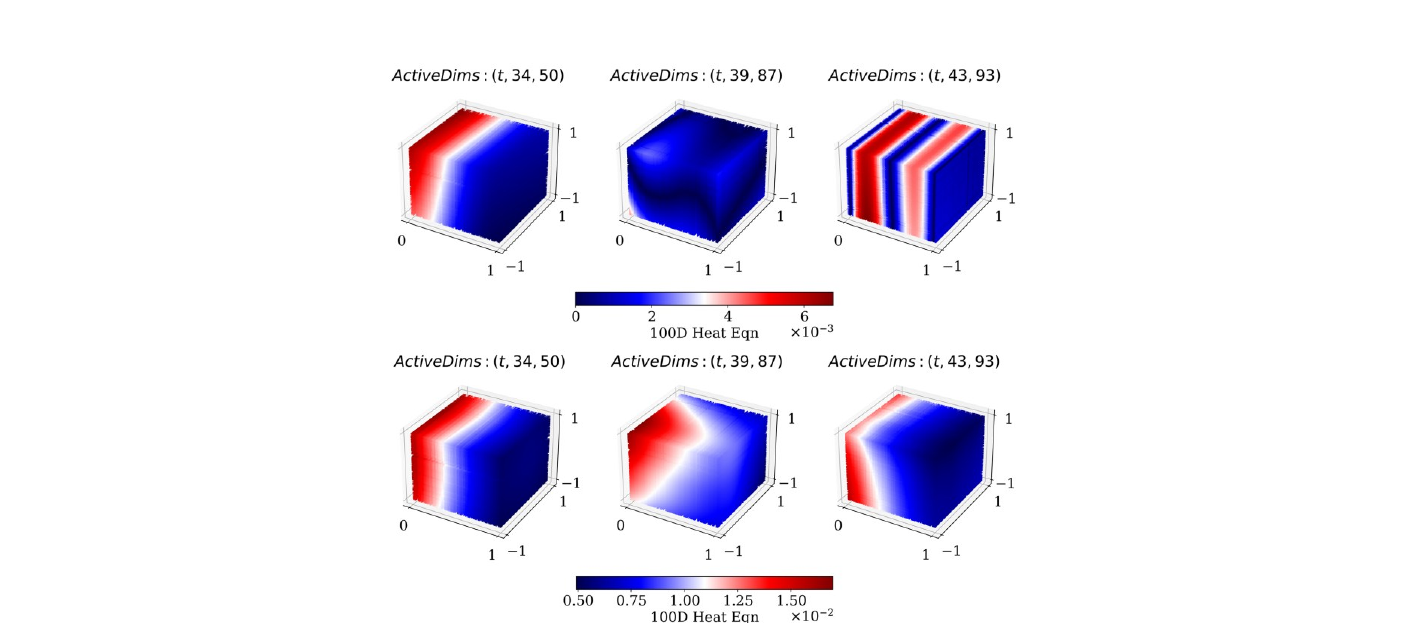}
}
\caption{Point-wise absolute testing error for random 3D hyper-surfaces of 100D (\textit{Top}) Anant-Net (\textit{Bottom}) Anant-KAN, for Heat equation. Here, "$\mathsf{ActiveDims}$: $(p, q, r)$" represents the active dimensions in the $d$-dimensional space where gradients are non-zero and $\mathrm{D} - \{p,q,r\}$ are inactive. Also, $\mathrm{D} = \{1,2,\dots,d\}$ is a set which constitutes all the dimensions. Since, the problems involves time derivative, a body network is dedicated towards learning the dynamics. Hence, $p = \text{time } (t)$ is fixed for all of the above cases.}  
\label{fig:AbsError_AnantNetKAN}
\end{figure}
\noindent
We also perform an empirical study for the current case using Anant-KAN in order to show its capability for solving high-dimensional parabolic PDEs with time derivatives. As it was the case for Anant-Net, we dedicate a body network of Anant-KAN exclusively for learning the dynamics of the high-dimensional PDE and time is always an active dimension for this body network. However, all the other dimensions in the high-dimensional space are distributed to rest of the body networks in Anant-KAN and the dimensions are simultaneously active or inactive for these body networks. Similar to Anant-Net, Anant-KAN also showcases good accuracy for 100D non-homogeneous heat equation as shown in Figure \ref{fig:Heat_AKAN_ANET}. Also note that Anant-KAN was run on a High-RAM CPU as opposed to the T4 GPU since the implementation of Kolmogorov Arnold Networks in the high-dimensional setting poses a major memory bottleneck. Although, currently we overcome this by switching to a high RAM CPU configuration, the implementation can be optimized further to overcome the memory bottleneck and this may be considered for a future work.

\subsection{Comparison}
In this section, we compare the testing performance and runtime of various methods. Specifically, we focus on methods that are designed to solve high-dimensional PDEs in hypercube domains with $ L > 1 $ employing boundary data sampling strategy. We report both the testing error and the runtime on GPU and CPU platforms.

\begin{table}[htbp]
  \centering
  \renewcommand{\arraystretch}{1.2}
  \begin{tabular}{||p{2cm}|c||c|c|c|c|c|c||}
    \hline
    \multirow{2}{5cm}{\textbf{Method }} &
    \multirow{2}{1.2cm}{\textbf{Device}} &
    \multicolumn{3}{c|}{\textbf{Run time (min)}} & \multicolumn{3}{c||}{\textbf{(\%) Relative $L_2$ Error}}\\
    \cline{3-8}
     &&\textbf{21D} & \textbf{99D} & \textbf{300D} & \textbf{21D} & \textbf{99D} & \textbf{300D}\\
    \hline
    WAN \cite{zang2020weak}   & CPU$^*$ & 250 & - & - & \textbf{0.9} $\pm$  \textbf{0.08} & NaN & NaN \\ \hline
    Anant-Net  & CPU$^*$ & 14 & 53 & 366 & 4.5 $\pm$ 0.037 & \textbf{8.6} $\pm$ \textbf{0.038} & \textbf{4.5} $\pm$ \textbf{0.037} \\ \hline \hline
        
    PINN & GPU$^\#$ & 34 & 100 & 250 & 16.5 $\pm$ 0.099 & 100.6 $\pm$ 0.90 & 174.5 $\pm$ 0.15 \\ \hline
    Anant-Net & GPU$^\#$ & 25 & 69 & 127 & \textbf{1.5} $\pm$ \textbf{0.013} & \textbf{7.1} $\pm$ \textbf{0.037} & \textbf{6.07} $\pm$ \textbf{0.038} \\ \hline

  \end{tabular}
  \caption{Run time and (\%) relative $L_2$ testing error on randomly selected test points for high-dimensional Poisson equation. ($^*$High RAM CPU with 51 GB RAM; $^\#$T4 GPU with 15 GB RAM). The best testing error is shown in \textbf{bold numbers}.}
  \label{tab:weak_strong_error_rtime}
\end{table}

\begin{table}[htbp]
  \centering
  \renewcommand{\arraystretch}{1.2}
  \begin{tabular}{||p{2cm}|c||c|c|c|c|c|c||}
    \hline
    \multirow{2}{5cm}{\textbf{Method }} &
    \multirow{2}{1.2cm}{\textbf{Device}} &
    \multicolumn{3}{c|}{\textbf{Run time (min)}} & \multicolumn{3}{c||}{\textbf{(\%) Relative $L_2$ Error}}\\
    \cline{3-8}
     &&\textbf{21D} & \textbf{99D} & \textbf{300D} & \textbf{21D} & \textbf{99D} & \textbf{300D}\\
    \hline
    WAN \cite{zang2020weak}  & CPU$^*$ & 167 & - & - & \textbf{1.21} $\pm$ \textbf{0.05} & NaN & NaN\\ \hline
    Anant-Net  & CPU$^*$ & 41 & 100 & 147 & 5.8 $\pm$ 0.030 & \textbf{7.4} $\pm$ \textbf{0.053} & \textbf{7.9} $\pm$\textbf{ 0.025} \\ \hline \hline 
    PINN  & GPU$^\#$ & 37 & 124 & 297 & 49.1 $\pm$ 0.32 & 35.3 $\pm$ 0.23 & 246.6 $\pm$ 0.35\\ \hline
    Anant-Net  & GPU$^\#$ & 53 & 108 & 168 & \textbf{2.7} $\pm$ \textbf{0.024}& \textbf{6.3} $\pm$ \textbf{0.052}& \textbf{6.7} $\pm$ \textbf{0.041} \\ \hline

  \end{tabular}
  \caption{Run time and (\%) relative $L_2$ error on randomly selected test points for high-dimensional sine-Gordon equation. ($^*$High RAM CPU with 51 GB RAM; $^\#$T4 GPU with 15 GB RAM). The best testing error is shown in \textbf{bold numbers}.}
  \label{tab:weak_strong_error_rtime2}
\end{table}
As shown in Table \ref{tab:weak_strong_error_rtime} and Table \ref{tab:weak_strong_error_rtime2}, WAN \cite{zang2020weak} demonstrates the efficiency of their method for problems with dimensions less than 50. For moderately high dimensions (e.g., 21D), WAN achieves better accuracy compared to Anant-Net, while Anant-Net attains significantly higher computational efficiency for the same volume of training data. Although WAN maintains stability and efficiency at lower dimensions, it still suffers from the curse of dimensionality at higher dimensions. For example, WAN becomes unstable at larger dimensions (99D and 300D), with the error diverging to NaN (not a number) during training. This instability is consistently observed regardless of the optimizer type, network architecture, or volume of training data. Furthermore, Table \ref{tab:weak_strong_error_rtime} and Table \ref{tab:weak_strong_error_rtime2} indicate that Anant-Net is both stable and computationally efficient at larger dimensions, whereas WAN is unstable and incurs high computational cost. The studies presented in Table \ref{tab:weak_strong_error_rtime} and Table \ref{tab:weak_strong_error_rtime2} were conducted to ensure a fair comparison, although the methods and architectures differ in various respects. For example, the number of parameters was kept approximately the same across all models despite differences in their architectures. The weak form based on WAN \cite{zang2020weak} was trained using the AdaGrad optimizer for both the primal and adversarial networks, as proposed in their paper. In contrast, the Anant-Net and PINN was trained using the Adam optimizer, with the pretrained model subsequently fine-tuned using the L-BFGS optimizer with appropriate learning rate selection. L-BFGS is a quasi-Newton optimizer that approximates the Hessian to avoid the high computational cost of direct computation for large problems. However, in non-convex settings, poor Hessian initialization can cause a mismatch between the true and approximated Hessians, leading to performance drops or optimizer divergence. Rafati et al. \cite{rafati2018improving} showed that batch size and Hessian initialization significantly influence L-BFGS performance, even though their study focused on the MNIST dataset. We believe these insights extend to solving high-dimensional PDEs. Thus, investigating batch size and initialization strategies for L-BFGS in the context of Anant-Net is a promising direction for future work.

\section{Conclusions}
High-dimensional partial differential equations (PDEs) arise in a broad range of scientific and engineering applications, including uncertainty quantification, plasma physics, quantitative finance, molecular dynamics, gas-kinetic theory, etc. The ability to efficiently solve such PDEs is important for advancing simulation, analysis, and informed decision-making in complex, real-world systems. In this work, we introduced Anant-Net, an efficient and scalable neural architecture tailored for solving high-dimensional PDEs defined over hypercubic domains. Traditional numerical and learning-based methods often struggle with the curse of dimensionality, especially when the side length of the domain $L > 1$, resulting in significant computational bottlenecks. Anant-Net overcomes these limitations through several key innovations, summarized below:
\begin{itemize}
    \item Anant-Net employs a tensor product formulation that enables efficient computation across batch dimensions. This design allows for dimension-wise sweeps over the full batch of collocation points, enabling the model to solve PDEs in hundreds of dimensions with high accuracy.
 \item To address the computational cost of automatic differentiation in high-dimensional spaces, Anant-Net adopts a selective differentiation approach. Instead of exhaustively computing derivatives in all dimensions, it samples a subset of dimensions (equal to batch-size) and applies differentiation only where needed. This significantly reduces computational overhead while preserving the essential structure of the governing PDEs.
\item  The model demonstrates the ability to solve both linear and nonlinear, steady state and transient  PDEs up to 300 dimensions within a few hours on a single GPU. We report the mean and standard deviation (variance) of relative $L_2$ testing error on randomly selected test points from the exponentially increasing high-dimensional spaces.
Furthermore, the accuracy can be enhanced by increasing the number of training and collocation points.
\item  We also propose an interpretable variant of Anant-Net, based on the Kolmogorov–Arnold Network (KAN) for high-dimensional PDEs. This variant offers deeper insights into the learned solution structure and enhances interpretability in deep neural networks.
\end{itemize}
For extremely high-dimensional problems where the data volume exceeds single-GPU memory, Anant-Net supports efficient training through mini-batching. Moreover, it can be easily adapted to parallelization techniques such as data and model parallelism \cite{shukla2021parallel}, tensor parallelism \cite{hu2024tackling}, and domain decomposition \cite{jagtap2020conservative,jagtap2020extended,moseley2023finite}, enabling deployment across multi-GPU architectures. Anant-Net presents a promising architecture for solving high-dimensional problems with both high accuracy and computational efficiency. 

Some of the known limitations of Anant-Net are as follows: Anant-Net is primarily designed for high-dimensional PDEs in Euclidean domains where a tensor product structure is present, enabling efficient computations with complexity scaling as $\mathcal{O}(Nd)$. This structure allows straightforward handling of linear additive operators such as the Laplacian, where contributions from individual dimensions are independent and additive. However, for more complex operators, like variable-coefficient diffusion $\nabla \cdot (k(x) \nabla u)$ or nonlinear advection $u \cdot \nabla u$, dimensional interactions become nontrivial due to cross-derivatives and nonlinear couplings. In such cases, decomposing the operator across active dimensions is less straightforward. While Anant-Net offers architectural flexibility to incorporate variable coefficients as inputs or represent nonlinear terms using network outputs and their gradients, doing so increases model complexity and requires careful redesign of the current architecture. These extensions remain the subject of ongoing work. Moreover, Anant-Net could also be extended to solve high-dimensional PDEs at selected points or along trajectories in the space rather than approximating the entire domain. To this end, we emphasize that solving PDEs in high dimensions over exponentially increasing domains is inherently difficult due to the curse of dimensionality, and Anant-Net aims to mitigate this challenge through a structured, scalable framework that lays the foundation for further generalizations.

\section*{Acknowledgement}
We would like to thank all the reviewers for their valuable feedback and constructive suggestions, which have helped improve the quality and clarity of this work.

\bibliography{Ref}
\bibliographystyle{elsarticle-num}

\pagebreak

\appendix
\section{Universal Approximation Theorem: Through the lens of Anant-Net}
\label{appendix:a}
\begin{theorem}
\label{Th:Theorem1}
(\text{SPINN}\cite{cho2024separable}) Let X, Y be compact subsets of $\mathbb{R}^d$. With $u \in L^2(X \times Y).$ Then, for arbitrary $\varepsilon > 0$, we can find a sufficiently large $\tilde{r} > 0$ and neural networks $f_j$ and $g_j$ such that,
\begin{equation}
\biggl\|u - \sum_{j=1}^{\tilde{r}}f_jg_j\biggr\|_{L^2(X \times Y)} < \varepsilon    
\end{equation}
\end{theorem}
\noindent
Theorem \ref{Th:Theorem1} was the universal approximation theorem proposed by Cho et al.\cite{cho2024separable} in their work. However, as explained in the previous sections of this paper, naively implementing this approach for a high-dimensional partial differential equation is infeasible. To this end, we propose Anant-Net and the corresponding universal approximation theorem for Anant-Net, Theorem \ref{Th:Theorem2}.

\begin{theorem}
\label{Th:Theorem2}
(Anant-Net) Let X, Y be compact subsets of $\mathbb{R}^d$. With $u^k \in L^2(X \times Y).$ Then, for arbitrary $\varepsilon > 0$, we can find a sufficiently large $r > 0$ and neural networks $f_j \triangleq f_{1,j}^{(\theta_1)}$ and $g_j \triangleq f_{2,j}^{(\theta_2)}$ such that,
\begin{equation}
\frac{1}{|\hat{\mathcal{B}}|}\sum_{k=1}^{|\hat{\mathcal{B}}|}\biggl\|u^k - \sum_{j=1}^{r}f_jg_j\biggr\|_{L^2(X \times Y)} < \varepsilon 
\end{equation}
\begin{equation}
f_j = f_j(\hat{\mathcal{B}}_k^1,X\text{\textbackslash}\hat{\mathcal{B}}_k^1) \hspace{1cm}\text{and}\hspace{1cm} g_j = g_j(\hat{\mathcal{B}}_k^2, Y\text{\textbackslash}\hat{\mathcal{B}}_k^2) 
\end{equation}
\\
where, $\hat{\mathcal{B}} = \{\hat{\mathcal{B}}_1,\hat{\mathcal{B}}_2, \dots \hat{\mathcal{B}}_k\}$ is the set of sets consisting of active dimensions where at any given step-$k$, $\hat{\mathcal{B}}_k = \{\hat{\mathcal{B}}_k^1, \hat{\mathcal{B}}_k^2\} = \{x_p, x_q\}$ is the set of active dimensions, $\{X\text{\textbackslash}\hat{\mathcal{B}}_k^1\}$ and $\{Y\text{\textbackslash}\hat{\mathcal{B}}_k^2\}$ is the set of inactive dimensions. Here, $X = \{x_1,x_2,\dots x_p \dots x_{d/2}\}$, $Y = \{x_{d/2+1},\dots x_q \dots x_d\}.$ 

\noindent
While Theorem \ref{Th:Theorem1} learns to approximate the solution, $u$ in the $d$-dimensional space with an equivalent number of $d$-body networks, this is often intractable for very high-dimensional problems. On the other hand, Theorem \ref{Th:Theorem2} proposes the model to approximate multiple slices of the $d$-dimensional space with active gradients only in selected dimensions leading to $|\hat{\mathcal{B}_k}|$-body networks with $|\hat{\mathcal{B}_k}|<<d$. However, in this scenario, model learns to approximate solution in the $d$-dimensional space in $k$-steps where $u^k$ is the solution and $\hat{\mathcal{B}}_k$ is the set of active dimensions at step-$k$.

\end{theorem}

\noindent

\textit{Proof.} Let ${\phi_i}$ and ${\psi_j}$ be the orthonormal basis for $L^2(X)$ and $L^2(Y)$ respectively. Then ${\phi_i \psi_j}$ forms the orthonormal basis for $L^2(X\times Y)$. Therefore, we can find a sufficiently large r, consider an active dimension set $\mathcal{B}_k$ that results in the solution $u^k$ such that,
\begin{equation}
\biggl\|u^k - \sum_{i,j}^{r}a^k_{ij}\phi_i\psi_j\biggr\|_{L^2(X \times Y)} < \frac{\varepsilon}{2}    
\end{equation}
where $a^k_{ij}$ denotes
\begin{equation}
a^k_{ij} = \int_{X\times Y} u^k(x,y)\phi_i(x)\psi_j(y)dxdy 
\end{equation}
On the other hand, by the universal approximation theorem, we can find neural networks $f_i$ and $g_j$ such that,
\begin{equation}
\|\phi_i - f_j \|_{L^2(X)} \leq \frac{\varepsilon}{3^j\|u^k\|_{L^2(X \times Y)}} \text{ and } \|\psi_j - g_j \|_{L^2(Y)} \leq \frac{\varepsilon}{3^j\|u^k\|_{L^2(X \times Y)}} 
\end{equation}
We consider the difference between $u^k$ and $\sum_{i,j}^{r} a^k_{ij}f_ig_j$ by adding and subtracting $\sum_{i,j}^{r} a^k_{ij}\phi_i\psi_j$ as shown below: 

$$\biggl\|u^k - \sum_{i,j}^ra^k_{ij}f_ig_j\biggr\|_{L^2(X\times Y)} = \biggl\|u^k - \sum_{i,j}^ra^k_{ij}f_ig_j + \sum_{i,j}^ra^k_{ij}\phi_i\psi_j - \sum_{i,j}^ra^k_{ij}\phi_i\psi_j\biggr\|_{L^2(X\times Y)}$$
\\
Appropriately grouping the terms and applying the triangular inequality $\|p + q\| \leq \|p\| + \|q\|$, we get the following.
$$\biggl\|u^k - \sum_{i,j}^ra^k_{ij}f_ig_j\biggr\|_{L^2(X\times Y)} \leq \biggl\|u^k - \sum_{i,j}^ra^k_{ij}\phi_i\psi_j\biggr\|_{L^2(X\times Y)} + \biggl\|\sum_{i,j}^ra^k_{ij}\phi_i\psi_j - \sum_{i,j}^ra^k_{ij}f_ig_j\biggr\|_{L^2(X\times Y)}$$
But from equation, equation can be further simplified to the following form.
$$\biggl\|u^k - \sum_{i,j}^ra^k_{ij}f_ig_j\biggr\|_{L^2(X\times Y)} \leq \frac{\varepsilon}{2} + \biggl\|\sum_{i,j}^ra^k_{ij}\phi_i\psi_j - \sum_{i,j}^ra^k_{ij}f_ig_j\biggr\|_{L^2(X\times Y)}$$
Now we go ahead to quantify the below unknown to complete the proof.
$$\biggl\|\sum_{i,j}^ra^k_{ij}\phi_i\psi_j - \sum_{i,j}^ra^k_{ij}f_ig_j\biggr\|_{L^2(X\times Y)}$$
\\
Consider the terms inside the parenthesis as follows,
$$\sum_{i,j}^ra^k_{ij}\phi_i\psi_j - \sum_{i,j}^ra^k_{ij}f_ig_j = \sum_{i,j}^ra^k_{ij}\phi_i(\psi_j - g_j) - \sum_{i,j}^ra^k_{ij}(\phi_i - f_i)g_j = II_1 + II_2 = II$$
\\
We first compute $II_1$.
$$
\|II_1\|^2_{L^2(X\times Y)} = \int_{X\times Y}\biggl\{\sum_{i,j}^ra_{i,j}^k\phi_i(\psi_j - f_j) \biggr\}^2 dxdy
$$
$$
\hspace{2.1cm} = \int_{X\times Y}\biggl\{\sum_j^r\biggl(\sum_i^ra_{i,j}^k\phi_i\biggr)(\psi_j - f_j) \biggr\}^2 dxdy
$$
$$
\hspace{0.6cm} = \int_{X\times Y}\biggl\{\sum_j^rA_j(x)B_j(y)\biggr\}^2 dxdy
$$
We apply Cauchy-Scharwz inequality, $|\langle p ,q\rangle| \leq \|p\|\cdot\|q\|$ to further break-down the above expression as follows.
$$
\|II_1\|^2_{L^2(X\times Y)} \leq \biggl(\int_{X}\sum_j^r|A_j(x)|^2dx\biggr)\biggl(\int_{Y}\sum_j^r|B_j(y)|^2dy\biggr)
$$
Then,
$$
\int_{X}\sum_j^r|A_j(x)|^2dx = \int_{X}\sum_j^r\biggl(\sum_i^ra^k_{i,j}\phi_i\biggr)^2dx
$$
$$
\hspace{2.2cm} = \sum_j^r\biggl\|\sum_i^ra^k_{i,j}\phi_i\biggr\|^2_{L(X)}
$$
$$
\hspace{2cm} = \sum_j^r\biggl(\sum_i^r|a^k_{i,j}|\biggr)^2
$$
$$
\hspace{2cm} \leq 2\sum_j^r\sum_i^r|a^k_{i,j}|^2 < 2\sum_{i,j}^{\infty}|a^k_{i,j}|^2 = 2\|u^k\|^2_{L^2(X\times Y)}
$$
On the other hand, we have the following form.
$$
\int_{Y}\sum_j^r|B_j(y)|^2dy = \sum_j^r\int_{Y}|\psi_j(y) - g_j(y)|^2dy
$$
$$
\hspace{2cm} = \sum_j^r\|\psi_j(y) - g_j(y)\|^2_{L^2(Y)}
$$
$$
\hspace{2cm} \leq \sum_j^r\frac{\varepsilon^2}{9^j \|u^k\|^2_{L^2(X \times Y)}}
$$
$$
\hspace{2cm} < \frac{\varepsilon^2}{8 \|u^k\|^2_{L^2(X \times Y)}}
$$
Hence we have.
$$
\|II_1\|^2_{L^2(X\times Y)} < \frac{\varepsilon^2}{8 \|u^k\|^2_{L^2(X \times Y)}}2\|u^k\|^2_{L^2{(X\times Y)}} = \frac{\varepsilon^2}{4}
$$
Likewise,
$$
\|II_2\|^2_{L^2(X\times Y)} < \frac{\varepsilon^2}{4}
$$
Therefore,
$$
\|II\|^2_{L^2(X\times Y)} < \frac{\varepsilon^2}{2}
$$
We go back to with estimates,
$$
\biggl\|u^k - \sum_{i,j}^ra^k_{i,j}f_ig_j\biggr\|_{L_2(X \times Y)} < \frac{\varepsilon}{2} + \frac{\varepsilon}{2} = \varepsilon
$$
\\
However, we know that the approximation error is an average of the error across all the sets of active dimensions and accounting for $k$ sets of active dimensions results in the following.
$$
\sum_{k=1}^{|\hat{\mathcal{B}}|}\biggl\|u^k - \sum_{i,j}^ra^k_{i,j}f_ig_j\biggr\|_{L_2(X \times Y)} < \sum_{k=1}^{|\hat{\mathcal{B}}|}\varepsilon = |\hat{\mathcal{B}}|\cdot\varepsilon 
$$
$$
\frac{1}{|\hat{\mathcal{B}}|}\sum_{k=1}^{|\hat{\mathcal{B}}|}\biggl\|u^k - \sum_{i,j}^ra^k_{i,j}f_ig_j\biggr\|_{L_2(X \times Y)} < \varepsilon
$$
\\
Since, the summation of the product of $f_i$ and $g_j$ is performed element-wise, it is sensible to reorder the notations as shown below.
$$
\frac{1}{|\hat{\mathcal{B}}|}\sum_{k=1}^{|\hat{\mathcal{B}}|}\biggl\|u^k - \sum_{i}^{\tilde{r}}b_i\tilde{f_i}\tilde{g_i}\biggr\|_{L_2(X \times Y)} < \varepsilon
$$
where, $b_i$ is a constant and grouping it with $\tilde{f_i}$ gives the following.
$$
\frac{1}{|\hat{\mathcal{B}}|}\sum_{k=1}^{|\hat{\mathcal{B}}|}\biggl\|u^k - \sum_{i}^{\tilde{r}}\{b_i\tilde{f_i}\}\tilde{g_i}\biggr\|_{L_2(X \times Y)} = \frac{1}{|\hat{\mathcal{B}}|}\sum_{k=1}^{|\hat{\mathcal{B}}|}\biggl\|u^k - \sum_{i}^{\tilde{r}}t_i\tilde{g_i}\biggr\|_{L_2(X \times Y)} < \varepsilon
$$
\\
Replacing, $\tilde{r}$, $\tilde{g_i}$, $t_i$ with $r$, $g_i$, $f_i$ results in the following form.
$$
\frac{1}{|\hat{\mathcal{B}}|}\sum_{k=1}^{|\hat{\mathcal{B}}|}\biggl\|u^k - \sum_{i}^rf_ig_i\biggr\|_{L_2(X \times Y)} < \varepsilon 
$$
Hence proved $\blacksquare$ 

\section{Hyperparameters for Anant-Net}
\label{appendix:b}
The main hyperparameters for Anant-Net include: the number of body networks ($B$), grid resolution for collocation points ($N_C$), grid resolution for boundary/data points ($N_B$), number of collocation grids ($\#CG$), number of boundary/data grids ($\#BG$), and the sampling frequency ($S_{\text{FREQ}}$). The hyperparameters used for the main results presented in this paper are discussed in this section. For clarity, the tables presented below follow the format: $\#$Collocation $= (N_C \times N_C \times N_C)\times\#CG$ and $\#$Data $= (N_B \times N_B \times N_B)\times\#BG$ for the current cases dealing with 3 body networks. However, if any future work deals with 4 body networks, then $\#$Collocation $= (N_C \times N_C \times N_C \times N_C)\times\#CG$ and $\#$Data $= (N_B \times N_B \times N_B \times N_B)\times\#BG$ and likewise.
\\
\subsubsection*{Hyperparameters for Section \ref{sec:Poisson}}
\begin{table}[H]
\begin{center}
\scriptsize
\begin{tabular}{||c | c | c | c | c ||}  
 \hline
 Dimension & Architecture & Optimizer & Sampling Frequency & $\#$Iterations \\ [0.05ex] 
 \hline\hline
    \multirow{2}{*}{21D} & \multirow{2}{*}{$\bigl[7, 64, 64, 10\bigr] \times 3$} & Adam & 5000 & 70000\\
    \cline{3-5}
     & & L-BFGS & 5000 & 70000\\
    \hline\hline
    \multirow{2}{*}{99D} & \multirow{2}{*}{$\bigl[33, 64, 64, 40\bigr] \times 3$} & Adam & 5000 & 165000\\
    \cline{3-5}
     & & L-BFGS & 5000 & 165000\\
    \hline\hline
    \multirow{2}{*}{300D} & \multirow{2}{*}{$\bigl[100, 128, 128, 128\bigr] \times 3$} & Adam & 500 & 50000\\
    \cline{3-5}
     & & L-BFGS & 2000 & 200000\\
    \hline
\end{tabular}
\end{center}
\caption{Anant-Net architecture used for solving high-dimensional Poisson equation.}
\label{tab:arch-poisson1}
\end{table}

\begin{table}[H]
\begin{center}
\scriptsize
\begin{tabular}{||c | c | c||}  
 \hline
 Dimension & $\#$Collocation & $\#$Data \\ [0.05ex] 
 \hline\hline
 21D & $(14 \times 14 \times 14) \times 14 = 38416$ & $(6 \times 6 \times 6) \times 32 = 6912$\\ 
 \hline
 99D & $(14 \times 14 \times 14) \times 33 = 90552$ & $(6 \times 6 \times 6) \times 50 = 10800$\\
 \hline
 300D & $(14 \times 14 \times 14) \times 100 = 274400$ & $(6 \times 6 \times 6) \times 150 = 32400$\\
 \hline
\end{tabular}
\end{center}
\caption{Details on collocation and data points sampled for cases in Table \ref{tab:arch-poisson1} for solving high-dimensional Poisson equation.}
\label{tab:arch-poisson2}
\end{table}

\noindent
\subsubsection*{Hyperparameters for Section \ref{sec:Sine-Gordon}}
\begin{table}[H]
\begin{center}
\scriptsize
\begin{tabular}{||c | c | c | c | c ||}  
 \hline
 Dimension & Architecture & Optimizer & Sampling Frequency & $\#$Iterations \\ [0.05ex] 
 \hline\hline
    \multirow{2}{*}{21D} & \multirow{2}{*}{$\bigl[7, 64, 64, 10\bigr] \times 3$} & Adam & 5000 & 100000\\
    \cline{3-5}
     & & L-BFGS & 5000 & 100000\\
    \hline\hline
    \multirow{2}{*}{99D} & \multirow{2}{*}{$\bigl[33, 64, 64, 40\bigr] \times 3$} & Adam & 5000 & 200000\\
    \cline{3-5}
     & & L-BFGS & 5000 & 200000\\
    \hline\hline
    \multirow{2}{*}{300D} & \multirow{2}{*}{$\bigl[100, 64, 64, 128\bigr] \times 3$} & Adam & 1000 & 110000\\
    \cline{3-5}
     & & L-BFGS & 2000 & 140000\\
    \hline
\end{tabular}
\end{center}
\caption{Anant-Net architecture used for solving high-dimensional sine-Gordon equation.}
\label{tab:arch-sinegordon1}
\end{table}

\begin{table}[H]
\begin{center}
\scriptsize
\begin{tabular}{||c | c | c||}  
 \hline
 Dimension & $\#$Collocation & $\#$Data \\ [0.05ex] 
 \hline\hline
 21D & $(14 \times 14 \times 14) \times 20 = 54880$ & $(6 \times 6 \times 6) \times 60 = 12960$\\ 
 \hline
 99D & $(14 \times 14 \times 14) \times 40 = 109760$ & $(6 \times 6 \times 6) \times 65 = 14040$\\
 \hline
 300D & $(14 \times 14 \times 14) \times 110 = 301840$ & $(6 \times 6 \times 6) \times 160 = 34560$\\
 \hline
\end{tabular}
\end{center}
\caption{Details on collocation and data points sampled for cases in Table \ref{tab:arch-sinegordon1} for solving high-dimensional sine-Gordon equation.}
\label{tab:arch-sinegordon2}
\end{table}

\noindent
\subsubsection*{Hyperparameters for Section \ref{sec:Allen-Cahn}}
\begin{table}[H]
\begin{center}
\scriptsize
\begin{tabular}{||c | c | c | c | c ||}  
 \hline
 Dimension & Architecture & Optimizer & Sampling Frequency & $\#$Iterations \\ [0.05ex] 
 \hline\hline
    \multirow{2}{*}{21D} & \multirow{2}{*}{$\bigl[7, 64, 64, 10\bigr] \times 3$} & Adam & 5000 & 70000\\
    \cline{3-5}
     & & L-BFGS & 5000 & 70000\\
    \hline\hline
    \multirow{2}{*}{99D} & \multirow{2}{*}{$\bigl[33, 64, 64, 40\bigr] \times 3$} & Adam & 3000 & 99000\\
    \cline{3-5}
     & & L-BFGS & 5000 & 165000\\
    \hline\hline
    \multirow{2}{*}{300D} & \multirow{2}{*}{$\bigl[100, 64, 64, 128\bigr] \times 3$} & Adam & 500 & 50000\\
    \cline{3-5}
     & & L-BFGS & 1000 & 200000\\
    \hline
\end{tabular}
\end{center}
\caption{Anant-Net architecture used for solving high-dimensional Allen-Cahn equation.}
\label{tab:arch-allencahn1}
\end{table}

\begin{table}[H]
\begin{center}
\scriptsize
\begin{tabular}{||c | c | c||}  
 \hline
 Dimension & $\#$Collocation & $\#$Data \\ [0.05ex] 
 \hline\hline
 21D & $(14 \times 14 \times 14) \times 14 = 38416$ & $(6 \times 6 \times 6) \times 40 = 8640$\\ 
 \hline
 99D & $(20 \times 20 \times 20) \times 33 = 264000$ & $(9 \times 9 \times 9) \times 50 = 36450$\\
 \hline
 300D & $(14 \times 14 \times 14) \times 100 = 274400$ & $(6 \times 6 \times 6) \times 200 = 43200$\\
 \hline
\end{tabular}
\end{center}
\caption{Details on collocation and data points sampled for cases in Table \ref{tab:arch-allencahn1} for solving high-dimensional Allen-Cahn equation.}
\label{tab:arch-allencahn2}
\end{table}

\noindent
\subsubsection*{Hyperparameters for Section \ref{sec:Anant-KAN}}

\begin{table}[H]
\begin{center}
\scriptsize
\begin{tabular}{||c | c | c | c | c | c ||}  
 \hline
 Equation & Dimension & Architecture & Optimizer & Sampling Frequency & $\#$Iterations \\ [0.05ex] 
 \hline\hline
    \multirow{2}{*}{Poisson}
    & 21D & $\bigl[7, 5, 5, 10\bigr] \times 3$ & Adam & 5000 & 70000\\
    \cline{2-6}
     & 99D & $\bigl[33, 5, 5, 14\bigr] \times 3$ & Adam & 2000 & 80000\\
    \hline\hline
    \multirow{2}{*}{sine-Gordon}     
     & 21D & $\bigl[7, 5, 5, 10\bigr] \times 3$ & Adam & 5000 & 70000\\
    \cline{2-6}
     & 99D & $\bigl[33, 5, 5, 14\bigr] \times 3$ & Adam & 2000 & 80000\\
    \hline\hline
    \multirow{2}{*}{Allen-Cahn}     
     & 21D & $\bigl[7, 5, 5, 10\bigr] \times 3$ & Adam & 5000 & 70000\\
    \cline{2-6}
     & 99D & $\bigl[33, 5, 5, 14\bigr] \times 3$ & Adam & 2000 & 80000\\
    \hline
\end{tabular}
\end{center}
\caption{Anant-KAN architecture used for solving high-dimensional Poisson equation, sine-Gordon equation and Allen-Cahn equation. Note that Spline basis is used for all the cases with polynomial order ($k$) = 2 and grid size ($G$) = 5.}
\label{tab:arch-Akan1}
\end{table}

\begin{table}[H]
\begin{center}
\scriptsize
\begin{tabular}{||c | c | c | c ||}  
 \hline
 Equation & Dimension & $\#$Collocation & $\#$Data \\ [0.05ex] 
 \hline\hline
    \multirow{2}{*}{Poisson}
     & 21D & $(14 \times 14 \times 14) \times 14 = 38416$ & $(6 \times 6 \times 6) \times 32 = 6912$\\
    \cline{2-4}
     & 99D & $(14 \times 14 \times 14) \times 40 = 109760$ & $(6 \times 6 \times 6) \times 65 = 14040$\\
    \hline\hline
    \multirow{2}{*}{sine-Gordon}     
     & 21D & $(14 \times 14 \times 14) \times 14 = 38416$ & $(6 \times 6 \times 6) \times 32 = 6912$\\
    \cline{2-4}
     & 99D & $(14 \times 14 \times 14) \times 40 = 109760$ & $(6 \times 6 \times 6) \times 50 = 10800$\\
    \hline\hline
    \multirow{2}{*}{Allen-Cahn}     
     & 21D & $(14 \times 14 \times 14) \times 14 = 38416$ & $(6 \times 6 \times 6) \times 32 = 6912$\\
    \cline{2-4}
     & 99D & $(14 \times 14 \times 14) \times 40 = 109760$ & $(6 \times 6 \times 6) \times 65 = 14040$\\
    \hline
\end{tabular}
\end{center}
\caption{Details on collocation and data points sampled for cases in Table \ref{tab:arch-Akan1} for solving high-dimensional Poisson equation, sine-Gordon equation and Allen-Cahn equation.}
\label{tab:arch-Akan2}
\end{table}

\noindent
{\subsubsection*{Hyperparameters for Section \ref{sec:Heat}}

\begin{table}[H]
\begin{center}
\scriptsize
\begin{tabular}{||c | c | c | c | c ||}  
 \hline
 Dimension & Architecture & Optimizer & Sampling Frequency & $\#$Iterations \\ [0.05ex] 
 \hline\hline
    20D & $\bigl[1, 64, 64, 15\bigr] \times 1, \bigl[10, 64, 64, 15\bigr] \times 2$ & Adam & 5000 & 100000\\
    \hline\hline
    100D & $\bigl[1, 64, 64, 50\bigr] \times 1, \bigl[50, 64, 64, 50\bigr] \times 2$ & Adam & 5000 & 300000\\
    \hline\hline
    300D & $\bigl[1, 64, 64, 160\bigr] \times 1, \bigl[150, 64, 64, 160\bigr] \times 2$ & Adam & 2500 & 375000\\
    \hline
\end{tabular}
\end{center}
\caption{Anant-Net architecture used for solving high-dimensional Heat equation.}
\label{tab:arch-heat111}
\end{table}

\begin{table}[H]
\begin{center}
\scriptsize
\begin{tabular}{||c | c | c||}  
 \hline
 Dimension & $\#$Collocation & $\#$Data \\ [0.05ex] 
 \hline\hline
 20D & $(14 \times 14 \times 14) \times 20 = 54880$ & $(10 \times 10 \times 10) \times 50 = 50000$\\ 
 \hline
 100D & $(14 \times 14 \times 14) \times 60 = 164640$ & $(10 \times 10 \times 10) \times 90 = 90000$\\
 \hline
 300D & $(14 \times 14 \times 14) \times 150 = 411600$ & $(10 \times 10 \times 10) \times 160 = 160000$\\
 \hline
\end{tabular}
\end{center}
\caption{Details on collocation and data points sampled for cases in Table \ref{tab:arch-heat111} for solving high-dimensional Heat equation.}
\label{tab:arch-heat112}
\end{table}

\begin{table}[H]
\begin{center}
\scriptsize
\begin{tabular}{||c | c | c | c | c ||}  
 \hline
 Dimension & Architecture & Optimizer & Sampling Frequency & $\#$Iterations \\ [0.05ex] 
 \hline\hline
    100D & $\bigl[1, 5, 5, 18\bigr] \times 1, \bigl[50, 5, 5, 18\bigr] \times 2$ & Adam & 1000 & 55000\\
    \hline
\end{tabular}
\end{center}
\caption{Anant-KAN architecture used for solving high-dimensional Heat equation. Note that Spline basis is used for all the cases with polynomial order ($k$) = 2 and grid size ($G$) = 5.}
\label{tab:arch-heatKAN111}
\end{table}

\begin{table}[H]
\begin{center}
\scriptsize
\begin{tabular}{||c | c | c||}  
 \hline
 Dimension & $\#$Collocation & $\#$Data \\ [0.05ex] 
 \hline\hline
 100D & $(14 \times 14 \times 14) \times 50 = 137200$ & $(10 \times 10 \times 10) \times 60 = 60000$\\
 \hline
\end{tabular}
\end{center}
\caption{Details on collocation and data points sampled for cases in Table \ref{tab:arch-heatKAN111} for solving high-dimensional Heat equation.}
\label{tab:arch-heatKAN112}
\end{table}}

\section{Nomenclature}
\label{appendix:c}
\begin{table}[H]
\begin{center}
\scriptsize
\label{symb-tables}
\begin{tabular}{ | m{5em} | m{5cm} | } 
  \hline
  Symbol & Definition \\ 
  \hline
  $\mathcal{L}$ & Differential operator for PDE \\
  \hline
  $\mathcal{B}$ & Boundary operator \\
  \hline
  $f$ & Source term of the PDE \\
  \hline
  $g$ & Boundary source term of the PDE \\
  \hline
  $u$ & Exact solution of the PDE \\
  \hline
  $\hat{u}_{\theta}$ & Neural Network (NN) ansatz \\
  \hline
  $\theta_i$ & Trainable NN parameters\\
  \hline
  $\mathrm{D}$ & Dimension set $\{1,2,\dots d\}$ for a $d-$dimensional PDE\\
  \hline
  $\lambda_i$ & Weight on loss component-$i$\\
  \hline
  $x_i$ & Input along dimension-$i$\\
  \hline
  $X_i$ & Input set of dimensional features $\{x_1, x_2\dots x_{d/3}\}$ for body network-$i$\\
  \hline
  $\{p, q, r\}$ & A stochastically sampled set of active dimensions where p, q, r are assigned to body network 1, 2, 3 respectively\\
  \hline
  $\mathcal{G}$ & Grid of boundary/collocation\\
  \hline
  $\mathcal{J}$ & Loss function\\
  \hline
  $N$ & Number of collocation points\\
  \hline
  $B$ & Number of body networks\\
  \hline
  $r$ & Size of the embedding layer in each body network\\
  \hline
  $H$ & Hessian of the loss function\\
  \hline
  $\Omega$, $\partial\Omega$ & Domain, Domain Boundary\\
  \hline
  $f_i$, $g_j$ & Body Networks in Anant-Net as mentioned in Section \ref{sec:UAT} on UAT\\
  \hline
  $\hat{\mathcal{B}}$ & Set of active dimension sets as mentioned in Section \ref{sec:UAT} on UAT\\
  \hline
\end{tabular}
\end{center}
\caption{Definition of symbols used in the paper.}
\end{table}

\end{document}